\documentclass{ucetd}

\usepackage{./myStyle}

\begin{document}
\maketitle


\tableofcontents
\listoffigures
\listoftables

\acknowledgments
I want to thank my parents who have always supported me. I owe this day to them.  I love you mom and pop.  
I want to thank all of my professors, especially the professors who went above and beyond as mentors.  Thank you Greg for helping me beyond researching ideas to also thinking about life, and what I will do beyond schooling.
Thank all of my friends and colleagues, Dan, Mohammadreza, Mantas, Gustav, and Payman all of whom in one way or another helped me achieve this goal.
Also would like to thank my friends Matt and Kevin.
I would also like to thank Keziah for helping me edit the thesis.
I would like to thank Victoria for also helping with edits, and being by my side.
Finally I would like to thank everyone who have helped to make this day possible.

\abstract
We introduce several new datasets namely ImageNet-A/O and ImageNet-R as well as a synthetic environment and testing suite we called CAOS.  ImageNet-A/O allow researchers to focus in on the blind spots remaining in ImageNet.  ImageNet-R was specifically created with the intention of tracking robust representation as the representations are no longer simply natural but include artistic, and other renditions.  The CAOS suite is built off of CARLA simulator which allows for the inclusion of anomalous objects and can create reproducible synthetic environment and scenes for testing robustness.  All of the datasets were created for testing robustness and measuring progress in robustness.  The datasets have been used in various other works to measure their own progress in robustness and allowing for tangential progress that does not focus exclusively on natural accuracy.

Given these datasets, we created several novel methods that aim to advance robustness research.  We build off of simple baselines in the form of Maximum Logit, and Typicality Score as well as create a novel data augmentation method in the form of DeepAugment that improves on the aforementioned benchmarks.  Maximum Logit considers the logit values instead of the values after the softmax operation, while a small change  produces noticeable improvements.  The Typicality Score compares the output distribution to a posterior distribution over classes.  We show that this improves performance over the baseline in all but the segmentation task.  Speculating that perhaps at the pixel level the semantic information of a pixel is less meaningful than that of class level information.  Finally the new augmentation technique of DeepAugment utilizes neural networks to create augmentations on images that are radically different than the traditional geometric and camera based transformations used previously.  DeepAugment improves SOTA by a significant margin while being able to be used with other augmentation schemes and demonstrates that neural augmentations are not only possible but provide a benefit with respect to robustness.

\mainmatter

\chapter{Introduction}

Machine Learning (ML) models are becoming more widespread in their use and adoption.  As their use becomes more prevalent in safety-critical applications or trust necessary situations, the models must exhibit reliability to ensure their continued adoption.  Some of the current domains that exhibit these properties include self-driving vehicles and in medical diagnoses; where in self-driving, both trust and safety are paramount.  In the second domain of medical diagnoses trust is the primary factor such that a model will need to be able to provide explanations or a confidence such that other professionals can best decide how to proceed with the outputted information.


Robustness is of especial importance in machine learning where many of the commonly used models are uncalibrated out of the box.
All of the works thus far that aimed at addressing this short-coming also come at the cost of model accuracy.  While calibration is of importance for building confidence in model predictions, it only represents one facet of robustness.  In this thesis we will cover several aspects of robustness and begin to show that robustness covers several related areas.  While we explore the related areas, we also aim to highlight the delineations between them.  These delineations help researchers better focus on specific problem domains, which then allows the community to make improvements. Beyond covering the delineations, we also demonstrate that there are still unknown areas of robustness that might not have proper categories or delineations at the moment. 

We define \textit{robustness} as the preservation of functionality under changing conditions. Within the ML community robustness has grown to encompass several different concepts which include the following domains: generalization, sensitivity, distribution shift, adversarial examples, out-of-distribution (OOD) detection, calibration, and even interpretability.  We shall attempt to make those distinctions clear.  
We give a detailed explanations of the domains in Section \ref{sec:detail_robustness} and a brief description of the OOD detection task in Section \ref{sec:Tasks}. This task fits into robustness by trying to preserve confidence on in-distribution (ID) examples in the presence of OOD examples.  Therefore, the OOD detection task is useful for building safe ML systems. For example in robotics systems it can serve as an indicator for the robot to hand off to an operator and in medicine, it can signal that the input is malformed and should be redone or handled by a doctor. 

We organize the thesis loosely based around machine learning tasks covering multi-class classification first, then segmentation, then multi-label, and finally meta-learning. Segmentation can be viewed as a multi-class classification problem applied to every individual pixel.  Multi-label classification can be thought of as an extension of multi-class classification so we cover multi-label after multi-class and segmentation. This extension is not perfect and we show some limitations that comes from trying to generalize successes from multi-class to multi-label.  We then discuss meta-learning with respect to robustness which is distinct enough from multi-class and multi-label to be covered last.

\pagebreak

\section{Contributions}

Our contributions to robustness research are as follows:

\begin{itemize}  
\item Datasets: In this thesis we contribute three distinct datasets to help push the research area of robustness further.  The first dataset explores the current errors that models make and are termed hard-negative examples or natural adversarial examples in Chapter \ref{ch:nae_overview}. The second dataset involves the creation of a controllable anomalous segmentation dataset created by modifying the CARLA simulator in Chapter \ref{ch:anom_segmentation}.  Finally, the third dataset aims to better explore distribution shift from ImageNet-1K to other representations such as cartoons, sculptures, or tattoos among others in Chapter \ref{ch:neural_aug}.

\item New Robustness Techniques: We utilize the Kullback–Leibler divergence (KL-Divergence) to improve OOD detection in the multi-class setting. We further show that using the maximum logit serves as a better indicator of OOD detection than the previous baseline. Both of those results are presented in Chapter \ref{ch:anom_segmentation}. We also develop a novel non-linear augmentation technique in Chapter \ref{ch:neural_aug}.  The novel technique is able to greatly improve robustness in the multi-class setting and highlights a new unexplored area that can also help generalization.  

\item Novel Field Criticisms:  By examining different variations of robustness, we are able to observe where current techniques are still lacking in Chapters \ref{ch:neural_aug} and \ref{ch:multilabelood}.  Most notably is the recognition that almost all of the previous robustness techniques failed to generalize to the multi-label setting.  

\item Few shot robustness: We explore the relationship of robustness and few shot learning in Chapter \ref{ch:fs_robustness}.  By converting the metric learning task into a set-membership task we find that while the models are able to learn set-membership for in-distribution classes, the models are unable to generalize to novel unseen classes.   

\end{itemize}


\chapter{Background}

\section{Robustness} \label{sec:detail_robustness}

We shall cover a few seemingly disparate robustness topics and show how they relate together.  In this thesis, we focus on natural out-of-distribution (OOD) detection. Since ML systems are deployed in natural settings, natural OOD detection is of utmost importance and should be solved immediately. In addition to natural OOD detection, in Chapter \ref{ch:anom_segmentation}, we explore anomaly segmentation in simulated and adversarial examples.  We explore these areas to study whether any of the techniques in simulations or improvements in adversarial robustness generalize to natural settings.  
In Section \ref{sec:adversarial_ml} we cover adversarial examples in greater detail. 

\begin{figure}[ht]
\begin{subfigure}{.45\textwidth}
    \centering
    \includegraphics[width=0.9\columnwidth]{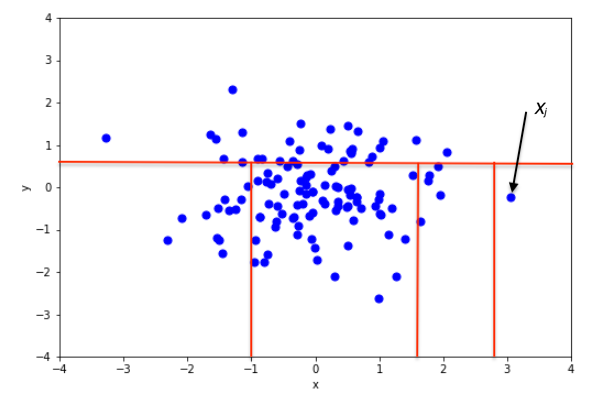}
    \caption{Points are drawn from a 2D Gaussian, and the isolation forest model is labeling the point X as an outlier.}
\end{subfigure}
\begin{subfigure}{.5\textwidth}
    \centering
    \includegraphics[width=0.9\columnwidth]{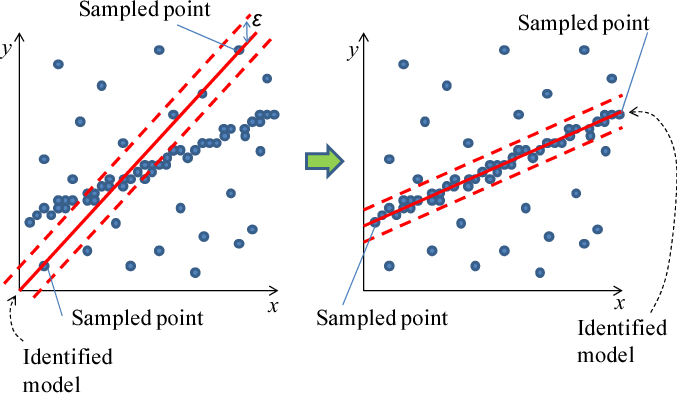}
    \caption{Given differently sampled points leads to different lines which causes some number of points to be labeled as outliers.  After sampling sufficient number of points then the best line is determined. Image credit: \cite{Watanabe2013AFR}.}
\end{subfigure}
\caption{Examples of two novelty detection methods.}
\end{figure}\label{fig:outlier_det}

OOD detection has its roots in outlier and novelty detection.  
The main distinction between the more recent OOD detection and outlier detection are the assumptions on what is considered an outlier.  In OOD detection, the task is set up similarly to a classification problem by having an in-distribution set, and an unknown and much larger out-of-distribution set.  Whereas in outlier detection the assumption would be that there exists a small subset of data which are considered outliers, and hence the data provided is considered ``poisoned.''  The commonly used approach in statistics is to first define the model and identify the outliers as points with low probability mass. More advanced methods exist such as using local outlier factor \parencite{Breunig2000LOFID} or using RANSAC \parencite{Fischler1981Ransac} to determine the inliers versus outliers.  See figure \ref{fig:outlier_det} for a visual demonstration of the techniques.  The former method works by comparing the distances of points to their neighbors and finding a threshold for the computed distance.  However many of these methods still rely on determining the model first to fit the data.

In a setting closely related to outlier detection, there is the problem of training with known label corruptions. Both domains deal with the case of data poisoning, while this sub-problem exclusively concerns itself with poison occurring in the labels.  Furthermore the problem of label corruption can be approached from the assumption whether there exists a subset of trusted labels.  \cite{Charikar2017SteinhardtsemiVerified} analyze both conditions to give theoretical guarantees and an algorithm to use under each setting.  
\cite{Patrini2017LabelNoise} address the problem in multi-class classification with no trusted data where they first estimate the level of corruption, then given the confusion matrix apply it after the network's outputs to correct for the uncertainty.  \cite{Ren2018ReweightExamplesLabelCorruption} also approach the problem under the same assumptions, but they use meta-learning (see Section \ref{sec:meta_learning}) to assign and reweight the examples used for training.  
Follow up work has shown that MixUp is also an effective way to increase the training given label corruption \parencite{arazo2019unsupervised}.  While the problem remains unsolved, some works highlight that for some models, namely neural networks, label noise might not be as much of a problem compared to other ML methods as the models tend to be robust to out of the box \parencite{Rolnick2017MassiveLabelNoise}.  

On the other hand, novelty detection, unlike in outlier detection, operates under the assumption that the dataset is not polluted with outliers, and instead tries to detect if a new example is an outlier.  Some classical techniques in novelty detection are that of isolation forest \parencite{Liu2008IsolationF} and one-class SVM \parencite{Schlkopf1999OCSVM}.  These methods employ a training and testing phase.  Some of the methods can not give estimations of outliers for the training set by construction.  This set-up more closely reflects the problems in OOD detection.  However, many of the novelty detection methods such as those listed exhibit poor performance \parencite{emmott2013ocsvmrobustnessreview}. 

In contrast to outlier and novelty detection, OOD detection frames the problem of detecting anomalies as a learning problem.  Novelty detection can be considered to be within unsupervised classification, while OOD detection expands on novelty detection by considering supervised and semi-supervised along with unsupervised classification.  
\cite{hendrycks2017baseline} uses a trained neural network's output as a confidence score for detecting anomalies in the setting of multi-class classification.  
As a followup work, ODIN \parencite{Liang2018ODIN} and Mahalanobis detector \parencite{Lee2018Mahalanobis} apply small gradient perturbations derived from label corruptions and then run the corrupted image through the network again to get the final output score.  
The work by \cite{Lee2018Mahalanobis} modifies this approach by training a per-class dependent model on the perturbations.  The main issue with the previous two approaches in particular is that they fine-tune their corruptions on the different perturbation types which leads to a form of training on the test set.  
Other methods such as the confidence estimator trains a separate confidence branch to directly predict both the class label and the confidence of the prediction \parencite{devries}.  In Chapter \ref{ch:neural_aug} we further explore the relationship between data augmentation and robustness by demonstrating how random corruptions by a neural network improve OOD detection.

\section{Adversarial Examples} \label{sec:adversarial_ml}

\begin{figure}[!ht]
    \centering
    \includegraphics[width=0.8\columnwidth]{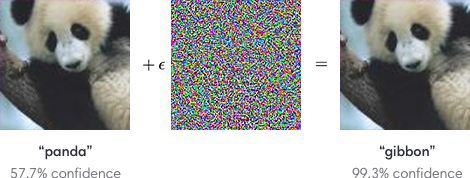}
    \caption{An adversarial example showing how a small perturbation can change the label of an image. The adversarial noise added to image is magnified by 1000x to be visible to a human.  Image credit: OpenAI}
    \label{fig:adversarial_panda_intro}
\end{figure}

Adversarial examples are modifications to inputs, such as images, that causes the label of a function (typically a deep neural network) to change, while a human would still classify the input as belonging to the same class. It was first discovered that neural networks are susceptible to these small perturbations in \cite{Szegedy2014IntriguingPO}. Figure \ref{fig:adversarial_panda_intro} shows how small adversarial noise can affect the outputs of a deep neural network (DNN).  This work spawned several works in interpretability such as \cite{olah2017DeepDream} and \cite{Selvaraju_2019GradCam} whereby the gradient perturbations were used to visualize what the network was attending to. 
Later on after \parencite{Carlini2017Wagner}'s research findings, the research community began to focus on the problem treating it as an actual security threat threat instead of as an anomaly.

Let us formally define adversarial examples.  Note for the purposes of these definitions we will restrict the domain of examples or inputs to be that of images but it can extended to other domains such as speech or language.  An input image is $\vec{x}$ and a trained model is $f$ (typically a neural network). $f(\cdot)$ is the output probability distribution for a set of classes the model was trained on.  We will perturb $\vec{x}$ by a small amount $\epsilon$ which will result in our adversarial example $\vec{x}' := \vec{x} + \epsilon$.  We say $\vec{x}'$ is an adversarial example if $\underset{\vec{x}}{\arg\max} \ f(\vec{x}) \neq \underset{\vec{x}'}{\arg\max} \ f(\vec{x}')$, it is also described as a successful attack.  $\epsilon$ is of the same dimensionality as $\vec{x}$ but is bounded by some distance metric.  We use several $L_p$ distance measures to bound $\epsilon$ as a proxy for the measure we actually care for, which is human perceptual distance.  More recently, there is work trying to move beyond $\ell_p$ distances such as \parencite{Bhattad2020UnrestrictedAEcolorizationAttack} because bounding $ \|\varepsilon\|_{\infty} \leq \delta$ distance is a somewhat arbitrary limitation on the attacker. 

It is common practice in the security research community to detail what the assumptions are concerning the threat model, which includes what the attacker has access to, capabilities, and limitations.  Most of what has so far been discussed relates to the scenario where the attacker has full access to the machine and the model such that they can inject signal noise $\epsilon$ into the system to be able to change the resulting classification output of $f$.  While this setting has received much attention, it is largely unrealistic as an actual security threat.  More recently, there have been efforts to change the threat model to consider more real world attacks such as \parencite{Eykholt2018PhysicalAE, Goodfellow2015ExplainingAH, Yakura_2019AudioAlexAttack} where the attacker has a more limited access to the model via its external sensors.   

As an implicit goal in desiring to make the models more similar to humans, we treat adversarial examples as inherent flaws in the model.  There are some discussions concerning whether or not adversarial examples are unavoidable \parencite{engstrom2019DiscussionofAdversarialBugs}, although this view is currently held by a minority in the research community.  We shall detail a few of the current hypotheses being investigated for why adversarial examples might arise.  One of the hypotheses by \cite{Goodfellow2015ExplainingAH} purports that the brittleness is a result of excessive linearity. The high-dimensionality of the inputs allows an adversary easy access to cross a decision boundary of a class.  The simple example is for a two class problem given a powerful enough adversary, the inputs can be moved along any direction and with high probability half of the directions will cross the decision boundary. This hypothesis has been indirectly challenged by subsequent work \parencite{wong2018adversarialpolytope}. 
Another hypothesis is that of robust and non-robust features \parencite{Ilyas2019AdversarialAreBugs}.  The argument is that the networks are picking up on highly predictive features that do well on the training set, but are too brittle and incomprehensible for humans which then allows for adversarial examples.  Some qualitative examples supporting this claim are that adversarially trained models have features that are more recognizable to humans.  The features themselves are trained on limited datasets which creates an artificial selection bias and lacks more theoretical underpinnings.  There is no definitive consensus for explanations or solutions to the problem of adversarial examples at the current time.

While there does not exist a solution to the problem, there are several defenses that can mitigate the power of adversarial attacks.  When a defense is ``broken'' it refers to the attacker reducing the accuracy of the model to zero (or nearly zero).  The most prominent defenses for adversarial examples are some form of training with adversarial examples.  The original formulation from \cite{Madry2018AdversarialTraining} adds an adversarial example during training as extra training data for the model to learn from.  Followup work from \cite{Zhang2019TRADES} explicitly considers the tradeoff between natural and adversarial accuracy by incorporating an additional loss term for the adversarial examples used during training.  There are other works that provide certifications for limited threat models such as $L_2$ distance \parencite{Cohen2019SmoothingAdversarialCertification} and other forms of certifications are being explored.  Methods which provide some amount of robustness include the following: Compression of either inputs, models, or both \parencite{Liu2019FeatureDD, Dziedzic2019BandlimitedTA};  injecting noise into either the input examples, models, or both \parencite{You2019AdversarialNL, Cohen2019SmoothingAdversarialCertification}; data augmentation, or pre-training \parencite{Hendrycks2019Pretraining}; and ensembling which use multiple models to give one output using a voting, or some other scheme \parencite{Tramr2018EnsembleAT}.

Part of the difficulty with comparing defenses is the lack of standardization to compare the different threat models and environments.  Recently, projected gradient descent attacks with $\ell_{\infty}$ perturbation of size 8/255 has become the standard because greater perturbations allowed for changing the label to a human annotator \parencite{Tramr2020FundamentalMnistSensitivityChangingLabel}. Given this limited adversary/scenario allows for direct comparison between methods.  The other issue is that the threat model or environment is being an unfair or being an unreasonable limitation to the adversary.  Due to this limitation researchers use adative attacks \parencite{Tramr2020OnAdaptiveAttacks} which allow the attacker to know the defense and modify the attack given this knowledge.  Given the relative infancy of the joint domains of ML and computer security, the field will eventually settle on realistic threat models and the issue of defense comparison will become moot.  

Although the focus is mostly on neural networks, other classical machine learning (ML) models are also susceptible to adversarial noise such as k Nearest Neighbors (k-NN), SVMs, and linear regression.  These other machine learning models have different failure modes for adversarial examples but have all been susceptible to the attacks.  \cite{gilmer19advconsequnce_of_noise_iclr} suggests that the previous methods fail due to the high dimensionality of the inputs.  Currently neural networks provide the best defenses against adversarial examples \parencite{Goodfellow2016comparemodels}.  However, the other ML models have been less explored both to attack and defend against this threat model.

It is worth noting the difficulty of transitioning to real-world attacks via adversarial examples.  \cite{Bhattad2020UnrestrictedAEcolorizationAttack} demonstrates that artificial constraints had to be placed on the model for the adversary to succeed.  
This is not to suggest that the signal based attacks are unimportant, but it should be studied to determine how well adversarial attacks can survive the transition to complex environments, to 3D, and against multiple modalities.  \cite{carla} have shown how difficult it is to construct adversarial examples in 3D environments.  

\section{Meta-Learning} \label{sec:meta_learning}

Meta-learning is most commonly understood as the task of ``learning to learn'', which refers to the process of improving a learning algorithm over distributions. This contrasts with the conventional machine learning, which is the process of learning a model for a single distribution or over many data instances.  
Meta-learning is comprised of a two stage process.  The first stage involves a base learner and the second stage (also confusingly called meta-learning) involves an outer algorithm that optimizes an outer objective.  An example of meta-learning the inner (or lower, base) learning algorithm solves a task such as image classification \parencite{franceschi2018bilevel}, 
defined by a dataset and objective. Then during meta-learning, an outer (or upper, meta) algorithm updates the inner one, so that the learned model can perform well in few-shot learning \parencite{hospedales2020metalearning}.

Recent work has tried to minimize the distinctions between meta-learning and supervised classification \parencite{chao2020revisiting}.  \cite{maurer2005algorithmic} extended the generalization error bounds from supervised learning to meta-learning.   Formulating the problem of meta-learning more closely with supervised learning better fits the theme of this thesis trying to improve OOD detection.  

We most prominently explore two techniques in meta-learning.  The first technique is from \cite{finn2017MAML} which uses the optimization algorithm to perform the meta-learning.  The inner learner does standard supervised learning for the task, e.g. image classification, and the outer optimization algorithm does an averaged gradient step for several tasks (different image classification tasks).  The second technique imbues the space of learned features with a metric to be later used for a nearest neighbor classification.
We ignore the other type of meta-learning which uses memory augmented networks to perform the meta-learning.  Those models are currently more akin to a novelty as opposed to offering a more utilized method such as the others described before it.

We explore meta-learning in Chapter \ref{ch:fs_robustness} to explore how the techniques from meta-learning namely the scenario of learning in the few shot setting can better assist in generalizing to unseen OOD examples.  In both scenarios the amount of OOD data far exceeds the amount of training data that the model is expected to generalize from. 


\section{Tasks} \label{sec:Tasks}

In this section, we will cover three computer vision tasks that will be addressed in this paper namely  multi-class classification, multi-label classification, and segmentation,.

\subsection{Multi-class Classification}

Multi-class classification is a rich sub-field in machine learning with its origins in binary classification \parencite{Cox1958logisticregression}.  From binary classification there are three approaches that have been taken to handle the multi-class setting.  The first approach is by reduction of the multi-class problem to binary classification, typically One-vs-Rest where one would use a classifier to distinguish one class from all of the others.  The size of these approaches scale linearly with the number of classes due to requiring a different model per class.  The second approach handles the problem by extending the binary classification to multi-class classification directly.  Finally, the last approach handles the problems by creating hierarchies for classification \parencite{Silla2010ASO}.

We will be covering the second approach for multi-class classification, the extension from binary, which include methods such as k-nearest-neighbors \parencite{Altman1992KNN}, logistic regression \parencite{menard2002appliedLR}, neural networks \parencite{Hopfield1988NeuralNA}, and random forests \parencite{Breiman2004RandomForests}.  We will briefly cover neural networks as they are utilized throughout the paper and are used in the other tasks as well.  For neural networks we mostly utilize convolutional neural networks (CNN). CNNs is composed of the following operations: learned filters which perform cross-correlation, convolution, non-linearity, and pooling which reduces the dimensionality of the inputs and features. These operations are repeated in succession to form what are considered layers before finally applying a differentiable loss function so that the network can learn via stochastic gradient descent.  
The specific implementation details of each neural network architecture is left to the respective chapter.


The problem in multi-class classification is to associate each instance with a unique label from a set of labels.  More formally given a set of training data $D = \{(\vec{x}_i, \vec{y}_i)\} , \forall i \in \{1, ..., N\}$ where $\vec{x} \in \mathbb{R}^{d}$, the goal is to learn (or induce) a model $f(\vec{x}) = \vec{\hat{y}}$ such that $\vec{\hat{y}}$ minimizes a loss function $L: \mathcal{R} \times \mathcal{R}\to\mathcal{R}$.  During classification, the assumption is training and test data points are drawn i.i.d. (independent and identically distributed) from a full joint distribution.  Although in most cases, the full joint distribution is unknown, a finite number of examples can be used to measure generalization by splitting the data into training and test sets. 
Generative and discriminative techniques are useful for learning model $f$. Generative techniques learn the full joint probability distribution while the latter techniques model the class or decision boundaries.  In this thesis, we focus on discriminative methods for classification. 

Some influential datasets in the area of multi-class classification have been MNIST \parencite{mnist}, SVHN \parencite{svhn}, CIFAR \parencite{cifar}, and ImageNet \parencite{imagenet}.  MNIST and SVHN are both datasets consisting of the digits 0-9. MNIST are black and white digits, and SVHN are numbers occurring on houses which are colored images and have a variety of backgrounds.  On the other hand, CIFAR and ImageNet are natural images consisting of animals, automotive vehicles, and many other categories.  The former are small images with ten categories and the latter are large images with one thousand categories.  These datasets, among others, have allowed for great advances in the task of multi-class classification \parencite{Ranzato2006LecunSparseEnergyModel,Beygelzimer2006CoverTreesNN,Cho2009KernelMethods,AlexNet,Carlini2017Wagner,}. 

\subsection{Multi-label Classification}

\begin{figure}[!ht]
    \centering
    \includegraphics[width=0.8\columnwidth]{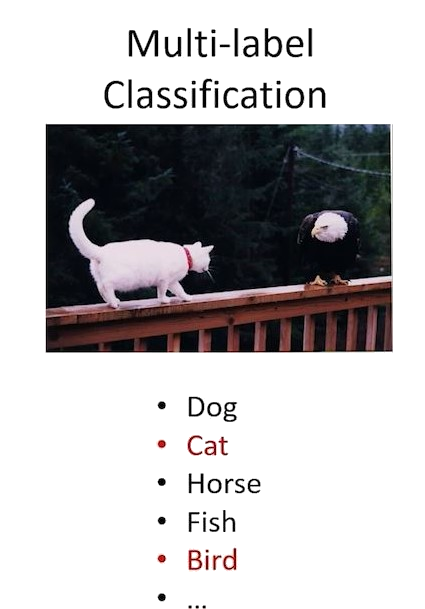}
    \caption{An example of an image with both a cat and a bird which shows multiple labels occurring together.}
    \label{fig:intro_multilabel_image}
\end{figure}

Building from multi-class classification, there is multi-label classification whereby instead of the traditional setting where an instance may be of only one label, in this setting each instance is associated with a set of labels.  
As an example consider the figure \ref{fig:intro_multilabel_image} where it can be classified as a bird, bald eagle, cat, porch and forest.  This task being a natural extension of multi-class is therefore a harder task because of determining the presence or absence of for all classes as opposed to the problem of selecting the most likely candidate.  Multi-label classification technique is useful for (1) text classification involving lots of documents about several topics, (2) audio classification, where some songs are a mixture of styles or genres and (3) image search where different images or videos can capture multiple genres or styles at the same time \parencite{Brhanie2016MultiLabelCM}. 

We will similarly define the task of multi-label classification.  The problem in multi-label (or sometimes referred to as multi-category or multi-topic) classification is to associate each instance with a set of labels.  This again follows the same conventions and notations as multi-class classification whereby training data $D = \{(\vec{x}_i, \vec{y}_i)\} , \forall i \in \{1, ..., N\}$ contain examples x that are a $d$-dimensional vector.  However, unlike in multi-class setting y is no longer a one-hot vector and are instead binary strings without the limitation of only one ``$1$'' in the string.

Multi-label classification is arguably better for studying robustness.  The main reasons for this is because
in multi-class classification the models are either forced to pick one out of the set of labels or add in a catch-all ``other'' class.  In the first scenario the model ``should'' output a uniform probability distribution over all classes, however, this also seems unrealistic as there will exist OOD examples which will appear to be similar to in-distribution classes and so a uniform output appears to be an unfeasible goal.  The second option seems more promising but still has a similar issue.  Classifying images or parts of image as background will lead to the model ``correctly'' classifying OOD examples as being background.  This scenario again presents issues that there is no longer any distinction between in and out-of-distribution examples, which can be a safety hazard in some scenarios.  Multi-label classification does not have these two issues because having a uniform output of no class is a feasible solution that is also realistic.  The secondary issue is still present unless modifications are made, that of being able to discern between in- and out-of-distribution examples.

Unfortunately, multi-label classification has not received as much attention within computer vision as that of multi-class classification.  This can be observed by the historical lag in multi-label datasets from Corel-5k \parencite{corel5k}, then PASCAL VOC \parencite{Everingham2009ThePV_pascal} and more recently we have MSCOCO \parencite{coco} and Tencent ML-image \cite{tencent-ml-images-2019}.  Similarly multi-label OOD is also a wholly underexplored area.  We hope to change this by releasing some baselines for this task in Chapter \ref{ch:multilabelood}.

\subsection{Segmentation}

\begin{figure}[!ht]
    \centering
    \includegraphics[width=0.8\columnwidth]{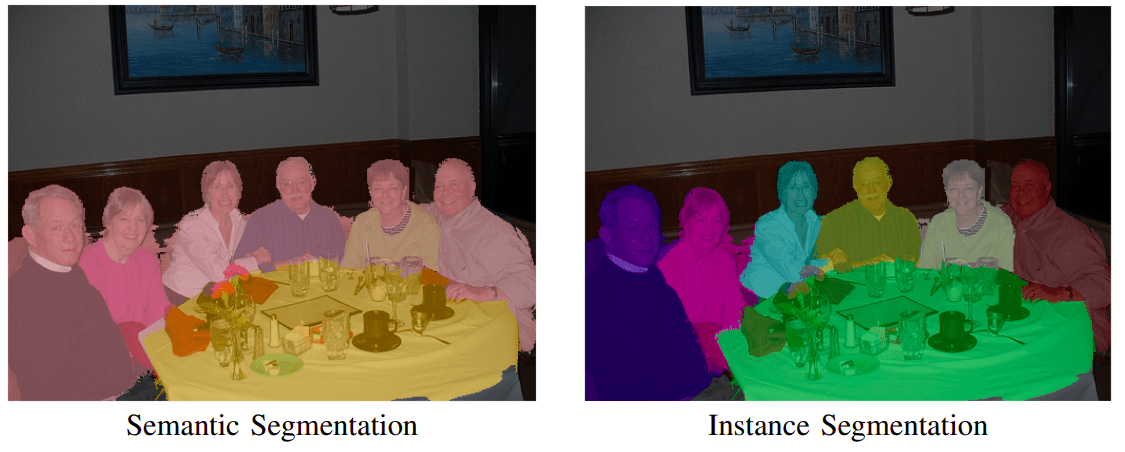}
    \caption{Highlighting the distinction between semantic segmentation on the left and instance segmentation on the right. Image source: \cite{Arnab2018ConditionalRFNN}}
    \label{fig:semantic_vs_instance_segmentation}
\end{figure}

Segmentation refers to the task of partitioning the pixels of an image into regions or segments, where the pixels in each region share similar attributes.  The goal of segmentation is to simplify and/or change the representation of an image into something  meaningful and easier to analyze \parencite{shapiro_stockman_2001}.  
We categorize segmentation methods into two main categories that of semantic and class agnostic segmentation.  Class agnostic segmentation is the task of segmenting an image into the region boundaries such as foreground and background. Semantic segmentation on the other hand, deals with the task of assigning each pixel to an element of a set of classes.  As a related task to semantic segmentation there is instance segmentation where the goal is to identify all distinct occurrences of the segmented objects in an image.  

Class agonostic segmentation involves partitioning an image into coherent regions.  \cite{BSDSmartin2001} show in Berkeley Segmentation Dataset (BSDS) that human annotations of edges and boundaries are nonrandom and useful for supervised class agnostic segmentation. However, the BSDS dataset has been most useful in unsupervised segmentation \parencite{arbelaez2010contour, martin2004learning, achanta2010slic, carreira2010constrained}, in such tasks as foreground/background segmentation \parencite{wu2008interactive}.    Unsupervised segmentation typically involve clustering \parencite{martin2004learning} and are trained by learning a distance between pixels within segments.  Their outputs have been used in other downstream tasks due to the reduction in the input space \parencite{mostajabi2015zoomout}.  Due to the hardware and algorithmic improvements, reliance on ``super pixels'' \parencite{achanta2010slic} has waned and the current approaches label each pixel directly in a given image.    

Unlike class agnostic segmentation, semantic segmentation is a supervised task where the goal is to assign every pixel in an image a label from a given set of classes.  In semantic segmentation multiple instances of the same object class are not differentiated.  While in instance segmentation the goal is to assign a unique label to every object instance in an image.  A limitation of instance segmentation is with dealing with uncountable classes such as pixels belonging to the sky or sand.  The union of both semantic and instance segmentation is known as panoptic segmentation \parencite{Kirillov2019PanopticS}. 
The goal of panoptic segmentation is to assign both a category label to every pixel and if the category is an instance based category, such as people, then also assign it a unique id. 

Prominent segmentation datasets include BSDS \parencite{BSDSmartin2001}, PASCAL-VOC \parencite{Everingham2009ThePV_pascal} and MS COCO \parencite{coco}.  These datasets have allowed for significant progress in unsupervised and supervised segmentation \parencite{arbelaez2010contour, objectdetectionanalysis, wu2008interactive, Arnab2018ConditionalRFNN, alex2019pointrend}.

\chapter{Natural Adversarial Examples}\label{ch:nae_overview}

\section{Overview}

\begin{figure}[t]
    \centering
    \includegraphics[width=\textwidth]{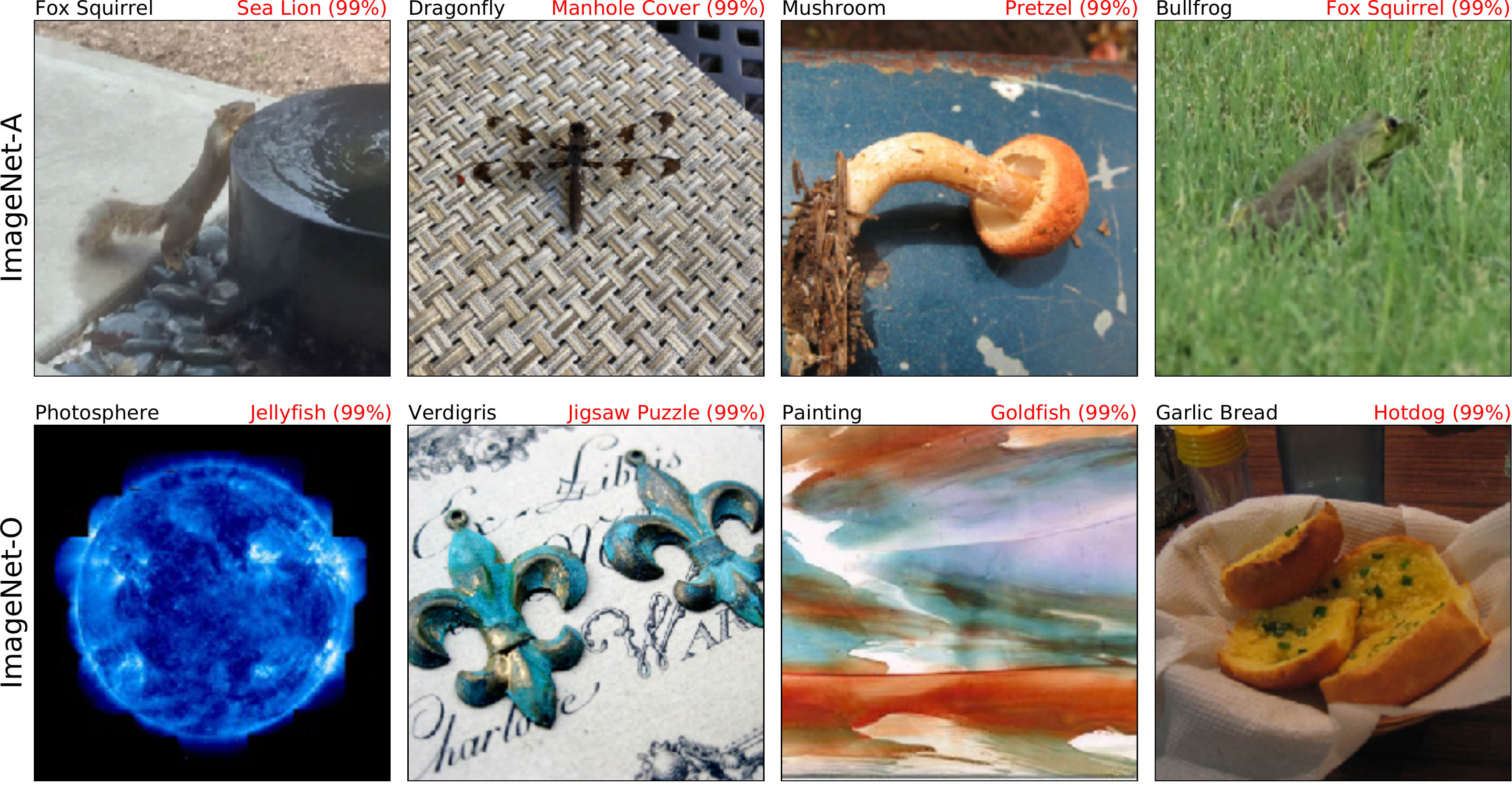}
\caption{Examples from \textsc{ImageNet-A} and \textsc{ImageNet-O}. The black text is the actual class, and the red text is a ResNet-50 prediction and its confidence. \textsc{ImageNet-A} contains images that classifiers should be able to classify. \textsc{ImageNet-O} contains out-of-distribution anomalies of unforeseen classes which should result in low-confidence predictions. ImageNet-1K models do not train on examples from "Photosphere" nor "Verdigris" classes, so these images are out-of-distribution.
}\label{fig:splash}
\vspace{4pt}
\end{figure}

Research on the ImageNet \parencite{imagenet} benchmark has led to numerous advances in classification \parencite{Krizhevsky2017Alexnet}, object detection \parencite{huang2017speed}, and segmentation \parencite{he2017mask}. 
ImageNet classification improvements are broadly applicable and highly predictive of improvements on many tasks \parencite{Kornblith2018DoBI}. Improvements on ImageNet classification have been so great that some call ImageNet classifiers "superhuman" \parencite{He2015DelvingDI}.
However, performance is decidedly subhuman when the test distribution does not match the training distribution \parencite{hendrycks2019robustness}. The distribution seen at test-time can include inclement weather conditions and obscured objects, and it can also include objects that are anomalous.\looseness=-1

\cite{Recht2019DoIC} remind us that ImageNet test examples tend to be simple, clear, close-up images, resulting in the current test set being perhaps too easy and not representative of harder images encountered in the real world. \cite{Geirhos2020ShortcutLI} and \cite{Arjovsky2019InvariantRM} argue that image classification datasets contain ``spurious cues'' or ``shortcuts.'' For instance, models may use an image's background to predict the foreground object's class; a cow tends to co-occur with a green pasture, and even though the background is inessential to the object's identity, models may predict ``cow'' primarily using the green pasture background cue. When datasets contain spurious cues, they can lead to performance estimates that are optimistic.

To counteract this, we curate two hard ImageNet test sets of adversarially filtered examples. By using adversarial filtration, we can test how well models perform when simple-to-classify examples are removed, including examples that are solved with simple spurious cues.  Some adversarially filtered examples are depicted in figure \ref{fig:splash}, which are simple for humans but hard for models.  Previously it has been shown that misclassified images can transfer between models; however, these demonstrations relied on synthetic distributions \parencite{geirhos,hendrycks2019robustness} and adversarial distortions \parencite{szegedy2013intriguing}.  Our examples demonstrate that is is possible to reliably fool many models with clean natural images highlighting a practical problem that needs to be addressed as opposed to a more theoretical or abstract problem.\looseness=-1

We demonstrate that clean examples that we collected can reliably degrade and transfer to other unseen classifiers with our first dataset.  Transfer in this setting means that the average performance of an unseen model is comparable (meaning not significantly better) to the model we used for filtering.  This phenomenon of transferring is hard to classify.   We call this dataset \textsc{ImageNet-A}, which contains images from a distribution unlike the ImageNet training distribution. \textsc{ImageNet-A} examples belong to ImageNet classes, but the examples are harder and transfer to other models. They cause consistent classification mistakes due to scene complications encountered in the long tail of scene configurations and by exploiting classifier blind spots (see Section \ref{sec:failures}). Since examples transfer reliably, this dataset shows models have previously unappreciated shared weaknesses.

The second dataset allows us to test model uncertainty estimates when semantic factors of the data distribution shift. Our second dataset is \textsc{ImageNet-O}, which contains image concepts from outside ImageNet-1K. These out-of-distribution images reliably cause models to mistake the examples as high-confidence in-distribution examples. To our knowledge this is the first dataset of anomalies or out-of-distribution examples developed to test ImageNet models. While \textsc{ImageNet-A} enables us to test image classification performance when the \emph{input data distribution shifts}, \textsc{ImageNet-O} enables us to test out-of-distribution detection performance when the \emph{label distribution shifts}.

\begin{figure}[t]
\centering
\begin{subfigure}{.5\textwidth}
\includegraphics[width=0.99\textwidth]{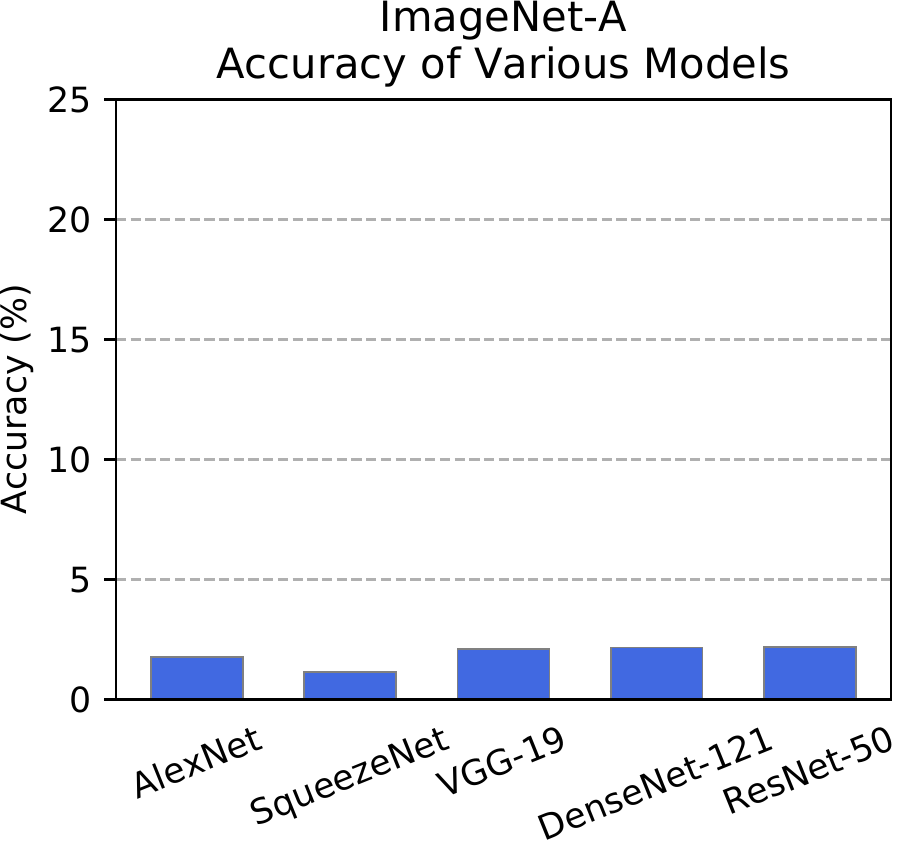}
\end{subfigure}%
\begin{subfigure}{.5\textwidth}
\includegraphics[width=0.99\textwidth]{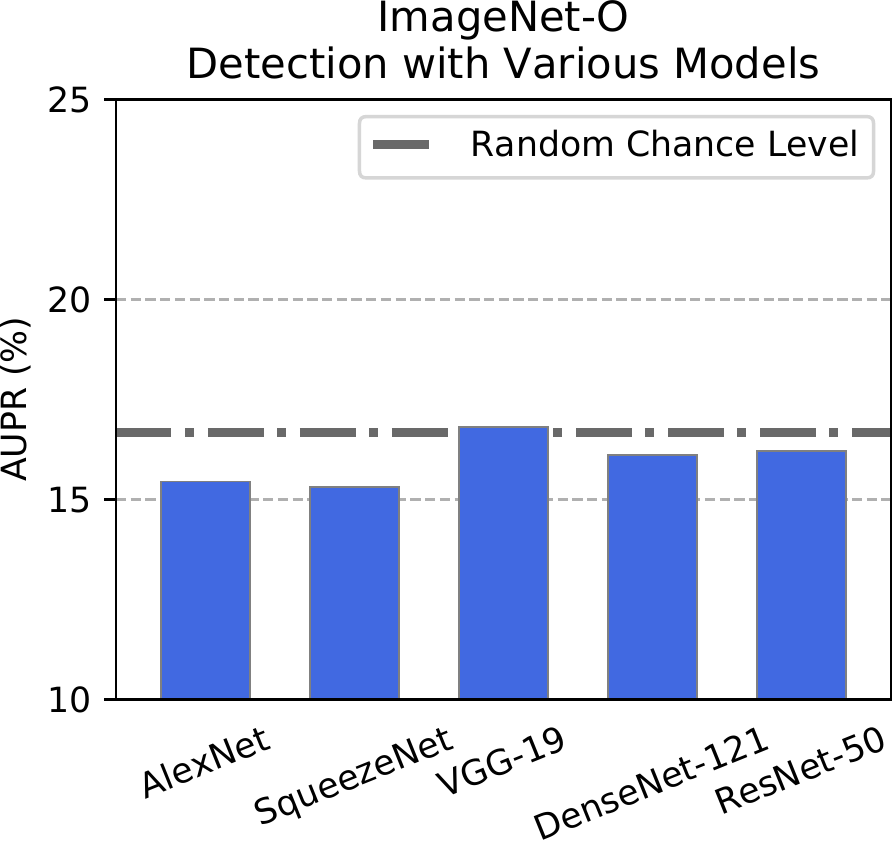}
\end{subfigure}
\caption{
Various ImageNet classifiers of different architectures fail to generalize well to \textsc{ImageNet-A} and \textsc{ImageNet-O}.
Higher Accuracy and higher AUPR is better. See \ref{sec:experiments} for a description of the AUPR out-of-distribution detection measure. These specific model parameters are unseen during adversarial filtration, so our adversarially filtered examples transfer across models. 
\looseness=-1
}\label{fig:modelsono}
\vspace{2pt}
\end{figure}

We examine methods to improve performance on adversarially filtered examples. However, this is difficult because figure \ref{fig:modelsono} shows that examples successfully transfer to unseen or black-box models.
To improve robustness, numerous techniques have been proposed. We find data augmentation techniques such as adversarial training decrease performance, while others can help by a few percent. We also find that a $10\times$ increase in training data corresponds to a less than a 10\% increase in accuracy. Finally, we show that improving model architectures is a promising avenue toward increasing robustness. Even so, current models have substantial room for improvement. 

\section{Related Work}

\noindent\textbf{Adversarial Examples.}\quad Real-world images may be chosen adversarially to cause performance decline. \cite{goodfellowblog} define adversarial examples \parencite{Szegedy2014IntriguingPO} as "inputs to machine learning models that an attacker has intentionally designed to cause the model to make a mistake". Most adversarial examples research centers around artificial $\ell_p$ adversarial examples, which are examples perturbed by nearly worst-case distortions that are small in an $\ell_p$ sense. Attackers can reliably and easily create black-box attacks by exploiting these consistent \emph{naturally occurring} model errors, and thus carefully applying gradient perturbations to create an artificial attack is unnecessary. This less restricted threat model has been discussed but not explored thoroughly before.

Several other forms of adversarial attacks have been considered in the literature, including elastic deformations \parencite{Xiao2018SpatiallyTA}, 
adversarial coloring \parencite{Bhattad2019BigBI,Hosseini2018SemanticAE}, 
 synthesis via generative models \parencite{Baluja2017AdversarialTN,Song2018ConstructingUA} 
and evolutionary search \parencite{Nguyen2015DeepNN}, among others. 
Other work has shown how to print 2D \parencite{Kurakin2017AdversarialEI,brown2017adversarial} 
or 3D \parencite{sharif2016accessorize,athalye2017synthesizing} objects that fool classifiers. 
These existing adversarial attacks are all based on synthesized images or objects, and some have questioned whether they provide a reliable window into real-world robustness \parencite{Gilmer2018MotivatingTR}.  
Our examples are closer in spirit to the hypothetical adversarial photographer discussed in \parencite{Brown2018UnrestrictedAE}, and by definition these adversarial photos occur in the real world.

\noindent\textbf{Adversarial Examples.}\quad Some types of adversarial attacks eschew $\ell_p$ norm ball constraints completely. For instance, \parencite{Baluja2017AdversarialTN} synthesize adversarial examples with generative adversarial networks, and \parencite{Song2018ConstructingUA} use variational autoencoders as well. Unfortunately, these examples are often classifier-specific and do not transfer to new models. Meanwhile, \textsc{ImageNet-A} adversarially filtered examples transfer and successfully attack current architectures. These adversarially filtered examples bear semblance to a theorized attack, the attack of the adversarial photographer \parencite{Brown2018UnrestrictedAE}. This attacker is free to take a photograph of an image with choice over the camera viewpoint, so that the attack is free of the confines of the $\ell_p$ norm ball constraints. This attack has not been thoroughly studied empirically, but these could be thought of as a type of natural adversarial example. \cite{Bhattad2019BigBI,Hosseini2018SemanticAE} attempt to color examples adversarially, and \cite{Xiao2018SpatiallyTA} create an adversarial elastic deformation. In both cases the modifications required to fool the network can become quite conspicuous \parencite{kang} and artificial. \cite{Kurakin2017AdversarialEI} show that $\ell_p$ adversarial examples can fool machine learning systems if they are carefully printed and if the perturbation is conspicuous. Meanwhile, adversarially filtered examples arise in nature and are not necessarily detectable by human vision alone. \cite{Nguyen2015DeepNN} also violate norm ball constraints and create adversarial examples through evolutionary algorithms, but these adversarial examples are far out-of-distribution and are not naturally manifested in the real world. 

\begin{figure}
\vspace{-10pt}
\centering
\includegraphics[width=0.48\textwidth]{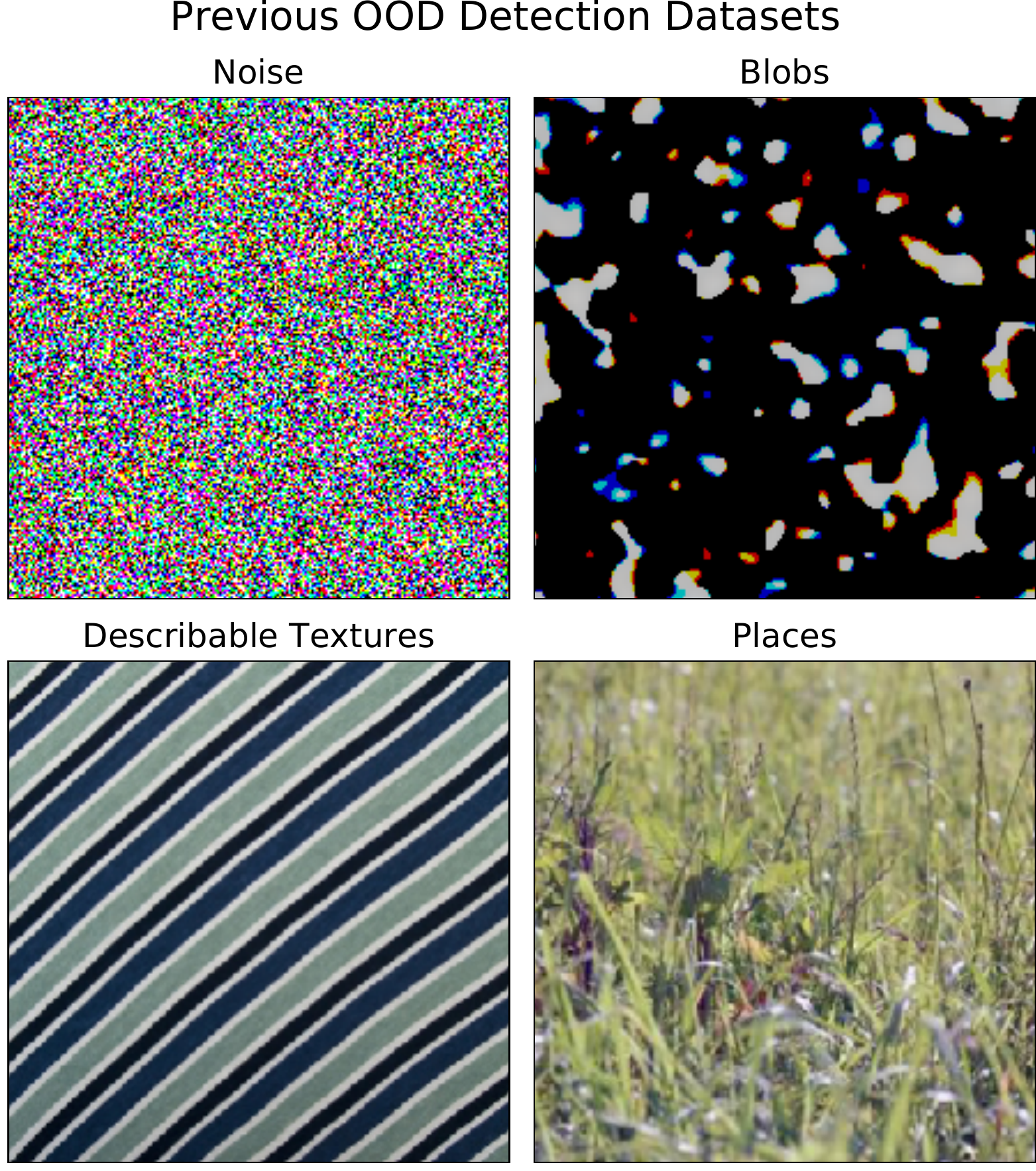}
\caption{
Previous work on out-of-distribution (OOD) detection uses synthetic anomalies and anomalies from wholly different data generating processes. For instance, previous work uses Bernoulli noise, blobs, the Describable Textures Dataset \parencite{cimpoi14describing}, and Places365 scenes \parencite{zhou2017places} to test ImageNet out-of-distribution detectors. To our knowledge, we propose the first dataset of out-of-distribution examples collected for ImageNet models.
}\label{fig:ood}
\vspace{2pt}
\end{figure}

\noindent\textbf{Out-of-Distribution Detection.}\quad 
Generally, models learn a distribution, such as the ImageNet-1K distribution, and are tasked with producing quality anomaly scores that distinguish between usual test set examples and examples from held-out anomalous distributions.
For instance, \cite{hendrycks2017baseline} treat CIFAR-10 as the in-distribution and treat Gaussian noise and the SUN scene dataset \parencite{Xiao2010SUNDL} as out-of-distribution data.
That paper also shows that the negative of the maximum softmax probability, or the the negative of the classifier prediction probability, is a high-performing anomaly score that can separate in- and out-of-distribution examples, so much so that it remains competitive to this day.
Since that time, other works on out-of-distribution detection continue to use datasets from other research benchmarks as stand-ins for out-of-distribution datasets. For example, some use the datasets shown in figure \ref{fig:ood} as out-of-distribution datasets \parencite{hendrycks2019oe}.
However, many of these anomaly sources are unnatural and deviate in numerous ways from the distribution of usual examples \parencite{Ahmed2019DetectingSA}. In fact, some of the distributions can be deemed anomalous from local image statistics alone. \cite{Meinke2019TowardsNN} propose studying adversarial out-of-distribution detection by detecting adversarially optimized uniform noise. In contrast, we propose a dataset for more realistic adversarial anomaly detection; our dataset contains hard anomalies generated by shifting the distribution's labels and keeping non-semantic factors similar to the in-distribution.

\noindent\textbf{Spurious Cues and Unintended Shortcuts.} Models may learn spurious cues and obtain high accuracy but for the wrong reasons \parencite{Lapuschkin2019UnmaskingCH}. 
Spurious cues are a studied problem in natural language processing \parencite{Cai2017PayAT,Gururangan2018AnnotationAI}. Many recently introduced datasets in NLP use adversarial filtration to create "adversarial datasets" by sieving examples solved with simple spurious cues \parencite{Sakaguchi2019WINOGRANDEAA,Bhagavatula2019AbductiveCR,Zellers2019HellaSwagCA,Dua2019DROPAR}. Like this recent concurrent research, we also use adversarial filtration \parencite{Sung1995LearningAE}, but the technique of adversarial filtration has not been applied to collecting image datasets before. Additionally, adversarial filtration in NLP uses filtration to remove only the easiest examples, while we use filtration to select only the hardest examples. Moreover, our examples transfer to weaker models, while in NLP the most used adversarial filtration technique AFLite \parencite{Sakaguchi2019WINOGRANDEAA} does not produce examples that transfer to less performant models \parencite{Bisk2019PIQARA}. We show that adversarial filtration algorithms can find examples that automatically and reliably transfer to both simpler and stronger models. Since adversarial filtration can remove examples that are solved by simple spurious cues, models must learn more robust features for our datasets.\looseness=-1 

\noindent\textbf{Robustness to Shifted Input Distributions.}\quad 
\cite{Recht2019DoIC} create a new ImageNet test set resembling the original test set as closely as possible. 
They found evidence that matching the difficulty of the original test set required selecting images deemed the easiest and most obvious by Mechanical Turkers.
\textsc{ImageNet-A} helps measure generalization to harder scenarios. \cite{Brendel2018ApproximatingCW} show that classifiers that do not know 
the spatial ordering of image regions can be competitive on the ImageNet test set, possibly due to the dataset's lack of difficulty. Judging classifiers by their performance on easier examples has potentially masked many of their shortcomings. For example, \cite{geirhos2019} artificially overwrite each ImageNet image's textures and conclude that classifiers learn to rely on textural cues and under-utilize information about object shape. Recent work shows that classifiers are highly susceptible to non-adversarial stochastic corruptions \parencite{hendrycks2019robustness}. While they distort images with 75 different algorithmically generated corruptions, our sources of distribution shift tend to be more heterogeneous and varied, and our examples are naturally occurring.

\begin{figure*}[t]
\centering
\begin{subfigure}{.25\textwidth}
    \centering
    \includegraphics[width=0.98\textwidth]{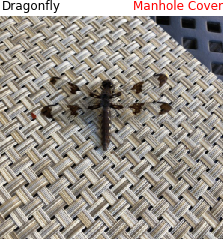}
\end{subfigure}%
\begin{subfigure}{.25\textwidth}
    \centering
    \includegraphics[width=0.98\textwidth]{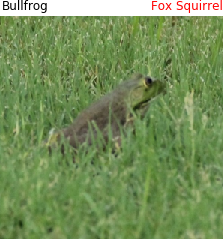}
\end{subfigure}%
\begin{subfigure}{.25\textwidth}
    \centering
    \includegraphics[width=0.98\textwidth]{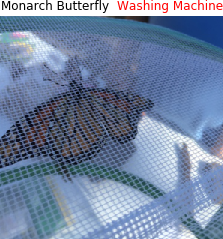}
\end{subfigure}%
\begin{subfigure}{.25\textwidth}
    \centering
    \includegraphics[width=0.98\textwidth]{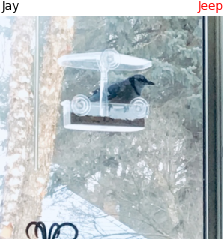}
\end{subfigure}
\caption{
Additional adversarially filtered examples from the \textsc{ImageNet-A} dataset. Examples are adversarially selected to cause classifier accuracy to degrade. The black text is the actual class, and the red text is a ResNet-50 prediction.
}\label{fig:imagenet-a}
\vspace{5pt}
\begin{subfigure}{.25\textwidth}
    \centering
    \includegraphics[width=0.98\textwidth]{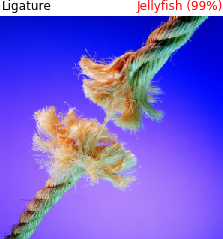}
\end{subfigure}%
\begin{subfigure}{.25\textwidth}
    \centering
    \includegraphics[width=0.98\textwidth]{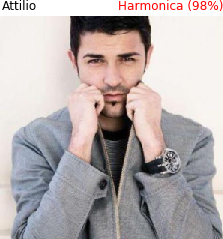}
\end{subfigure}%
\begin{subfigure}{.25\textwidth}
    \centering
    \includegraphics[width=0.98\textwidth]{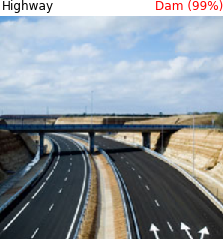}
\end{subfigure}%
\begin{subfigure}{.25\textwidth}
    \centering
    \includegraphics[width=0.98\textwidth]{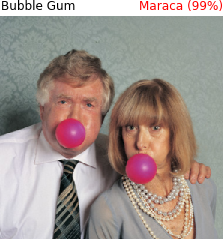}
\end{subfigure}
\caption{
Additional adversarially filtered examples from the \textsc{ImageNet-O} dataset. Examples are adversarially selected to cause out-of-distribution detection performance to degrade. Examples do not belong to ImageNet classes, and they are wrongly assigned highly confident predictions. The black text is the actual class, and the red text is a ResNet-50 prediction and the prediction confidence.
}\label{fig:imagenet-o}
\vspace{2pt}
\end{figure*}

\section{The Design and Construction of \textsc{ImageNet-A} and \textsc{ImageNet-O}}\label{sec:design}

\subsection{Design}
\textsc{ImageNet-A} is a dataset of adversarially filtered examples for ImageNet classifiers, or real-world examples that fool current classifiers. 
To find adversarially filtered examples, we first download numerous images related to an ImageNet class. Thereafter we delete the images that fixed ResNet-50 \parencite{resnet} classifiers correctly predict. We chose ResNet-50 due to its widespread use. Later we show that examples which fool ResNet-50 transfer reliably to other unseen models. With the remaining incorrectly classified images, we manually select a subset of high-quality images. 

Next, \textsc{ImageNet-O} is a dataset of adversarially filtered examples for ImageNet out-of-distribution detectors. To create this dataset, we download ImageNet-22K and delete examples from ImageNet-1K. With the remaining ImageNet-22K examples that do not belong to ImageNet-1K classes, we keep examples that are classified by a ResNet-50 as an ImageNet-1K class with high confidence. Then we manually select a subset of high-quality images.

Both datasets were manually labelled by graduate students over several months. This is because a large share of images in the ImageNet test set contain multiple classes per image \parencite{Stock2018ConvNetsAI}. Therefore, producing a high-quality dataset without multilabel images can be challenging with usual annotation techniques. By high-quality we refer to both being a singleton and recognizable instance of the target class. To ensure images do not fall into more than one of the several hundred classes, we had graduate students memorize the classes to build a high-quality test set.

\noindent\textbf{\textsc{ImageNet-A} Class Restrictions.}\quad 
We select a 200-class subset of ImageNet-1K's 1,000 classes so that errors among these 200 classes would be considered egregious \parencite{imagenet}.
For instance, wrongly classifying Norwich terriers as Norfolk terriers does less to demonstrate faults in current classifiers than mistaking a Persian cat for a candle.
We additionally avoid rare classes such as "snow leopard," classes that have changed much since 2012 such as "iPod," coarse classes such as "spiral," classes that are often image backdrops such as "valley," and finally classes that tend to overlap such as "honeycomb," "bee,'" "bee house," and "bee eater"; "eraser," "pencil sharpener" and "pencil case"; "sink," "medicine cabinet," "pill bottle" and "band-aid"; and so on. The 200 \textsc{ImageNet-A} classes cover most broad categories spanned by ImageNet-1K; see the Supplementary Materials \ref{app:imageneta-classes} for the full class list.

\noindent\textbf{\textsc{ImageNet-O} Class Restrictions.}\quad We again select a 200-class subset of ImageNet-1K's 1,000 classes. These 200 classes determine the in-distribution or the distribution that is considered usual. The remaining 800 classes could be used as data for Outlier Exposure \cite{hendrycks2019oe}.
As before, the 200 classes cover most broad categories spanned by ImageNet-1K; see the Supplementary Materials \ref{app:imageneto-classes} for the full class list. 

\subsection{Collection}
\noindent\textbf{\textsc{ImageNet-A} Data Aggregation.}\quad Curating a large set of adversarially filtered examples requires combing through an even larger set of images. Fortunately, the website iNaturalist has millions of user-labeled images of animals, and Flickr has even more user-tagged images of objects. We download images related to each of the 200 ImageNet classes by leveraging user-provided labels and tags. 
After exporting or scraping data from sites including iNaturalist, Flickr, and DuckDuckGo, we adversarially select images by removing examples that fail to fool our ResNet-50 models. Of the remaining images, we select low-confidence images and then ensure each image is valid through human review. For this procedure to work, many images are necessary; if we only used the original ImageNet test set as a source rather than iNaturalist, Flickr, and DuckDuckGo, some classes would have zero images after the first round of filtration.

For concreteness, we describe the selection process for the dragonfly class. We download 81,413 dragonfly images from iNaturalist, and after performing a basic filter with ResNet-50, we have 8,925 dragonfly images. In the algorithmically suggested shortlist, 1,452 images remain. From this shortlist, 80 dragonfly images are manually selected, but hundreds more could be chosen. Hence for just one class we may review over 1,000 images.

We now describe this process in more detail. 
We use a small ensemble of ResNet-50s for filtering, one pre-trained on ImageNet-1K then fine-tuned on the 200 class subset, and one pre-trained on ImageNet-1K where 200 of its 1,000 logits are used in classification. Both classifiers have similar accuracy on the 200 clean test set classes from ImageNet-1K. The ResNet-50s perform 10-crop classification of each image, and should any crop be classified correctly by the ResNet-50s, the image is removed. If either ResNet-50 assigns greater than 15\% confidence to the correct class, the image is also removed; this is done so that adversarially filtered examples yield misclassifications with low confidence in the correct class, like in untargeted adversarial attacks. Now, some classification confusions are greatly over-represented, such as Persian cat and lynx. We would like \textsc{ImageNet-A} to have great variability in its types of errors and cause classifiers to have a dense confusion matrix. Consequently, we perform a second round of filtering to create a shortlist where each confusion only appears at most 15 times. Finally, we manually select images from this shortlist in order to ensure \textsc{ImageNet-A} images are simultaneously valid, single-class, and high-quality. In all, the \textsc{ImageNet-A} dataset has 7,500 adversarially filtered examples.

\noindent\textbf{\textsc{ImageNet-O} Data Aggregation.}\quad Our dataset for adversarial out-of-distribution detection is created by fooling ResNet-50 out-of-distribution detectors. The negative of the prediction confidence of a ResNet-50 ImageNet classifier serves as our anomaly score \parencite{hendrycks2017baseline}. Usually in-distribution examples produce higher confidence predictions than OOD examples, but we curate OOD examples that have high confidence predictions. To gather candidate adversarially filtered examples, we use the ImageNet-22K dataset with ImageNet-1K classes deleted. We choose the ImageNet-22K dataset since it was collected in the same way as ImageNet-1K. ImageNet-22K allows us to have coverage of numerous visual concepts and vary the distribution's semantics without unnatural or unwanted non-semantic data shift. After excluding ImageNet-1K images, we process the remaining ImageNet-22K images and keep the images which cause the ResNet-50 to have high confidence, or a low anomaly score. We then manually select a high-quality subset of the remaining images to create \textsc{ImageNet-O}. We suggest only training models with data from the 1,000 ImageNet-1K classes, since the dataset becomes trivial if models train on ImageNet-22K. To our knowledge, this dataset is the first anomalous dataset curated for ImageNet models and enables researchers to study adversarial out-of-distribution detection. The \textsc{ImageNet-O} dataset has 2,000 adversarially filtered examples since anomalies are rarer; this has the same number of examples per class as ImageNetV2 \parencite{Recht2019DoIC}. Thus we use adversarial filtration to select examples that are difficult for a fixed ResNet-50, and we will show these examples straightforwardly transfer to unseen models. Additional example \textsc{ImageNet-O} images are in \ref{fig:imagenet-o}.\looseness=-1

For the collection of \textsc{ImageNet-O} we refrain from testing with classification models with a background class.  While there are many classification tasks which incorporate a ``background'' class into the final predictions such as many segmentation datasets \parencite{coco, Everingham2009ThePV_pascal}, most of the common multi-class classification datasets do not include such a category.  Most importantly the \textsc{ImageNet} dataset does not include a background class which is what our dataset aims to test models from.  There still remains the potential issue that if a model had been trained with a background class then the images from \textsc{ImageNet-O} would be classified as such.  It has been shown that depending on the score utilized such as from generative models out-of-distribution examples can actually achieve greater in-distribution (or lower anomaly) score in \cite{hendrycks2017baseline}.  We also show in a different chapter how poorly background classes perform in distinguishing out from in-distribution see \ref{tab:CAOS}.

\begin{figure*}[t]
\centering
\includegraphics[width=\textwidth]{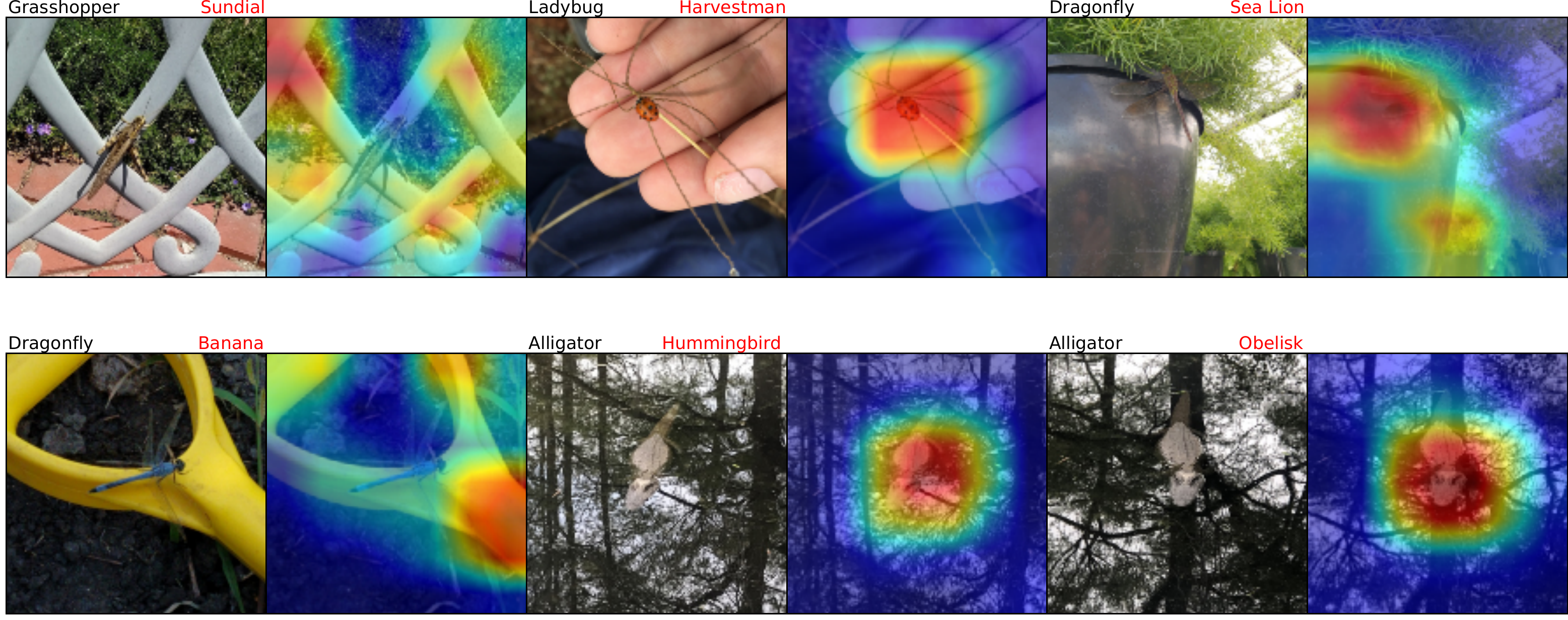}
\caption{
Examples from \textsc{ImageNet-A} demonstrating classifier failure modes.  Adjacent to each natural image is its heatmap generated by ``gradCAM'' \parencite{Selvaraju2019GradCAMVE}.  The heatmap is generated by taking the classification prediction and computing the gradients from the given output prediction back to the image.  The gradients are scaled to -1 to 1 range before adding back to the original image.  The -1 implies negative influence to the prediction and is depicted as blue, while 1 implies positive influence for the prediction and is the depicted as red.  The resulting image is rescaled after adding the image gradient.  Classifiers may use erroneous background cues for prediction and the technique gradCAM can be used. Further description of these failure modes is in Section \ref{sec:failures}.
}\label{fig:compilation}
\vspace{2pt}
\end{figure*}

\subsection{Illustrative Failure Modes}\label{sec:failures}

Examples in \textsc{ImageNet-A} uncover numerous failure modes of modern convolutional neural networks. We describe our findings after having viewed tens of thousands of candidate adversarially filtered examples. Some of these failure modes may also explain poor \textsc{ImageNet-O} performance, but for simplicity we describe our observations with \textsc{ImageNet-A} examples.

Observe figure \ref{fig:compilation}. The first two images suggest models may overgeneralize visual concepts. It may confuse metal with sundials, or thin radiating lines with harvestman bugs. We also observed that networks overgeneralize tricycles to bicycles and circles, digital clocks to keyboards and calculators, and more. We also observe that models may rely too heavily on color and texture, as shown with the dragonfly images. Since classifiers are taught to associate entire images with an object class, frequently appearing background elements may also become associated with a class, such as wood being associated with nails. Other examples include classifiers heavily associating hummingbird feeders with hummingbirds, leaf-covered tree branches being associated with the white-headed capuchin monkey class, snow being associated with shovels, and dumpsters with garbage trucks.
Additionally figure \ref{fig:compilation} shows an American alligator swimming. With different frames, the classifier prediction varies erratically between classes that are semantically loose and separate. For other images of the swimming alligator, classifiers predict that the alligator is a cliff, lynx, and a fox squirrel. Current convolutional networks have pervasive and diverse failure modes that are tested with \textsc{ImageNet-A}.




\section{Experiments}\label{sec:experiments}
We show that adversarially filtered examples collected to fool fixed ResNet-50 models reliably transfer to other models, indicating that current convolutional neural networks have shared weaknesses and failure modes. In the following sections, we analyze whether robustness can be improved by using data augmentation, using more real labeled data, and using different architectures. For the first two sections, we analyze performance with a fixed architecture for comparability, and in the final section we observe performance with different architectures. As a preliminary, we define our metrics.

\noindent\textbf{Metrics.} Our metric for assessing robustness to adversarially filtered examples for classifiers is the top-1 \emph{accuracy} on \textsc{ImageNet-A}. For reference, the top-1 accuracy on the 200 \textsc{ImageNet-A} classes using usual ImageNet images is usually greater than or equal to $90\%$ for ordinary classifiers.

Our metric for assessing out-of-distribution detection performance of \textsc{ImageNet-O} examples is the area under the precision-recall curve (\emph{AUPR}). This metric requires anomaly scores. Our anomaly score is the negative of the maximum softmax probabilities \parencite{hendrycks2017baseline} from a model that can classify the 200 \textsc{ImageNet-O} classes specified in Section \ref{sec:design}. We collect anomaly scores with the ImageNet validation examples for the said 200 classes.
Then, we collect anomaly scores for the \textsc{ImageNet-O} examples. Higher performing OOD detectors would assign \textsc{ImageNet-O} examples lower confidences, or higher anomaly scores. With these anomaly scores, we can compute the area under the precision-recall curve \parencite{auprbaseline}. Random chance levels for the AUPR is approximately $16.67\%$ with \textsc{ImageNet-O}, and the maximum AUPR is $100\%$.

\begin{figure}[t]
\centering
\vspace{-15pt}
\includegraphics[width=\textwidth]{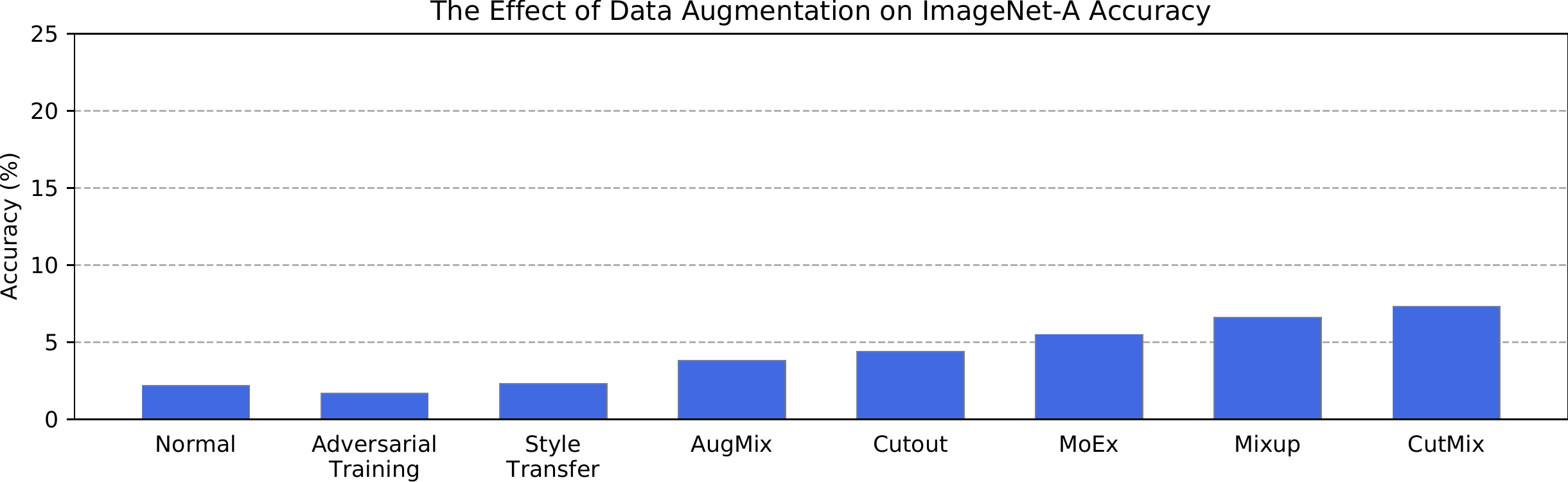}
\caption{Some data augmentation methods can slightly improve performance.}\label{fig:aug}
\vspace{-1pt}
\end{figure}

\paragraph{Data Augmentation.}
We examine popular data augmentation techniques and note their effect on robustness. In this section we exclude \textsc{ImageNet-O} results, as the data augmentation techniques hardly help with out-of-distribution detection as well.
As a baseline, we train a new ResNet-50 from scratch and obtain $2.17\%$ accuracy on \textsc{ImageNet-A}.
Now, one purported way to increase robustness is through adversarial training, which makes models less sensitive to $\ell_p$ perturbations. We use the adversarially trained model from \parencite{wong2020fast}, but accuracy decreases to $1.68\%$. Next, \cite{geirhos2019} propose making networks rely less on texture by training classifiers on images where textures are transferred from art pieces. They accomplish this by applying style transfer to ImageNet training images to create a stylized dataset, and models train on these images. While this technique is able to greatly increase robustness on synthetic corruptions \parencite{hendrycks2019robustness}, Style Transfer increases \textsc{ImageNet-A} accuracy by $0.13\%$ over the ResNet-50 baseline. A recent data augmentation technique, AugMix \parencite{hendrycks2019augmix}, which takes linear combinations of different data augmentations increases accuracy to $3.8\%$. Cutout augmentation \parencite{Devries2017ImprovedRO} randomly occludes image regions and corresponds to $4.4\%$ accuracy. Moment Exchange (MoEx) \parencite{Li2020OnFN} exchanges feature map moments between images, and this increases accuracy to $5.5\%$. Mixup \parencite{zhang2017mixup} trains networks on elementwise convex combinations of images and their interpolated labels; this technique increases accuracy to $6.6\%$. CutMix \parencite{Yun2019CutMixRS} superimposes images regions within other images and yields $7.3\%$ accuracy. At best these data augmentations techniques improve accuracy by approximately $5\%$ over the baseline. Results are summarized in figure \ref{fig:aug}.

\paragraph{More Labeled Data.}
One possible explanation for consistently low \textsc{ImageNet-A} accuracy is that all models are trained only with ImageNet-1K, and using additional data may resolve the problem.
To test this hypothesis we pre-train a ResNet-50 on Places365 \parencite{zhou2017places}, a large-scale scene recognition dataset. After fine-tuning the Places365 model on ImageNet-1K, we find that accuracy is $1.56\%$. Consequently, even though scene recognition models are purported to have qualitatively distinct features \parencite{Zhou2019InterpretingDV}, this is not enough to improve \textsc{ImageNet-A} performance. Likewise, Places365 pre-training does not improve \textsc{ImageNet-O} detection, as its AUPR is $14.88\%$. Next, we see whether labeled data from \textsc{ImageNet-A} itself can help. We take baseline ResNet-50 with $2.17\%$ \textsc{ImageNet-A} accuracy and fine-tune it on $80\%$ of \textsc{ImageNet-A}. This leads to no clear improvement on the remaining $20\%$ of \textsc{ImageNet-A} since the top1 and top5 accuracies are below $2\%$ and $5\%$ respectively. Last, we pre-train using an order of magnitude more training data with ImageNet-21K. This dataset contains approximately 21,000 classes and approximately 14 million images. To our knowledge this is the largest publicly available database of labeled natural images. Using a ResNet-50 pretrained on ImageNet-21K, we fine-tune the model on ImageNet-1K and attain $11.41\% $accuracy on \textsc{ImageNet-A}, a $9.24\%$ increase. Likewise, the AUPR for \textsc{ImageNet-O} improves from $16.20\%$ to $21.86\%$, although this improvement is less significant since \textsc{ImageNet-O} images overlap with ImageNet-21K images. Overall, an order of magnitude increase in labeled training data can provide some improvements in accuracy.

\begin{figure}[t]
\begin{subfigure}{0.5\textwidth}
    \centering
    \includegraphics[width=0.99\textwidth]{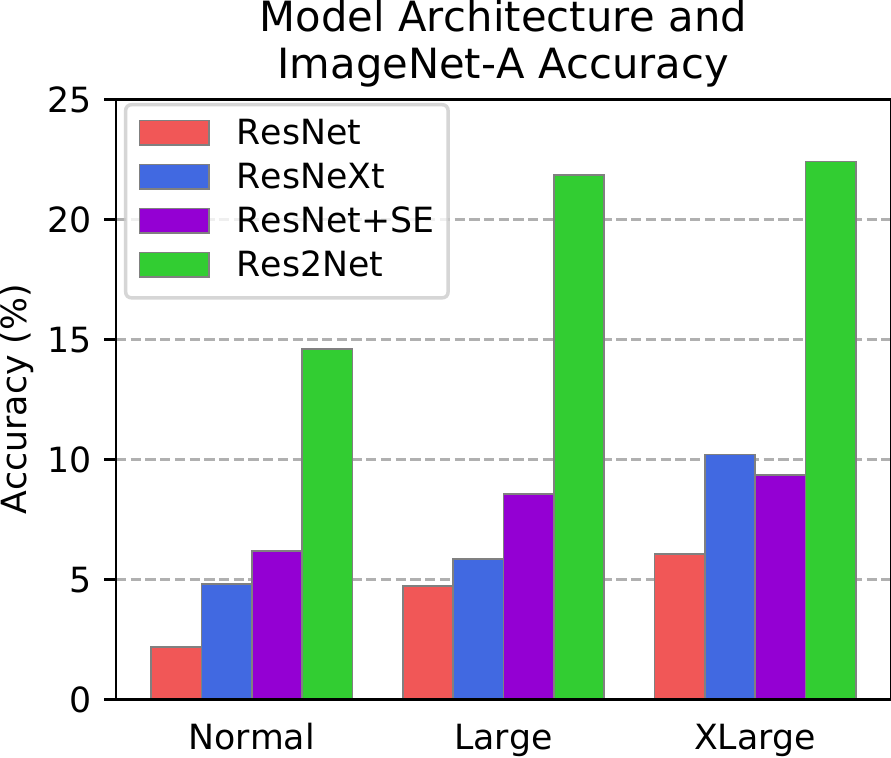}
\end{subfigure}%
\begin{subfigure}{0.5\textwidth}
    \centering
    \includegraphics[width=0.99\textwidth]{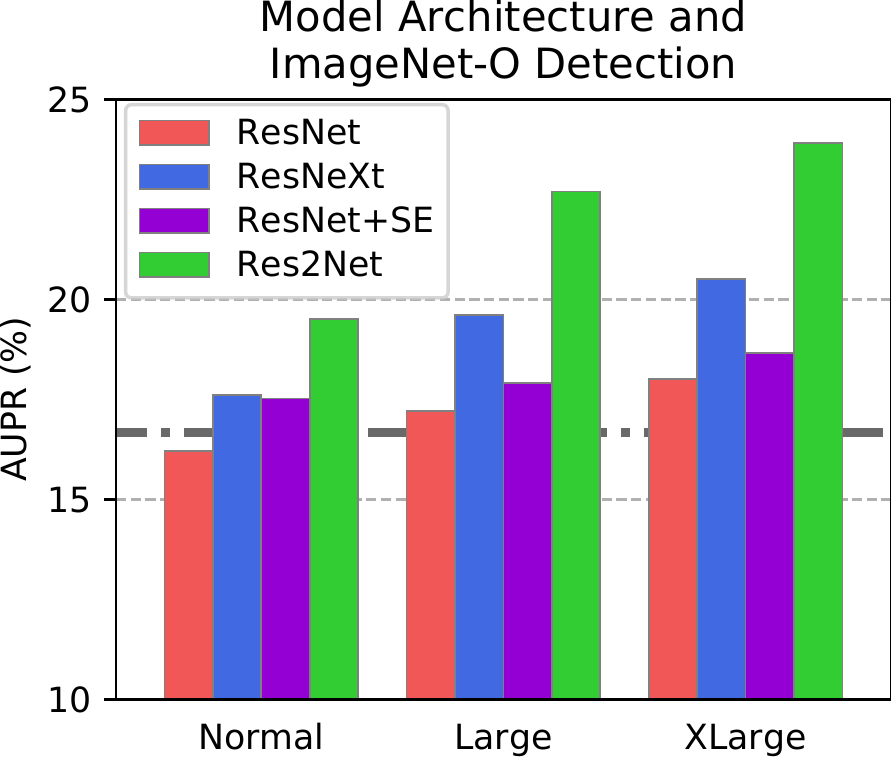}
\end{subfigure}%
\caption{Increasing model size and other architecture changes can greatly improve performance. Note Res2Net and ResNet+SE have a ResNet backbone. Normal model sizes are ResNet-50 and ResNeXt-50 ($32\times4$d), Large model size are ResNet-101 and ResNeXt-101 ($32\times4$d), and XLarge Model sizes are ResNet-152 and ($32\times8$d).}\label{fig:attn}
\vspace{2pt}
\end{figure}

\paragraph{Architectural Changes.}
We find that model architecture can play a large role in \textsc{ImageNet-A} accuracy and \textsc{ImageNet-O} detection performance.
Simply increasing the width and number of layers of a network is sufficient to automatically impart more \textsc{ImageNet-A} accuracy and \textsc{ImageNet-O} OOD detection performance. Increasing network capacity has been shown to improve performance on $\ell_p$ adversarial examples \parencite{kurakin}, common corruptions \parencite{hendrycks2019robustness}, and now also improves performance for adversarially filtered images. For example, a ResNet-50's top-1 accuracy and AUPR is $2.17\%$ and $16.2\%$, respectively, while a ResNet-152 obtains $6.1\%$ top-1 accuracy and $18.0\%$ AUPR. Another architecture change that reliably helps is using the grouped convolutions found in ResNeXts \parencite{resnext}. A ResNeXt-50 ($32\times4$d) obtains a $4.81\%$ top1 \textsc{ImageNet-A} accuracy and a $17.60\%$ \textsc{ImageNet-O} AUPR.
Another architectural change is self attention.
Convolutional neural networks with self-attention \parencite{Hu2018GatherExciteE} are designed to better capture long-range dependencies and interactions across an image. We consider the self-attention technique called Squeeze-and-Excitation (SE) \parencite{Hu2018SqueezeandExcitationN}, which won the final ImageNet competition in 2017. A ResNet-50 with Squeeze-and-Excitation attains $6.17\%$ accuracy. However, for larger ResNets, self-attention does little to improve \textsc{ImageNet-O} detection.
Finally, we consider the ResNet-50 architecture with its residual blocks exchanged with recently introduced Res2Net v1b blocks \parencite{Gao2019Res2NetAN}. This change increases accuracy to $14.59\%$ and the AUPR to $19.5\%$. A ResNet-152 with Res2Net v1b blocks attains $22.4\%$ accuracy and $23.9\%$ AUPR. Compared to data augmentation or an order of magnitude more labeled training data, some architectural changes can provide far more robustness gains. Consequently future improvements to model architectures is a promising path towards greater robustness.

\section{Conclusion}
In this chapter, we introduced adversarially filtered examples for image classifiers and out-of-distribution detectors. Our \textsc{ImageNet-A} dataset degrades classification accuracy across known classifiers, and it measures robustness to input data distribution shifts. Likewise, \textsc{ImageNet-O} adversarially filtered examples reliably degrade ImageNet out-of-distribution detection performance, and it measures robustness to label distribution shifts. \textsc{ImageNet-O} enables the measurement of adversarial out-of-distribution detection performance, and is the first ImageNet out-of-distribution detection dataset. Our adversarial filtration process removes examples solved by simple spurious cues, so our datasets enable researchers to observe performance when simple spurious cues are removed. Our naturally occurring images expose common blindspots of current convolutional networks, and solving these tasks will require addressing long-standing but under-explored failure modes of current models such as over-reliance on texture, over-generalization, and spurious cues. We found that these failures are slightly less pronounced with different data augmentation strategies. However, we identified that architectural improvements can provide large gains in model robustness, and there is much room for future research.
In this work, we introduce two new and difficult ImageNet test sets to measure model performance under distribution shift---an important research aim as computer vision systems are deployed in increasingly precarious real-world environments.\looseness=-1


\chapter{Scaling Anomaly Detection to Large Scale Images}\label{ch:anom_segmentation}

\section{Overview}

\begin{figure*}  
    \centering
    \includegraphics[width=\textwidth]{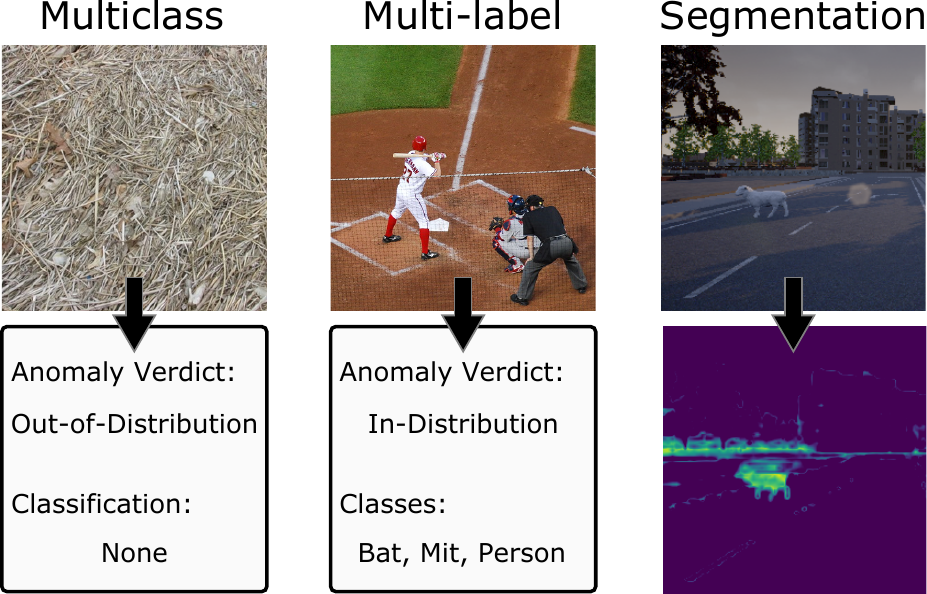}
    \caption{We scale up out-of-distribution detection to large-scale multi-class datasets with hundreds of classes, multi-label datasets with complex scenes, and anomaly segmentation in driving environments. In all three settings, we find that an OOD detector based on the maximum logit outperforms previous methods, establishing a strong and versatile baseline for future work on large-scale OOD detection.}\vspace{-10pt}
    \label{fig:StreetHazards1}
\vspace{6pt}
\end{figure*} 

Detecting out-of-distribution inputs is important in real-world applications of deep learning. When faced with anomalous inputs flagged as such, systems may initiate a conservative fallback policy or defer to human judgment. This is especially important in safety-critical applications of deep learning, such as self-driving cars or medical applications. Accordingly, research on out-of-distribution detection has a rich history spanning several decades \parencite{OC-SVM,  lof, emmott2015metaanalysis}. Recent work leverages deep neural representations for out-of-distribution detection in complex domains, such as image data \parencite{hendrycks2017baseline, kimin, hendrycks2019oe}. However, these works still primarily use small-scale datasets with low-resolution images and few classes. As the community moves towards more realistic, large-scale settings, strong baselines and high-quality benchmarks are imperative for future progress.

In addition to focusing on small-scale datasets, previous formulations of anomaly detection treat entire images as anomalies. In practice, an image could be anomalous in localized regions while being in-distribution elsewhere. Knowing which regions of an image are anomalous could allow for safer handling of unfamiliar objects in the case of self-driving cars. Creating a benchmark for this task is difficult, though, as simply cutting and pasting anomalous objects into images introduces various unnatural giveaway cues such as edge effects, mismatched orientation, and lighting, all of which trivialize the task of anomaly segmentation \parencite{blum2019fishyscapes}.

To overcome these issues, we utilize a simulated driving environment to create the novel StreetHazards dataset for anomaly segmentation. Using the Unreal Engine and the open-source CARLA simulation environment \parencite{carla}, we insert a diverse array of foreign objects into driving scenes and re-render the scenes with these novel objects. This enables integration of the foreign objects into their surrounding context with correct lighting and orientation.

To complement the StreetHazards dataset, we convert the BDD100K semantic segmentation dataset \parencite{bdd100k} into an anomaly segmentation dataset, which we call BDD-Anomaly. By leveraging the large scale of BDD100K, we reserve infrequent object classes to be anomalies. We combine this dataset with StreetHazards to form the Combined Anomalous Object Segmentation (CAOS) benchmark. The CAOS benchmark improves over previous evaluations for anomaly segmentation in driving scenes by evaluating detectors on realistic and diverse anomalies. We evaluate several baselines on the CAOS benchmark and discuss problems with porting existing approaches from earlier formulations of out-of-distribution detection.

In more traditional whole-image anomaly detection, large-scale datasets such as ImageNet \parencite{imagenet_cvpr09} and Places365 \parencite{zhou2017places} present unique challenges not seen in small-scale settings, such as a plethora of fine-grained object classes. We demonstrate that the MSP detector, a state-of-the-art method for small-scale problems, does not scale well to these challenging conditions. Moreover, in the common real-world case of multi-label data, the MSP detector cannot naturally be applied in the first place, as it requires softmax probabilities. 

Through extensive experiments, we identify a detector based on the maximum logit (MaxLogit) that greatly outperforms strong baselines in large-scale multi-class, and anomaly segmentation settings. In each of these three settings, we discuss why MaxLogit provides superior performance, and we show that these gains are hidden if one looks at small-scale problems alone. The code for our experiments and the CAOS benchmark datasets are available at \href{https://github.com/hendrycks/anomaly-seg}{\text{github.com/hendrycks/anomaly-seg}}.

\section{Related Work}

\noindent\textbf{Anomaly Segmentation.}\quad
Several prior works explore segmenting anomalous image regions. One line of work uses the WildDash dataset \parencite{zendel2018wilddash}, which contains numerous annotated driving scenes in conditions such as snow, fog, and rain. The WildDash test set contains fifteen "negative image" from different domains for which the goal is to mark the entire image as out-of-distribution. Thus, while the task is segmentation, the anomalies do not exist as objects within an otherwise in-distribution scene. This setting is similar to that explored by \parencite{hendrycks2017baseline}, in which whole images from other datasets serve as out-of-distribution examples.

To approach anomaly segmentation on WildDash, \parencite{kreo2018robust} train on multiple semantic segmentation domains and treat regions of images from the WildDash driving dataset as out-of-distribution if they are segmented as regions from different domains, i.e. indoor classes. \parencite{bev2018discriminative} use ILSVRC 2012 images and train their network to segment the entirety of these images as out-of-distribution.

In medical anomaly segmentation and product fault detection, anomalies are regions of otherwise in-distribution images. \parencite{Baur_2019} segment anomalous regions in brain MRIs using pixel-wise reconstruction loss. Similarly, \parencite{haselmann2018anomaly} perform product fault detection using pixel-wise reconstruction loss and introduce an expansive dataset for segmentation of product faults. In these relatively simple domains, reconstruction-based approaches work well. In contrast to medical anomaly segmentation and fault detection, we consider complex images from street scenes. These images have high variability in scene layout and lighting, and hence are less amenable to reconstruction-based techniques.

\begin{table*}
	\centering
	\begin{tabularx}{1\textwidth}{*{1}{>{\hsize=1\hsize}X} | *{1}{>{\hsize=1\hsize}Y} *{1}{>{\hsize=1\hsize}Y} *{1}{>{\hsize=1\hsize}Y} *{1}{>{\hsize=1\hsize}Y}}
		& Fishyscapes & Lost and Found & BDD-Anomaly (Ours) & StreetHazards (Ours) \\ \hline
		Train Images & 0 & 1036 & 6280 & 5125 \\
		Test Images & 1000 & 1068 & 810 & 1500 \\
		Anomaly Types & 12 & 9 & $3$ & 250 \\
		\Xhline{3\arrayrulewidth}
	\end{tabularx}
	\caption{Quantitative comparison of the CAOS benchmark with related datasets. The BDD-Anomaly dataset treats three categories as anomalous and has many unique object instances within those categories. By contrast, Lost and Found has the same objects in multiple images and has only nine unseen objects at test time. StreetHazards leverages a simulated environment to naturally insert hundreds of varied anomalies.}
	\label{tab:dataset_comparison}
\end{table*}

The two works closest to our own are the Lost and Found \parencite{Pinggera2016LostAF} and Fishyscapes \parencite{blum2019fishyscapes} datasets. In table \ref{tab:dataset_comparison}, we quantitatively compare the CAOS benchmark to these datasets. The Lost and Found dataset consists of real images in a driving environment with small road hazards. The images were collected to mirror the Cityscapes dataset \parencite{Cordts2016Cityscapes} but are only collected from one city and so have less diversity. The dataset contains 35 unique anomalous objects, and methods are allowed to train on many of these. For Lost and Found, only nine unique objects are truly unseen at test time. Crucially, this is a different evaluation setting from our own, where anomalous objects are not revealed at training time, so their dataset is not directly comparable. Nevertheless, the BDD-Anomaly dataset fills several gaps in Lost and Found. First, the images are more diverse, because they are sourced from a more recent and comprehensive semantic segmentation dataset. Second, the anomalies are not restricted to small, sparse road hazards. Concretely, anomalous regions in Lost and Found take up 0.11\% of the image on average, whereas anomalous regions in the BDD-Anomaly dataset are larger and fill 0.83\% of the image on average. Finally, although the BDD-Anomaly dataset treats three categories as anomalous, compared to Lost and Found it has far more unique anomalous objects.

The Fishyscapes benchmark for anomaly segmentation consists of cut-and-paste anomalies from out-of-distribution domains. This is problematic, because the anomalies stand out as clearly unnatural in context. For instance, the orientation of anomalous objects is unnatural, and the lighting of the cut-and-paste patch differs from the lighting in the original image, providing an unnatural cue to anomaly detectors that would not exist for real anomalies. Techniques for detecting image manipulation \parencite{zhou2018learning} are competent at detecting artificial image elements of this kind. Our StreetHazards dataset overcomes these issues by leveraging a simulated driving environment to naturally insert anomalous \textit{3D models} into a scene rather than overlaying 2D images. These anomalies are integrated into the scene with proper lighting and orientation, mimicking real-world anomalies and making them significantly more difficult to detect.

\begin{figure}
\centering
\includegraphics{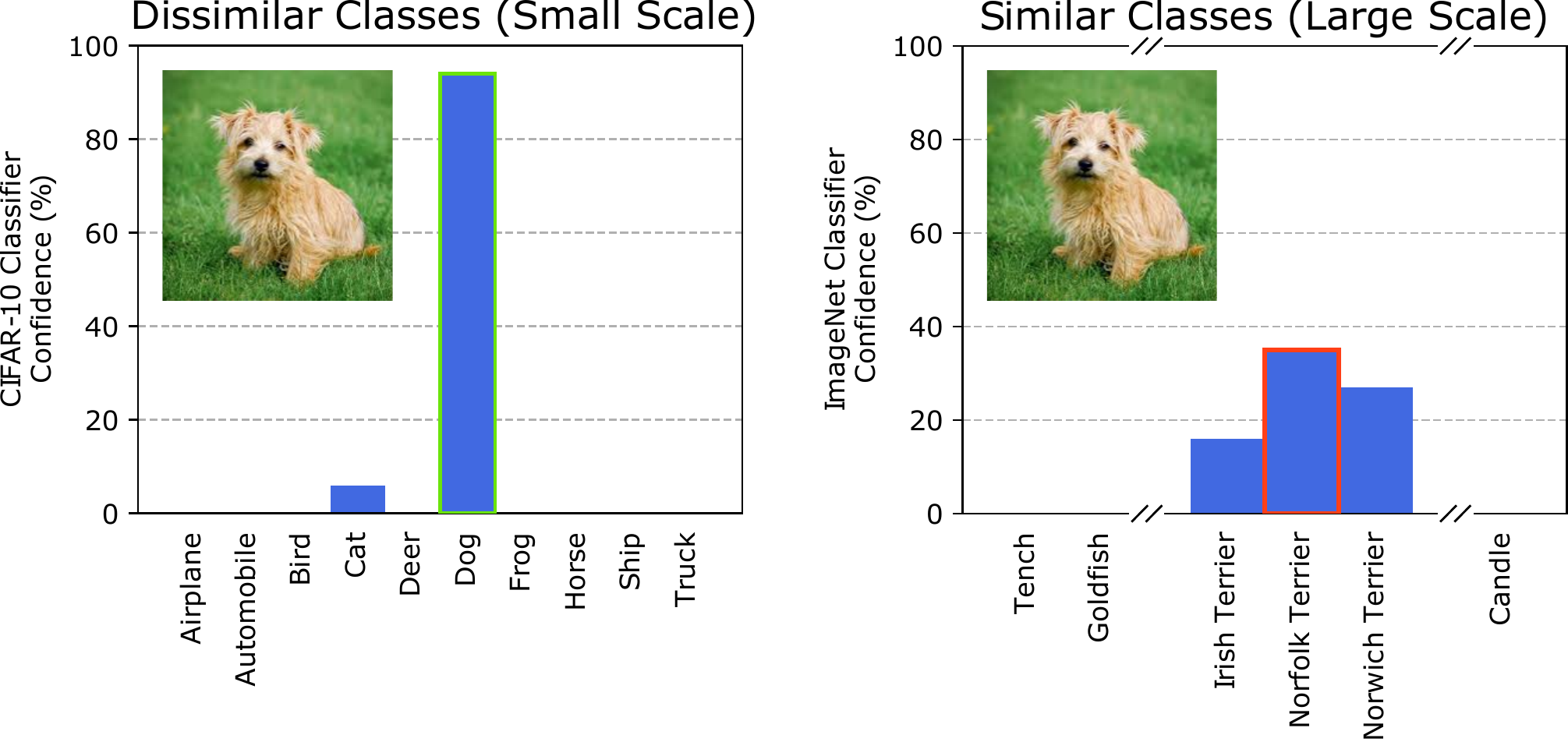}
\caption{Small-scale datasets such as CIFAR-10 have relatively disjoint classes, but larger-scale datasets including ImageNet have several classes with high visual similarity to other classes. The implication is that large-scale classifiers disperse probability mass among several classes. If the prediction confidence is used for out-of-distribution detection, then images which have similarities to other classes will often wrongly be deemed out-of-distribution due to dispersed confidence. The dog is lower resolution for the CIFAR-10 classifier.}
\label{fig:smaxvkl}
\end{figure}

\noindent\textbf{Multi-Class Out-of-Distribution Detection.}\quad
A recent line of work leverages deep neural representations from multi-class classifiers to perform out-of-distribution (OOD) detection on high-dimensional data, including images, text, and speech data. \cite{hendrycks2017baseline} formulate the task and propose the simple baseline of using the maximum softmax probability of the classifier on an input to gauge whether the input is out-of-distribution. In particular, they formulate the task
as distinguishing between examples from an in-distribution dataset and various out-of-distribution datasets. Importantly, entire images are treated as out-of-distribution.

Continuing this line of work, \cite{kimin} propose to improve the neural representation of the classifier to better separate in- and out-of-distribution examples. They use generative adversarial networks to produce near-distribution examples. In training, they encourage the classifier to output a uniform posterior on these synthetic out-of-distribution examples. \cite{hendrycks2019oe} observe that outliers are often easy to obtain in large quantity from diverse, realistic datasets and demonstrate that training out-of-distribution detectors to detect these outliers generalizes to completely new, unseen classes of anomalies.
Other work investigates improving the anomaly detectors themselves given a fixed classifier \parencite{devries, Liang2018ODIN}. However, as observed in \cite{hendrycks2019oe}, most of these works tune hyperparameters on a particular type of anomaly that is also seen at test time, so their evaluation setting is more lenient. We ensure that all anomalies seen at test time come from entirely unseen categories and are not tuned on in any way, hence we do not compare to techniques such as \cite{Liang2018ODIN}. Additionally, in a point of departure from prior work, we focus primarily on large-scale images and datasets with many classes.




\section{Multi-Class Prediction for OOD Detection}\label{anom_seg:sec:multiclass}


\noindent\textbf{Problem with existing baselines.}\quad In large-scale image classification, a network is often tasked with predicting an object's identity from one of hundreds or thousands of classes, where class distinctions tend to be more fine and subtle. An increase in similarity and overlap between classes spells a problem for the multi-class out-of-distribution baseline \parencite{hendrycks2017baseline}. This baseline uses the negative maximum softmax probability as the anomaly score, or $-\max_k p(y=k \mid x)$. Classifiers tend to have higher confidence on in-distribution examples than out-of-distribution examples, enabling OOD detection. Assuming single-model evaluation and no access to other anomalies or test-time adaptation, the maximum softmax probability (MSP) is the state-of-the-art multi-class out-of-distribution detection method. However, we show that the MSP is problematic for large-scale datasets with many classes including ImageNet-1K and Places365 \parencite{zhou2017places}. Probability mass can be dispersed among visually similar classes, as shown in figure \ref{fig:smaxvkl}. Consequently, a classifier may produce a low confidence prediction for an in-distribution image, not because the image is unfamiliar or out-of-distribution, but because the object's exact class is difficult to determine. To circumvent this problem, we propose using the negative of the maximum unnormalized logit for an anomaly score, which we call MaxLogit. 

The MaxLogit has several benefits over the previous baseline of MSP.  
Empirically we find that MaxLogit outperforms the MSP, although the difference is marginal in small scale image datasets such as CIFAR-10 but show a larger improvement on CAOS benchmark and ImageNet see \ref{tab:multiclass}.  The two main reasons other than emperical performance to switch from MSP to MaxLogit are that the MaxLogit can work even if it is not a distributions and MaxLogit is unaffected by the number of classes.  To expand upon each point, for several image tasks the output comes in the form of multi-label categories which the results are not from a softmax and not a distribution as the results do not sum to one, and MaxLogit does not require such a condition to in order to operate.  The other point is a bit more subtle but none-the-less important.  While the softmax operation does not change the maximum score, and does not change the ordering of classes per item, it does affect the relative ordering across items.  This is because the softmax operation will convert the absolute differences per item in relative differences which loses information when comparing across items.  


\noindent\textbf{Datasets.}\quad To evaluate the MSP baseline out-of-distribution detector and the MaxLogit detector, we use ImageNet-1K object recognition dataset and Places365 scene recognition dataset as in-distribution datasets $\mathcal{D}_\text{in}$. We use several out-of-distribution test datasets $\mathcal{D}_\text{out}$, all of which are unseen during training.
The first out-of-distribution dataset is \emph{Gaussian} noise, where each pixel of these out-of-distribution examples are i.i.d.\ sampled from $\mathcal{N}(0,0.5)$ and clipped to be contained within $[-1,1]$. Another type of test-time noise is \emph{Rademacher} noise, in which each pixel is i.i.d.\ sampled from $2\cdot\text{Bernoulli}(0.5) - 1$, i.e.~ each pixel is $1$ or $-1$ with equal probability. \emph{Blob} examples are more structured than noise; they are algorithmically generated blob images. Meanwhile, \emph{Textures} is a dataset consisting in images of describable textures \parencite{textures}. When evaluating the ImageNet-1K detector, we use \emph{LSUN} images, which is a dataset for scene recognition \parencite{lsun}. Our final $\mathcal{D}_\text{out}$ is \emph{Places69}, a scene classification dataset that does not share classes with Places365. In all, we evaluate against out-of-distribution examples spanning synthetic and realistic images.

\begin{table*}
 \setlength{\tabcolsep}{8pt}
 \centering
 \begin{tabularx}{\textwidth}{*{1}{>{\hsize=0.65\hsize}X}
 | *{3}{>{\hsize=0.4\hsize}Y}
 |*{3}{>{\hsize=0.4\hsize}Y}
 | *{3}{>{\hsize=0.4\hsize}Y} }
 & \multicolumn{3}{c}{FPR95 $\downarrow$} & \multicolumn{3}{c}{AUROC $\uparrow$} &\multicolumn{3}{c}{AUPR  $\uparrow$}\\ \cline{2-10}
 $\mathcal{D}_\text{in}$ &
 {MSP} & {MaxLogit} & {KL} & {MSP} & {MaxLogit} & {KL} & {MSP} & {MaxLogit} & {KL} \\ \hline
 ImageNet & {42.42} & {{35.77}} & {36.22} & {84.60} & {{87.20}} & {87.29} & {48.26} & {{45.68}} & {37.32} \\
 Places365 & {52.68} & {{36.6}} & {49.14} & {75.67} & {{85.9}} & {80.01} & {8.13} & {{19.2}} & {24.61} \\
 \Xhline{3\arrayrulewidth}
 \end{tabularx}
 \caption{Multi-class out-of-distribution detection results using the maximum softmax probability, maximum logit basline and KL Divergence between predicted and posterior. Results are on ImageNet and Places365. Values are rounded so that $99.995\%$ rounds to $100\%$. Full results on individual $\mathcal{D}_\text{out}$ datasets and additional baselines are in the supplementary material.}
 \label{tab:multiclass}
\end{table*}

\noindent\textbf{Results.}\quad Results are shown in table \ref{tab:multiclass}. Observe that the proposed MaxLogit method outperforms the maximum softmax probability baseline for all three metrics on both ImageNet and Places365. These results were computed using a ResNet-50 trained on either ImageNet-1K or Places365. In the case of Places365, the AUROC improvement is over 10\%. We note that the utility of the maximum logit could not as easily be appreciated in previous work's small-scale settings. For example, using the small-scale CIFAR-10 setup of \parencite{hendrycks2019oe}, the MSP attains an average AUROC of 90.08\% while the maximum logit attains 90.22\%, a 0.14\% difference. However, in a large-scale setting, the difference can be over 10\% on individual $\mathcal{D}_\text{out}$ datasets. We are not claiming that utilizing the maximum logit is a mathematically innovative formulation, only that it serves as a consistently powerful baseline for large-scale settings that went unappreciated in small-scale settings. In consequence, we suggest using the maximum logit as a new baseline for large-scale multi-class out-of-distribution detection.

\section{The CAOS Benchmark}
The Combined Anomalous Object Segmentation (CAOS) benchmark is comprised of two complementary datasets for evaluating anomaly segmentation systems on diverse, realistic anomalies.  First is the StreetHazards dataset, which leverages simulation to provide a large variety of anomalous objects realistically inserted into driving scenes. Second is the BDD-Anomaly dataset, which consists of real images taken from the BDD100K dataset \parencite{bdd100k}. StreetHazards contains a highly diverse array of anomalies; BDD-Anomaly contains anomalies in real-world images. Together, these datasets allow researchers to judge techniques on their ability to segment diverse anomalies as well as anomalies in real images. All images have $720\times 1280$ resolution, and we recommend evaluating with the AUROC, AUPR, and FPR$K$ metrics, which we describe in Section \ref{section:experiments}.

\noindent\textbf{The StreetHazards Dataset.}\quad
StreetHazards is an anomaly segmentation dataset that leverages simulation to provide diverse, realistically-inserted anomalous objects. To create the StreetHazards dataset, we use the Unreal Engine along with the CARLA simulation environment \parencite{carla}. From several months of development and testing including customization of the Unreal Engine and CARLA, we can insert foreign entities into a scene while having them be properly integrated. Unlike previous work, this avoids the issues of inconsistent chromatic aberration, edge effects, differences in environmental lighting, and other simple cues that an object is anomalous. Additionally, using a simulated environment allows us to dynamically insert diverse anomalous objects in any location and have them render properly with changes to lighting and weather including time of day, cloudy skies, and rain.

\begin{figure}
    \centering
    \includegraphics[width=0.45\textwidth]{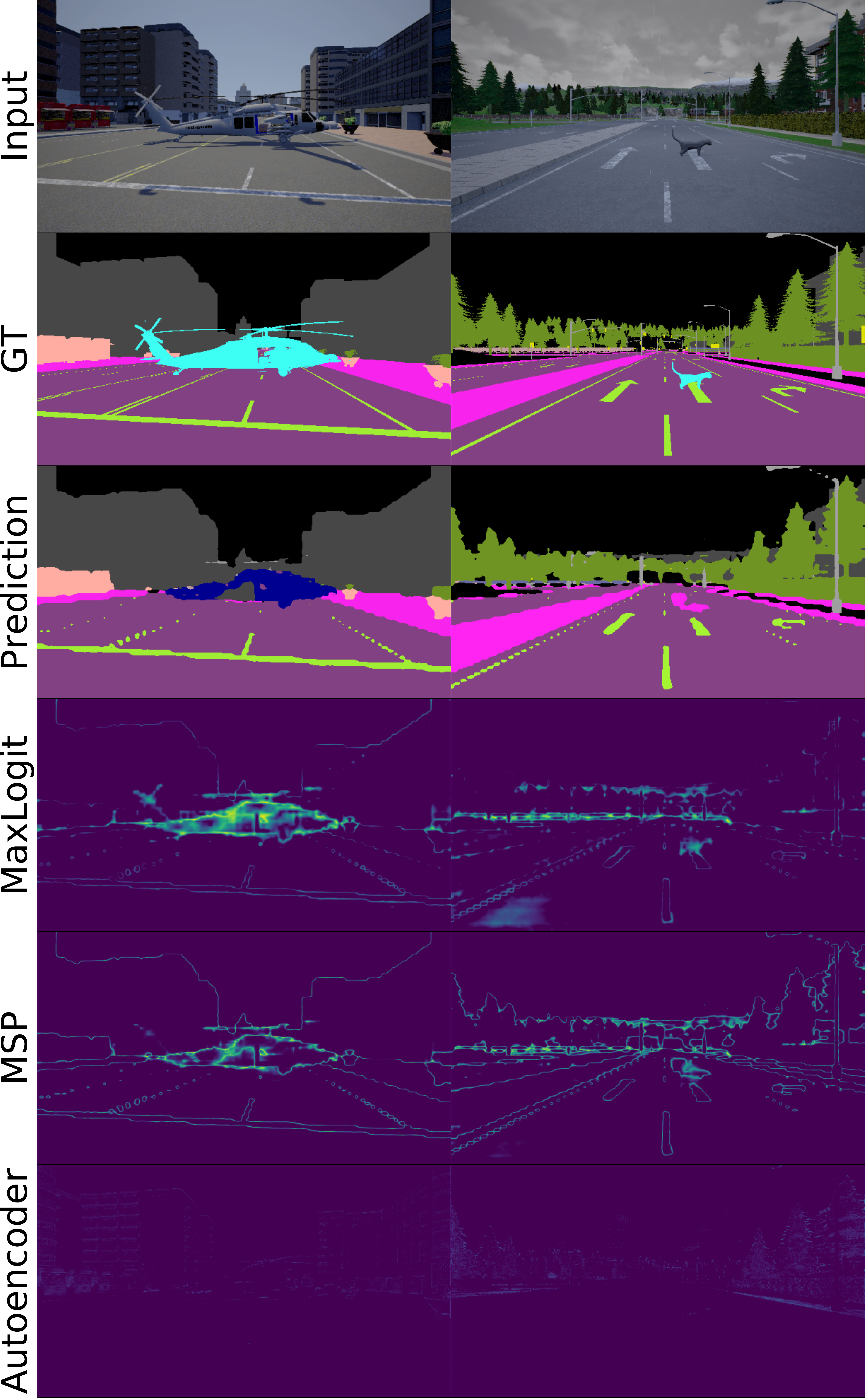}
    \caption{A sample of anomalous scenes, model predictions, and anomaly scores. The anomaly scores are thresholded to the top 10\% of values for visualization. GT is ground truth, the autoencoder model is based on the spatial autoencoder used in \parencite{Baur_2019}, MSP is the maximum softmax probability baseline \parencite{hendrycks2017baseline}, and MaxLogit is the method we propose as a new baseline for large-scale settings.}
    \label{fig:StreetHazards2}
\end{figure}

We use 3 towns from CARLA for training, from which we collect RGB images and their respective semantic segmentation maps to serve as our training data for our semantic segmentation model. We generate a validation set from the fourth town. Finally, we reserve the fifth and sixth town as our test set. We insert anomalies taken from the Digimation Model Bank Library and semantic ShapeNet (ShapeNetSem) \parencite{shapenet} into the test set in order to evaluate methods for out-of-distribution detection. In total, we use 250 unique anomaly models of diverse types. There are 12 classes used for training: background, road, street lines, traffic signs, sidewalk, pedestrian, vehicle, building, wall, pole, fence, and vegetation. The thirteenth class is the anomaly class that is only used at test time. We collect 5,125 image and semantic segmentation ground truth pairs for training, 1,031 pairs without anomalies for validation, and 1,500 test pairs with anomalies.

\noindent\textbf{The BDD-Anomaly Dataset.}\quad
BDD-Anomaly is an anomaly segmentation dataset with real images in diverse conditions. We source BDD-Anomaly from BDD100K \parencite{bdd100k}, a large-scale semantic segmentation dataset with diverse driving conditions. The original data consists in 7000 images for training and 1000 for validation. There are 18 original classes. We choose \textit{motorcycle}, \textit{train}, and \textit{bicycle} as the anomalous object classes and remove all images with these objects from the training and validation sets. This yields 6,280 training pairs, 910 validation pairs without anomalies, and 810 testing pairs with anomalous objects.

\subsection{Experiments}\label{section:experiments}
\noindent\textbf{Metrics.}\quad 
To evaluate out-of-distribution detectors in large-scale settings, we use three standard metrics of detection performance: area under the ROC curve (AUROC), false positive rate at 95\% recall (FPR95), and area under the precision-recall curve (AUPR). The AUROC and AUPR are important metrics, because they give a holistic measure of performance when the cutoff for detecting anomalies is not a priori obvious or when we want to represent the performance of a detection method across several different cutoffs.\looseness=-1

The AUROC can be thought of as the probability that an anomalous example is given a higher score than an ordinary example. Thus, a higher score is better, and an uninformative detector has a AUROC of 50\%. AUPR provides a metric more attuned to class imbalances, which is relevant in anomaly and failure detection, when the number of anomalies or failures may be relatively small. Last, the FPR95 metric consists of measuring the false positive rate at 95\%. This is important because it tells us how many false positives (i.e.~false alarms) are necessary for a given method to achieve a desired recall. This desired recall may be thought of as a safety threshold. Moreover, anomalies and system failures may require human intervention, so a detector requiring little human intervention while still detecting most anomalies is of pecuniary importance.

In anomaly segmentation experiments, each pixel is treated as a prediction, resulting in many predictions to evaluate. To fit these in memory, we compute the metrics on each image and average over the images to obtain final values.

\noindent\textbf{Methods.}\quad
Our first baseline is pixel-wise Maximum Softmax Probability (MSP). Introduced in \cite{hendrycks2017baseline} for multi-class out-of-distribution detection, we directly port this baseline to anomaly segmentation. 
Alternatively, the background class might serve as an anomaly detector, because it contains everything not in the other classes. To test this hypothesis, "Background" uses the posterior probability of the background class as the anomaly score. 
The Dropout method leverages MC Dropout \parencite{gal} to obtain an epistemic uncertainty estimate.  We follow the implementation by \cite{Kendall2015BayesianSM}.
MC Dropout is computed by leaving dropout on during inference and then one runs several forward passes of the image through the network.  This creates a set of predictions for the given object that we use to compute the variance over the set of predictions. The variance over the set of predictions serves as the anomaly score.  A higher variance implies higher uncertainty about which class the object belongs to.  
We also experiment with an autoencoder baseline similar to \cite{Baur_2019, haselmann2018anomaly} where pixel-wise reconstruction loss is used as the anomaly score.  By this we mean that we run the image through the autoencoder and subtract the resulting image from the input.  The absolute difference in magnitude serves as the anomaly score.  This method is called AE. 
The "Branch" method is a direct port of the confidence branch detector from \cite{devries} to pixel-wise prediction.  This method trains a separate final output score along with a classification output.  The output score is multiplied to the predictions so that the predictions are scaled based on the model's confidence.  The training involves the slight modification to cross entropy $\mathcal{L} = - y log(p \cdot c)$ where y is the label, p is the probability, and c is the confidence score.  The confidence score is always between 0-1 by applying the sigmoid function ensuring that the cross entropy function is still valid.  The confidence is trained via backpropagation similarly to the classification prediction.  
Finally, we use the MaxLogit method described in earlier sections.

For all of the baselines except the autoencoder, we train a PSPNet \parencite{Zhao2017PyramidSP} decoder with a ResNet-101 encoder \parencite{resnet} for $20$ epochs.  The PSPNet follows a similar pattern to Zoomout \parencite{mostajabi2015zoomout} or Hypercolumns \parencite{hypercol} whereby the activations of the convolutional layers are concatenated together before finally undergoing a fully connected layer to arrive at the appropriate dimensionality.  We train both the encoder and decoder using SGD with momentum of 0.9, a learning rate of $2\times10^{-2}$, and learning rate decay of $10^{-4}$. 
For the autoencoder, we use a 4-layer U-Net \parencite{Ronneberger_2015} with a spatial latent code as in \parencite{Baur_2019}. The U-Net also uses batch norm and is trained for $10$ epochs.  

To evaluate the methods, we take all of the scores per pixel that belong to anomalies and all of the scores for the remaining pixels. Then we sort the scores, thereby allowing us to compute the false positive rate and true positive rate at every threshold. We use the thresholds to compute the AUROC giving us the probability that we correctly select an in-distribution pixel with the method we're evaluating. We also predefine a set threshold at 5\% and compute the false positive rate for that threshold. Due to the large number of scores to be evaluated and sorted, we take the mean over all the images of each evaluation as a final report.

\begin{table*}[t]
 \setlength{\tabcolsep}{8pt}
 \centering
 \begin{tabularx}{\textwidth}{*{1}{>{\hsize=1\hsize}X} *{1}{>{\hsize=0.9cm}X } c
 | *{1}{>{\hsize=0.4\hsize}Y} *{1}{>{\hsize=0.6\hsize}Y} *{1}{>{\hsize=0.8\hsize}Y} *{1}{>{\hsize=0.6\hsize}Y} *{1}{>{\hsize=0.6\hsize}Y} *{1}{>{\hsize=0.6\hsize}Y}}
 & & & MSP & Branch & Background & Dropout & AE & MaxLogit \\ \hline
 \parbox[t]{50mm}{\multirow{3}{*}{\rotatebox{0}{StreetHazards}}}
 & FPR95 &$\downarrow$ & 33.7 & 68.4 & 69.0 & 79.4 & 91.7 & \textbf{26.5} \\
 & AUROC &$\uparrow$ & 87.7   & 65.7 & 58.6 & 69.9 & 66.1 & \textbf{89.3} \\
 & AUPR &$\uparrow$ & 6.6 & 1.5 & 4.5  & 7.5 & 2.2  & \textbf{10.6} \\
 \Xhline{3\arrayrulewidth}
 \parbox[t]{50mm}{\multirow{3}{*}{\rotatebox{0}{BDD-Anomaly}}}
 & FPR95 &$\downarrow$ & 24.5 & 25.6 & 40.1 & 16.6 & 74.1 & \textbf{14.0} \\
 & AUROC &$\uparrow$ & 87.7 & 85.6 & 69.7 & 90.8 & 64.0 & \textbf{92.6} \\
 & AUPR &$\uparrow$ & 3.7 & 3.9 & 1.1 & 4.3 & 0.7 & \textbf{5.4} \\
 \Xhline{3\arrayrulewidth}
 \end{tabularx}
 \caption{Results on the Combined Anomalous Object Segmentation benchmark. AUPR is low across the board due to the large class imbalance, but all methods perform substantially better than chance. MaxLogit obtains the best performance. All results are percentages.}
 \label{tab:CAOS}
\end{table*}

\noindent\textbf{Results and Analysis.}\quad
MaxLogit outperforms all other methods across the board by a substantial margin. The intuitive baseline of using the posterior for the background class to detect anomalies performs poorly, which suggests that the background class may not align with rare visual features. Even though reconstruction-based scores succeed in product fault segmentation, we find that the AE method performs poorly on the CAOS benchmark, which may be due to the more complex domain. AUPR for all methods is low, indicating that the large class imbalance presents a serious challenge. However, the substantial improvements with the MaxLogit method suggest that progress on this task is possible and there is much room for improvement.

In  figure \ref{fig:StreetHazards2}, we can qualitatively see that both MaxLogit and MSP have a high number of false positives, as they assign high anomaly scores to semantic boundaries, a problem also observed in the recent works of \cite{blum2019fishyscapes, thesis_canny}. However, the problem is less severe in MaxLogit.  A potential explanation for this could be due to two effects.  The first we mentioned earlier in the benefits of MaxLogit over MSP (see section \ref{anom_seg:sec:multiclass}) in that the inter-class variance is better preserved in MaxLogit over MSP.  The second effect builds off of the first, by having a greater range in MaxLogit as compared to MSP bilinear upsampling from the models final output to the final output image creates much sharper boundaries.  This is because the interpolation of points that are already close (with MSP) will blur the boundaries much more and cause more pixels to become classified as in-distribution by exceeding the threshold.  

Autoencoder-based methods are qualitatively different from approaches using the softmax probabilities, because they model the input itself and can avoid boundary effects seen in the MaxLogit and MSP rows of figure \ref{fig:StreetHazards2}. While autoencoder methods are successful in medical anomaly segmentation and product fault detection, we find the AE baseline to be ineffective in the more complex domain of street scenes. The last row of figure \ref{fig:StreetHazards2} shows pixel-wise reconstruction loss on example images from StreetHazards. Anomalies are not distinguished well from in-distribution elements of the scene. New methods must be developed to mitigate the boundary effects faced by softmax-based methods while also attaining good detection performance.

\section{Conclusion}
We scaled out-of-distribution detection to more realistic, large-scale settings by developing a novel benchmark for OOD segmentation. The CAOS benchmark for anomaly segmentation consists of diverse, naturally-integrated anomalous objects in driving scenes. Baseline methods on the CAOS benchmark substantially improve on random guessing but are still lacking, indicating potential for future work. We also investigated using multi-label classifiers for out-of-distribution detection and established an experimental setup for this previously unexplored setting. On large-scale multi-class image datasets, we identified an issue faced by existing baselines and proposed the maximum logit detector as a natural solution. Interestingly, this detector also provides consistent and significant gains in the multi-label and anomaly segmentation settings, thereby establishing it as a new baseline in place of the maximum softmax probability baseline on large-scale OOD detection problems. In all, we hope that our simple baseline and our new OOD segmentation benchmark will enable further research on out-of-distribution detection for real-world safety-critical environments.\looseness=-1

\chapter{Neural augmentations}\label{ch:neural_aug}

\section{Overview}
While the research community must create robust models that generalize to new scenarios, the robustness literature \parencite{dodge2017study, geirhos2020shortcut} lacks consensus on evaluation benchmarks and contains many dissonant hypotheses.
While \cite{hendrycks2020pretrained} find that many recent language models are already robust to many forms of distribution shift, \cite{yin2019fourier}, and \cite{geirhos2019} find that vision models are largely fragile and argue that data augmentation offers a solution. In contrast, \cite{taori2020when} provide results suggesting that using pretraining and improving in-distribution test set accuracy improve natural robustness, whereas other methods do not.

In this chapter we articulate and systematically study seven robustness hypotheses. The first four hypotheses concern \emph{methods} for improving robustness, while the last three hypotheses concern abstract \emph{properties} about robustness. These hypotheses are as follows.

\begin{itemize}[noitemsep,topsep=0pt,parsep=0pt,partopsep=0pt,leftmargin=12pt]
\item \textit{ Larger Models}: increasing model size improves robustness  \parencite{ Xie2020Intriguing}. 
\item \textit{ Self-Attention}: adding self-attention layers to models improves robustness \parencite{Hendrycks2019NaturalAE}.\looseness=-1
\item \textit{ Diverse Data Augmentation}: robustness can increase through data augmentation \parencite{yin2019fourier}.
\item \textit{ Pretraining}: pretraining on larger and more diverse datasets improves robustness \parencite{Orhan2019RobustnessPO,Hendrycks2019Pretraining}.
\item \textit{ Texture Bias}: convolutional networks are biased towards texture, which harms robustness \parencite{geirhos2019}.
\item \textit{ Only IID Accuracy Matters}: accuracy on independent and identically distributed test data entirely determines natural robustness \parencite{Taori2020RechtDistributionShiftImagenet}.
\item \textit{ Synthetic $\notimplies$ Natural}: \emph{synthetic} robustness interventions including diverse data augmentations do not help with robustness on \emph{naturally occurring} distribution shifts \parencite{taori2020when}.
\end{itemize}

\begin{figure}[t]
\begin{center}
\includegraphics[width=\textwidth]{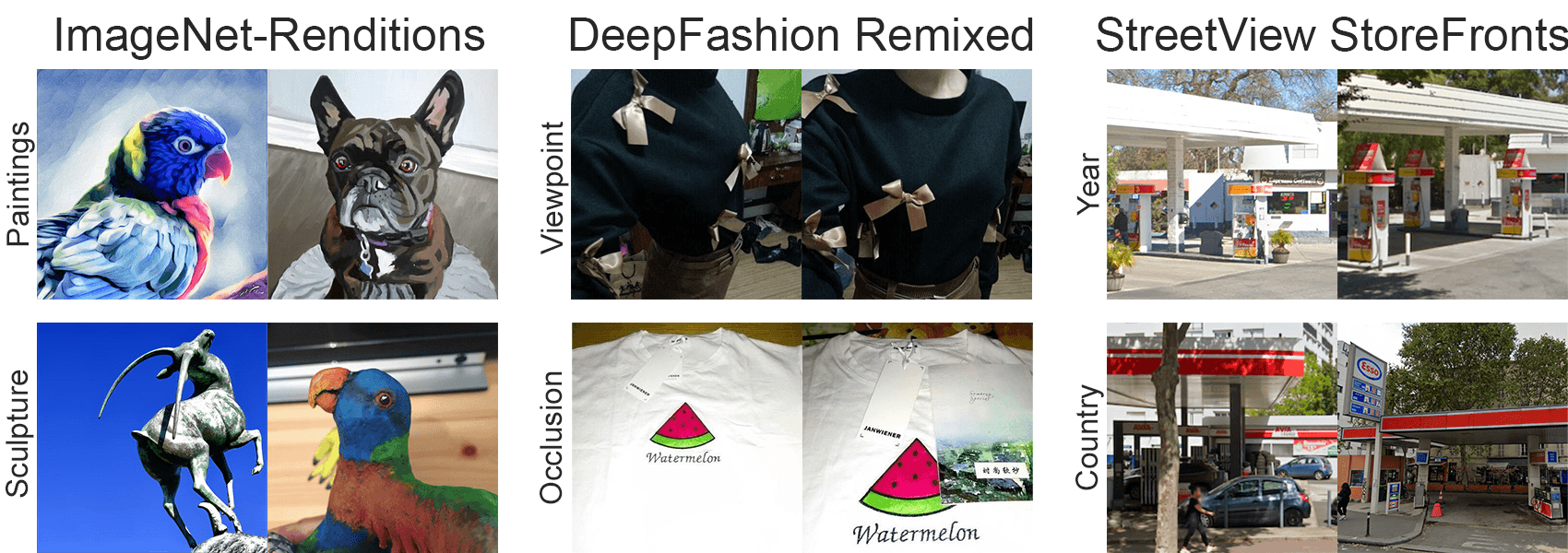}
\end{center}
\caption{
Images from our three new datasets ImageNet-Renditions (ImageNet-R), DeepFashion Remixed (DFR), and StreetView StoreFronts (SVSF).
The SVSF images are recreated from the public Google StreetView, copyright Google 2020. Our datasets test robustness to various naturally occurring distribution shifts including rendition style, camera viewpoint, and geography.
}\label{fig:ch5_splash}
\end{figure}

It has been difficult to arbitrate these hypotheses because existing robustness datasets preclude the possibility of controlled experiments by varying multiple aspects simultaneously. For instance, \textit{ Texture Bias} was initially investigated with synthetic distortions \parencite{geirhos}, which conflicts with the \textit{ Synthetic $\notimplies$ Natural} hypothesis.
On the other hand, natural distribution shifts often affect many factors (e.g., time, camera, location, etc.) simultaneously in unknown ways \parencite{Recht2019DoIC, Hendrycks2019NaturalAE}.
Existing datasets also lack diversity such that it is hard to extrapolate which methods will improve robustness more broadly. To address these issues and test the seven hypotheses outlined above, we introduce three new robustness benchmarks and a new data augmentation method.

First we introduce ImageNet-Renditions (ImageNet-R), a 30,000 image test set containing various renditions (e.g., paintings, embroidery, etc.) of ImageNet object classes. These renditions are naturally occurring, with textures and local image statistics unlike those of ImageNet images, allowing us to more cleanly separate the \emph{Texture Bias} and \emph{Synthetic $\notimplies$ Natural} hypotheses. 

Next, we investigate natural shifts in the image capture process with StreetView StoreFronts (SVSF) and DeepFashion Remixed (DFR).  SVSF contains business storefront images taken from Google Streetview, along with metadata allowing us to vary location, year, and even the camera type. DFR leverages the metadata from DeepFashion2 \parencite{ge2019deepfashion2} to systematically shift object occlusion, orientation, zoom, and scale at test time.
Both SVSF and DFR provide distribution shift controls and do not alter texture, which remove possible confounding variables affecting prior benchmarks.  DFR is discussed in greater detail in \ref{ch:multilabel_sec:DFR}.


Finally, we contribute DeepAugment to increase robustness to some new types of distribution shift. This augmentation technique uses image-to-image neural networks for data augmentation, not data-independent Euclidean augmentations like image shearing or rotating as in previous work. DeepAugment achieves state-of-the-art robustness on our newly introduced ImageNet-R benchmark and a corruption robustness benchmark. DeepAugment can also be combined with other augmentation methods to outperform a model pretrained on $1000\times$ more labeled data.

After examining our results on these three datasets and others, we can rule out several of the above hypotheses while strengthening support for others. 
As one example, we find that synthetic data augmentation robustness interventions improve accuracy on ImageNet-R and real-world image blur distribution shifts, providing clear counterexamples to \textit{ Synthetic $\notimplies$ Natural} while lending support to the \emph{Diverse Data Augmentation} and \textit{Texture Bias} hypotheses. In the conclusion, we summarize the various strands of evidence for and against each hypothesis.
Across our many experiments, we do not find a general method that consistently improves robustness, and some hypotheses require additional qualifications. While robustness is often spoken of and measured as a single scalar property like accuracy, our investigations suggest that robustness is not so simple. In light of our results, we hypothesize in the conclusion that robustness is \emph{multivariate}.

\section{Related Work}

\paragraph{Robustness Benchmarks.}
Recent works \parencite{hendrycks2019robustness, Recht2019DoIC, hendrycks2020pretrained} have begun to characterize model performance on out-of-distribution (OOD) data with various new test sets, with dissonant findings. For instance, \cite{hendrycks2020pretrained} demonstrate that modern language processing models are moderately robust to numerous naturally occurring distribution shifts, and that \emph{Only IID Accuracy Matters} is inaccurate for natural language tasks. For image recognition, \cite{hendrycks2019robustness} analyze image models and show that they are sensitive to various simulated image corruptions (e.g., noise, blur, weather, JPEG compression, etc.) from their ``ImageNet-C'' benchmark.

\cite{Recht2019DoIC} reproduce the ImageNet \parencite{imagenet} validation set for use as a benchmark of naturally occurring distribution shift in computer vision. Their evaluations show a 11-14\% drop in accuracy from ImageNet to the new validation set, named ImageNetV2, across a wide range of architectures. \cite{taori2020when} use ImageNetV2 to measure natural robustness and dismiss \emph{Diverse Data Augmentation}. 
\cite{engstrom2020identifying} identify statistical biases in ImageNetV2's construction, and they estimate that reweighting ImageNetV2 to correct for these biases results in a less substantial 3.6\% drop.

In contrast to \emph{adversarial} robustness \parencite{szegedy2013intriguing, goodfellow2014explaining}, we focus instead on robustness to unconstrained out-of-distribution data measured on non-interactive benchmarks.
\parencite{carlini2019evaluating} detail the inherent practical difficulty in evaluating adversarial robustness, and \parencite{Gilmer2018MotivatingTR} argue for the inclusion of unconstrained input modifications in the threat model of attacks against machine learning systems.

\paragraph{Data Augmentation.}

\cite{geirhos2019, yin2019fourier, hendrycks2019augmix} demonstrate that data augmentation can improve robustness on ImageNet-C.
The space of augmentations that help robustness includes various types of noise \parencite{madry, rusak2020increasing, Lopes2019ImprovingRW}, highly unnatural image transformations \parencite{geirhos2019, Yun2019CutMixRS, zhang2017mixup}, or compositions of simple image transformations such as Python Imaging Library operations \parencite{Cubuk2018AutoAugmentLA, hendrycks2019augmix}.
Some of these augmentations can improve accuracy on in-distribution examples as well as on out-of-distribution (OOD) examples.

\paragraph{Transfer learning}
Pretraining models on larger datasets has been been demonstrated to improve robustness on ImageNet-C. \cite{Orhan2019RobustnessPO} report that models trained on the JFT-300m dataset \parencite{Sun_2017_ICCV} and the weakly-supervised Instagram dataset \parencite{instagram2018} improve robustness on the ImageNet-C by significant margins. 

\section{New Benchmarks}
In order to evaluate the seven robustness hypotheses, we introduce three new benchmarks that capture new types of naturally occurring distribution shifts. 
ImageNet-Renditions (ImageNet-R) is a newly collected test set intended for ImageNet classifiers, whereas StreetView StoreFronts (SVSF) and DeepFashion Remixed (DFR) each contain their own training sets and multiple test sets.
SVSF and DFR split data into a training and test sets based on various image attributes stored in the metadata. For example, we can select a test set with images produced by a camera different from the training set camera. We now describe the structure and collection of each dataset.

\subsection{ImageNet-Renditions (ImageNet-R)}

\begin{figure}[t]
\begin{center}
\includegraphics[width=\textwidth]{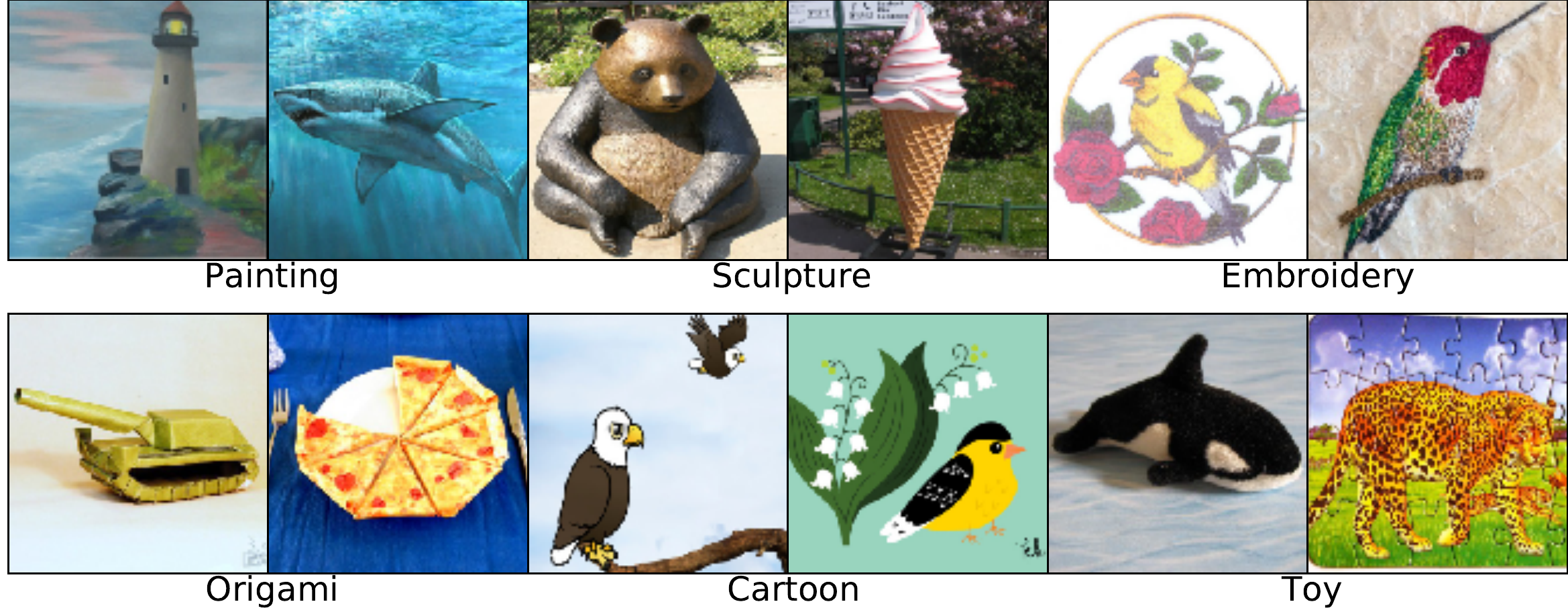}
\end{center}
\caption{
ImageNet-Renditions (ImageNet-R) contains 30,000 images of ImageNet objects with different textures and styles. This figure shows only a portion of ImageNet-R's numerous rendition styles.
The rendition styles (e.g., ``Toy'') are for clarity and are \emph{not} ImageNet-R's classes; ImageNet-R's classes are a subset of 200 ImageNet classes. ImageNet-R emphasizes shape over texture.
}\label{fig:renditions}
\end{figure}

While current classifiers can learn some aspects of an object's shape \parencite{mordvintsev2015inceptionism}, they nonetheless rely heavily on natural textural cues \parencite{geirhos2019}. In contrast, human vision can process abstract visual renditions.
For example, humans can recognize visual scenes from line drawings as quickly and accurately as they can from photographs \parencite{biederman1988surface}.
Even some primates species have demonstrated the ability to recognize shape through line drawings \parencite{Itakura1994,Tanaka2006}.

To measure generalization to various abstract visual renditions, we create the ImageNet-Rendition (ImageNet-R) dataset. ImageNet-R contains various artistic renditions of object classes from the original ImageNet dataset. Note the original ImageNet dataset discouraged such images since annotators were instructed to collect "photos only, no painting, no drawings, etc." \parencite{deng2012large}. We do the opposite.





\paragraph{Data Collection.}
ImageNet-R contains 30,000 image renditions for 200 ImageNet classes. We collect images primarily from Flickr and use queries such as "art," "cartoons," "graffiti," "embroidery," "graphics," "origami," "paintings," "patterns," "plastic objects," "plush objects," "sculptures," "line drawings," "tattoos," "toys," "video game," and so on. Examples are depicted in figure \ref{fig:renditions}. Images are filtered by Amazon MTurk workers using a modified collection interface from ImageNetV2 \parencite{Recht2019DoIC}. The resulting images are then manually filtered by graduate students. 
ImageNet-R also includes the line drawings from \parencite{wang2019learning}, excluding horizontally mirrored duplicate images, pitch black images, and images from the incorrectly collected "pirate ship" class.

\subsection{StreetView StoreFronts (SVSF)} \label{ch:neuraug_sec:svsf}

Computer vision applications often rely on data from complex pipelines that span different hardware, times, and geographies. Ambient variations in this pipeline may result in unexpected performance degradation, such as degradations experienced by health care providers in Thailand deploying laboratory-tuned diabetic retinopathy classifiers in the field \parencite{beede2020human}. In order to study the effects of shifts in the image capture process we collect the StreetView StoreFronts (SVSF) dataset, a new image classification dataset sampled from Google StreetView imagery \parencite{anguelov2010google} focusing on three distribution shift sources: country, year, and camera.

\paragraph{Data Collection.}

SVSF consists of cropped images of business store fronts extracted from StreetView images by an object detection model.
Each store front image is assigned the class label of the associated Google Maps business listing through a combination of machine learning models and human annotators.
We combine several visually similar business types (e.g. drugstores and pharmacies) for a total of 20 classes, listed Appendix \ref{app:classes}.
We are currently unable to release the SVSF data publicly.

Splitting the data along the three metadata attributes of country, year, and camera, we create one training set and five test sets.
We sample a training set and an in-distribution test set (200K and 10K images, respectively) from images taken in US/Mexico/Canada during 2019 using a "new" camera system.
We then sample four OOD test sets (10K images each) which alter one attribute at a time while keeping the other two attributes consistent with the training distribution.
Our test sets are year: 2017, 2018; country: France; and camera: ``old.''

\section{DeepAugment}



In order to further explore the \emph{Diverse Data Augmentation} hypothesis, we introduce a new data augmentation technique we call DeepAugment. DeepAugment works by passing an image through an image-to-image network (such as an image autoencoder or a superresolution network), but rather than processing the image normally, we distort the internal weights and activations. We distort the image-to-image network's weights by applying randomly sampled operations such as zeroing, negating, convolving, transposing, applying activation functions, and so on. This creates diverse but semantically consistent images as illustrated in figure  \ref{fig:deepaugment}. We provide the pseudocode in Appendix~\ref{app:deepaugment}. Whereas most previous data augmentations techniques use simple augmentation primitives applied to the raw image itself, we stochastically distort the internal representations of image-to-image networks to augment images.  We did not experiment with any geometric image-to-image networks such as NeRF \parencite{mildenhall2020nerf}.  We have results in table \ref{tab:imagenetr} that shows how DeepAugment works well with Augmix which does incorporate geometric transformations.

\begin{figure}[t]
\begin{center}
\includegraphics[width=\textwidth]{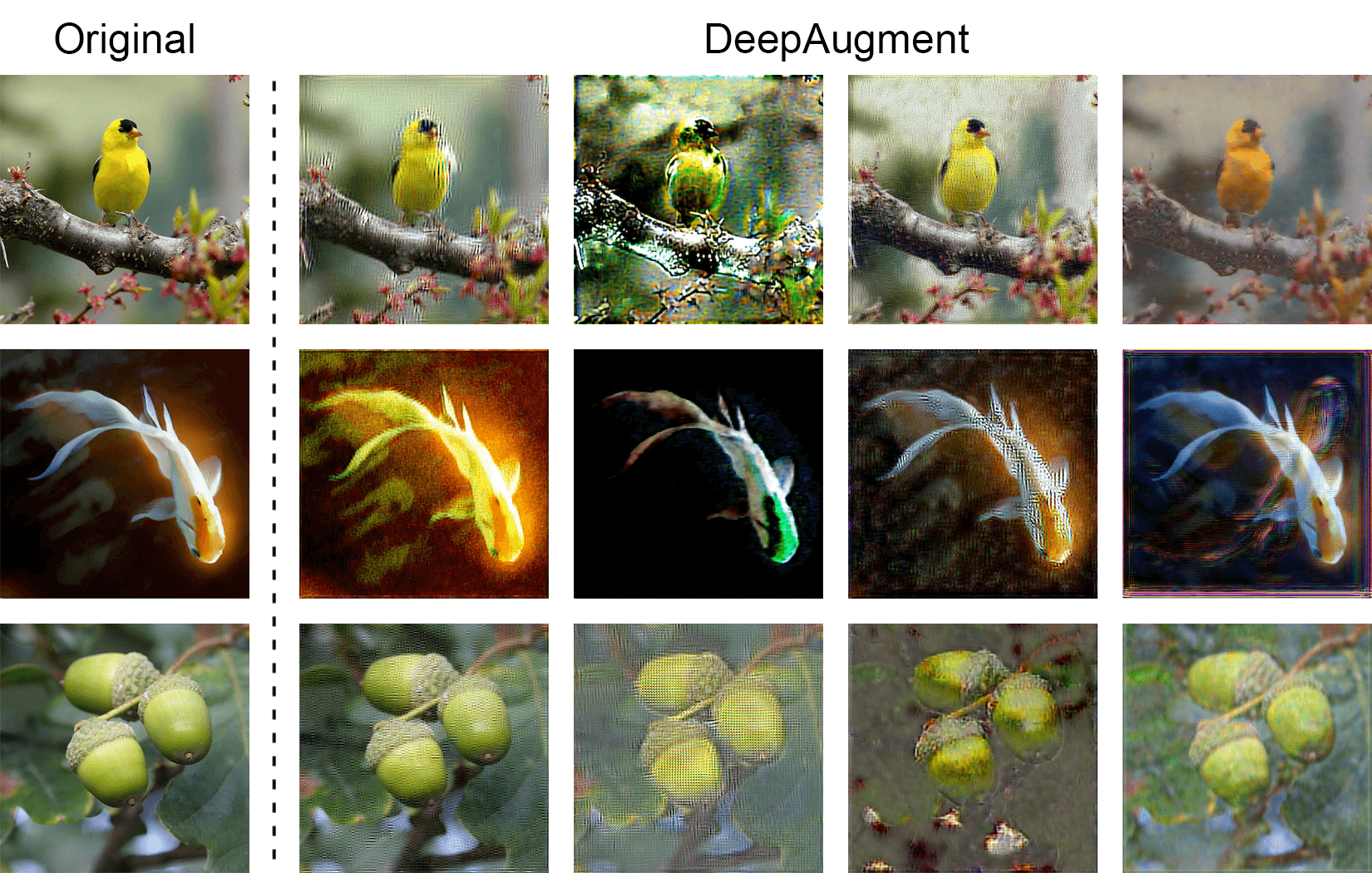}
\end{center}
\caption{
DeepAugment examples preserve semantics, are data-dependent, and are far more visually diverse than augmentations such as rotations.
}\label{fig:deepaugment}
\end{figure}

\section{Experiments}
\subsection{Setup}

In this section we briefly describe the evaluated models, pretraining techniques, self-attention mechanisms, data augmentation methods, and note various implementation details.

\noindent\textbf{Model Architectures and Sizes.}\quad Most experiments are evaluated on a standard ResNet-50 model \parencite{resnet}. Model size evaluations use ResNets or ResNeXts \parencite{resnext} of varying sizes.

\noindent\textbf{Pretraining.}\quad 
For pretraining we use ImageNet-21K which contains approximately 21,000 classes and approximately 14 million labeled training images, or around $10\times$ more labeled training data than ImageNet-1K. We tune \cite{kolesnikov2019large}'s ImageNet-21K model.
We also use a large pre-trained ResNeXt-101 model from \cite{instagram2018}. This was pre-trained on on approximately 1 billion Instagram images with hashtag labels and fine-tuned on ImageNet-1K. This Weakly Supervised Learning (WSL) pretraining strategy uses approximately $1000\times$ more labeled data.

\noindent\textbf{Self-Attention.}\quad
When studying self-attention, we employ CBAM \parencite{woo2018cbam} and SE \parencite{Hu2018SqueezeandExcitationN} modules, two forms of self-attention that help models learn spatially distant dependencies.

\noindent\textbf{Data Augmentation.}\quad
We use Style Transfer, AugMix, and DeepAugment to analyze the \emph{Diverse Data Augmentation} hypothesis, and we contrast their performance with simpler noise augmentations such as Speckle Noise and adversarial noise.
Style transfer \parencite{geirhos2019} uses a style transfer network to apply artwork styles to training images. AugMix \parencite{hendrycks2019augmix} randomly composes simple augmentation operations (e.g., translate, posterize, solarize).
DeepAugment, introduced above, distorts the weights and feedforward passes of image-to-image models to generate image augmentations. Speckle Noise data augmentation muliplies each pixel by $(1+x)$ with $x$ sampled from a normal distribution \parencite{rusak2020increasing,hendrycks2019robustness}. We also consider adversarial training as a form of adaptive data augmentation and use the model from \parencite{wong2020fast} trained against $\ell_\infty$ perturbations of size $\varepsilon = 4/255$.

\begin{table*}[t]
\begin{center}
\begin{tabular}{lcc>{\color{gray}}c}
\toprule
                Error Rates    & ImageNet-200 (\%) & ImageNet-R (\%) & Gap      \\ \midrule
ResNet-50               &  7.9         &  63.9       &  56.0    \\
+ ImageNet-21K \emph{Pretraining} ($10\times$ labeled data)          &  7.0                  &  62.8             &  55.8     \\
+ CBAM (\emph{Self-Attention})                        &  7.0                  &  63.2             &  56.2     \\
+ $\ell_\infty$ Adversarial Training    &  25.1        &  68.6       &  43.5    \\
+ Speckle Noise              &  8.1                  &  62.1             &  54.0     \\
+ Style Transfer Augmentation          &  8.9         &  58.5       &  49.6    \\ 
+ AugMix                  &  7.1         &  58.9       &  51.8    \\
+ DeepAugment             &  7.5         &  57.8       &  50.3    \\
+ DeepAugment + AugMix      &  8.0         &  53.2 &  45.2 \\\midrule
ResNet-152 (\emph{Larger Models})                             &  6.8                  &  58.7             &  51.9     \\
\bottomrule
\end{tabular}
\end{center}
\caption{ImageNet-200 and ImageNet-R top-1 error rates. ImageNet-200 uses the same 200 classes as ImageNet-R. DeepAugment+AugMix improves over the baseline by over 10 percentage points. ImageNet-21K Pretraining tests \emph{Pretraining} and  CBAM tests \emph{Self-Attention}. Style Transfer, AugMix, and DeepAugment test \emph{Diverse Data Augmentation} in contrast to simpler noise augmentations such as $\ell_\infty$ Adversarial Noise and Speckle Noise. While there remains much room for improvement, results indicate that progress on ImageNet-R is tractable.}
\label{tab:imagenetr}
\end{table*}

\subsection{Results}
We now perform experiments on ImageNet-R, and StreetView StoreFronts leaving results on DeepFashion Remixed to Chapter \ref{ch:multilabelood} on multilabel OOD. We also evaluate on ImageNet-C and compare and contrast it with real distribution shifts.


\paragraph{ImageNet-R.}
Table \ref{tab:imagenetr} shows performance on ImageNet-R as well as on ImageNet-200 (the original ImageNet data restricted to ImageNet-R's 200 classes). This has several implications regarding the four method-specific hypotheses. \emph{Pretraining} with ImageNet-21K (approximately $10\times$ labeled data) hardly helps. Appendix \ref{app:additional} shows WSL pretraining can help, but Instagram has renditions, while ImageNet excludes them; hence we conclude comparable pretraining was ineffective. Notice \emph{Self-Attention} increases the IID/OOD gap. Compared to simpler data augmentation techniques such as Speckle Noise, the \emph{Diverse Data Augmentation} techniques of Style Transfer, AugMix, and DeepAugment improve generalization. Note AugMix and DeepAugment improve in-distribution performance whereas Style transfer hurts it. Also, our new DeepAugment technique is the best standalone method with an error rate of 57.8\%.
Last, \emph{Larger Models} reduce the IID/OOD gap. Full results for all evaluated models can be found in the Appendix in table \ref{tab:imagenetr_full}.

Regarding the three more abstract hypotheses, 
biasing networks away from natural textures through diverse data augmentation improved performance, so we find support for the \emph{Texture Bias} hypothesis.
The IID/OOD generalization gap varies greatly which condtradicts \emph{Only IID Accuracy Matters}. Finally, since ImageNet-R contains real-world examples, and since synthetic data augmentation helps on ImageNet-R, we now have clear evidence against the \emph{Synthetic $\notimplies$ Natural} hypothesis.




\noindent\textbf{StreetView StoreFronts.}\quad
In table \ref{tab:svsf}, we evaluate data augmentation methods on SVSF and find that all of the tested methods have mostly similar performance and that no method helps much on country shift, where error rates roughly double across the board. Images captured in France contain noticeably different architectural styles and storefront designs than those captured in US/Mexico/Canada; meanwhile, we are unable to find conspicuous and consistent indicators of the camera and year.
This may explain the relative insensitivity of evaluated methods to the camera and year shifts.
Overall \emph{Diverse Data Augmentation} shows limited benefit, suggesting either that data augmentation primarily helps combat texture bias as with ImageNet-R, or that existing augmentations are not diverse enough to capture high-level semantic shifts such as building architecture.

\begin{table}[h]
\begin{center}
{
\begin{tabular}{@{}l ? c | c | c c | c}
\multicolumn{2}{c}{} & \multicolumn{1}{c}{Hardware} & \multicolumn{2}{c}{Year} & \multicolumn{1}{c}{Location} \\
\hline
Network & \multicolumn{1}{c|}{\,IID\,}
    & Old & 2017 & 2018 & France 
    \\ \hline 
ResNet-50               & 27.2 & 28.6 & 27.7 & 28.3 & 56.7 \\
+ Speckle Noise         & 28.5 & 29.5 & 29.2 & 29.5 & 57.4 \\
+ Style Transfer        & 29.9 & 31.3 & 30.2 & 31.2 & 59.3 \\
+ DeepAugment           & 30.5 & 31.2 & 30.2 & 31.3 & 59.1\\
+ AugMix                & 26.6 & 28.0 & 26.5 & 27.7 & 55.4 \\

\Xhline{2\arrayrulewidth}
\end{tabular}}
\caption{SVSF classification error rates. Networks are robust to some natural distribution shifts but are substantially more sensitive the geographic shift. Here \emph{Diverse Data Augmentation} hardly helps.}
\label{tab:svsf}
\end{center}
\end{table}

\begin{figure}
	\begin{center}
      \centering
      \includegraphics[width=\linewidth]{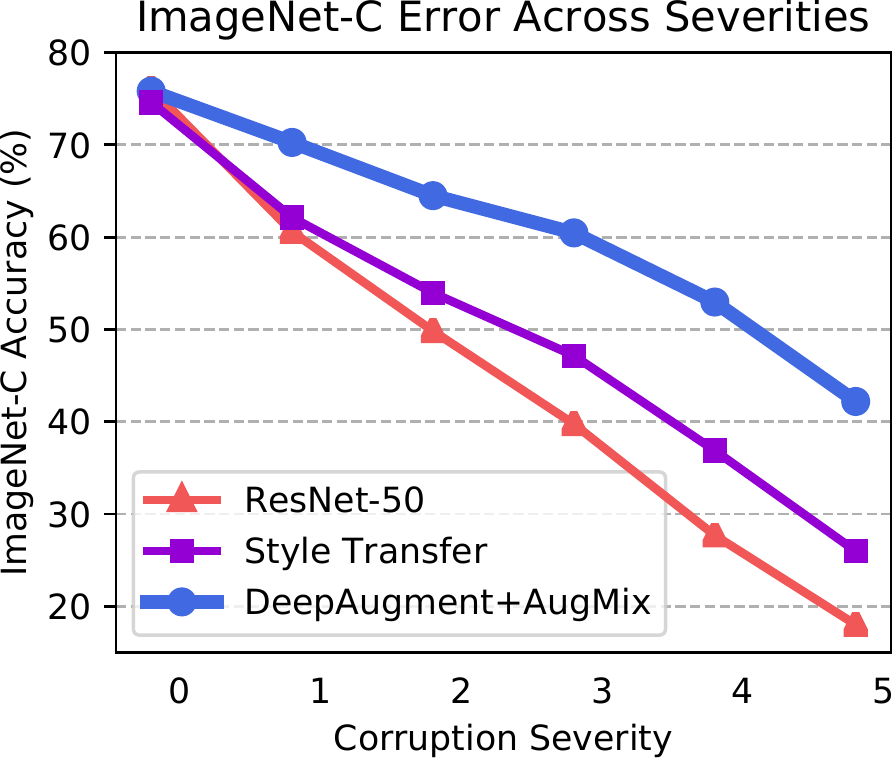}
      \caption{Accuracy as a function of corruption severity. Severity ``0'' denotes clean data. Data augmentation methods such as DeepAugment with AugMix shift the entire Pareto frontier outward.}
  \end{center}
\end{figure}

\begin{figure}{htbp}
	\centering
      \includegraphics[width=\linewidth]{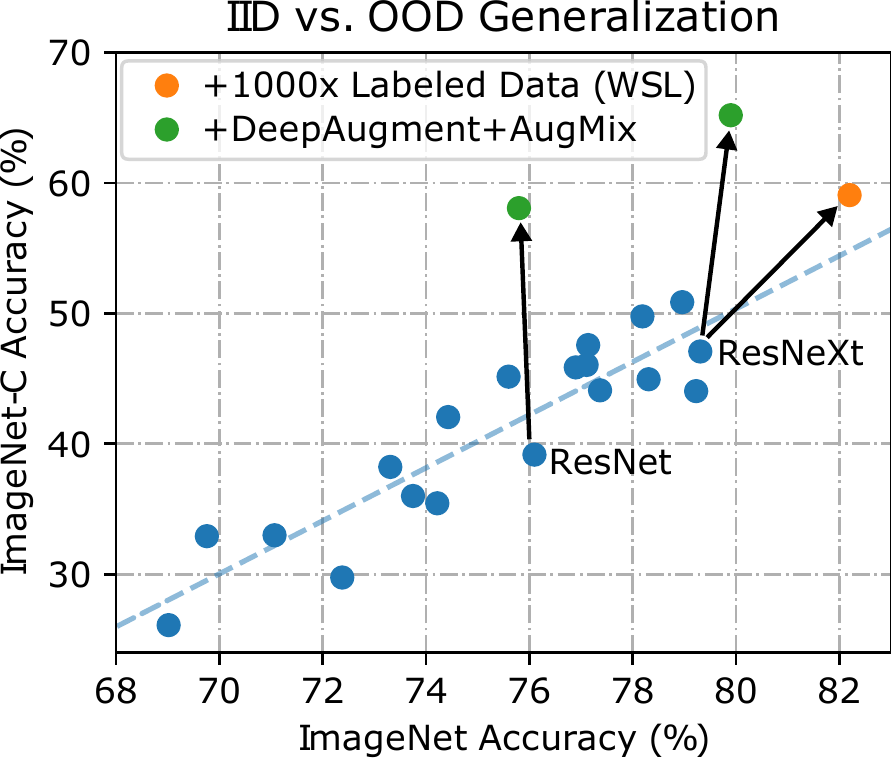}
  \caption{ImageNet accuracy and ImageNet-C accuracy. Previous architectural advances slowly translate to ImageNet-C performance improvements, but DeepAugment+AugMix on a ResNet-50 yields a $\approx19\%$ accuracy increase.
  }
  \label{fig:imagenetc}
\end{figure}

\noindent\textbf{ImageNet-C.}\quad
We now consider a previous robustness benchmark to reassess all seven hypotheses. We use the ImageNet-C dataset \parencite{hendrycks2019robustness} which applies 15 common image corruptions (e.g., Gaussian noise, defocus blur, simulated fog, JPEG compression, etc.) across 5 severities to ImageNet-1K validation images. We find that DeepAugment improves robustness on ImageNet-C. Figure \ref{fig:imagenetc} shows that when models are trained with AugMix and DeepAugment, they attain the state-of-the-art, break the trendline, and exceed the corruption robustness provided by training on $1000\times$ more labeled training data. Note the augmentations from AugMix and DeepAugment are disjoint from ImageNet-C's corruptions. Full results are shown in Appendix \ref{app:additional}'s \ref{tab:imagenetc_table}. This is evidence against the \emph{Only IID Accuracy Matters} hypothesis and is evidence for the \emph{Larger Models}, \emph{Self-Attention}, \emph{Diverse Data Augmentation}, \emph{Pretraining}, and \emph{Texture Bias} hypotheses.

\cite{taori2020when} remind us that ImageNet-C uses various \emph{synthetic} corruptions and suggest that they are divorced from real-world robustness. Real-world robustness requires generalizing to naturally occurring corruptions such as snow, fog, blur, low-lighting noise, and so on, but it is an open question whether ImageNet-C's simulated corruptions meaningfully approximate real-world corruptions.

We collect a small dataset of 1,000 real-world blurry images and find that ImageNet-C can track robustness to real-world corruptions. We collect the "Real Blurry Image" dataset with Flickr and query ImageNet object class names concatenated with the word "blurry." We then evaluate various models on real-world blurry images and find that \emph{all} the robustness interventions that help with ImageNet-C also help with real-world blurry images. Hence ImageNet-C can track performance on real-world corruptions. Moreover, DeepAugment+AugMix has the lowest error rate on Real Blurry Images, which again contradicts the \textit{ Synthetic $\notimplies$ Natural} hypothesis. Appendix \ref{app:blur} has full results. The upshot is that ImageNet-C is a controlled and systematic proxy for real-world robustness.

\section{Conclusion}

\begin{table}[t]
\setlength{\tabcolsep}{5pt}
\centering
\begin{tabular}{lcccccc}
  Hypothesis       & ImageNet-C & Real Blurry Images & ImageNet-R  & DFR & SVSF \\ 
\hline
\emph{Larger Models}    &  $+$ & $+$ & $+$ & $-$ \\
\emph{Self-Attention}   & $+$ & $+$ & $-$ & $-$  \\
\emph{Diverse Data Augmentation}       & $+$ & $+$ & $+$ & $-$ & $-$  \\
\emph{Pretraining}      & $+$ & $+$ & $-$ & $-$  \\
\bottomrule
\end{tabular}
\caption{A highly simplified account of each hypothesis when tested against different datasets. Evidence for is denoted "$+$", and "$-$" denotes an absence of evidence or evidence against.}
\label{tab:hypothesissummary}
\end{table}

We introduced two new multi-class benchmarks, ImageNet-Renditions, and StreetView StoreFronts.
With these benchmarks, we thoroughly tested seven robustness hypotheses--four about methods for robustness, and three about the nature of robustness.

Let us consider the first four hypotheses, using the new information from ImageNet-C and our three new benchmarks.
The \emph{Larger Models} hypothesis was supported with ImageNet-C and ImageNet-R, but not with DFR.
While \emph{Self-Attention} noticeably helped ImageNet-C, it did not help with ImageNet-R and DFR.
\emph{Diverse Data Augmentation} was ineffective for SVSF and DFR, but it greatly improved ImageNet-C and ImageNet-R accuracy.
\emph{Pretraining} greatly helped with ImageNet-C but hardly helped with ImageNet-R. This is summarized in table \ref{tab:hypothesissummary}. It was not obvious \emph{a priori} that synthetic \emph{Diverse Data Augmentation} could improve ImageNet-R accuracy, nor did previous research suggest that \emph{Pretraining} would sometimes be ineffective. While no single method consistently helped across all distribution shifts, some helped more than others.


Our analysis of these four hypotheses have implications for the remaining three hypotheses. Regarding \emph{Texture Bias}, ImageNet-R shows that networks do not generalize well to renditions (which have different textures), but that diverse data augmentation (which often distorts textures) can recover accuracy. More 
generally, larger models and diverse data augmentation consistently helped 
on ImageNet-R, ImageNet-C, and Blurry Images, suggesting that these two 
interventions reduce texture bias. However, these methods helped little 
for geographic shifts, showing that there is more to robustness than texture 
bias alone. Regarding \emph{Only IID Accuracy Matters}, while IID accuracy 
is a strong predictor of OOD accuracy, it is not decisive---Table~\ref{tab:hypothesissummary} shows that many methods improve robustness across multiple distribution shifts, and recent experiments in NLP provide further counterexamples \parencite{hendrycks2020pretrained}. Finally, \emph{Synthetic $\notimplies$ Natural} has clear counterexamples given that DeepAugment greatly increases accuracy on ImageNet-R and Real Blurry Images. 
In summary, some previous hypotheses are implausible, and the Texture Bias hypothesis has the most support.

Our seven hypotheses presented several conflicting accounts of robustness. What led to this conflict?
We suspect it is because robustness is not one scalar like accuracy.
The research community is reasonable in judging IID accuracy with a \emph{univariate} metric like ImageNet classification accuracy, as models with higher ImageNet accuracy reliably have better fine-tuned classification accuracy on other tasks \parencite{Kornblith2018DoBI}.
In contrast, we argue 
it is too simplistic to judge OOD accuracy with a univariate metric like, say,  ImageNetV2 or ImageNet-C accuracy. Instead we hypothesize that robustness is multivariate.
This \emph{Multivariate} hypothesis means that there is not a single scalar model property that wholly governs natural model robustness. 

If robustness has many faces, future work should evaluate robustness using many distribution shifts; for example, ImageNet models should at least be tested against ImageNet-C and ImageNet-R.
Future work could further characterize the space of distribution shifts. However there are now more out-of-distribution robustness datasets than there are published robustness methods. Hence the research community should prioritize creating new robustness methods. If our \emph{Multivariate} hypothesis is true, multiple tests are necessary to develop models that are both robust and safe.\looseness=-1
\chapter{Multi-label Out-Of-Distribution Detection} \label{ch:multilabelood}

\section{Overview}

Research in multi-label classification has been in the shadow of multi-class classification for the greater part of a decade \parencite{tidake2018multilabelsurvey}.  This lack of focus is not entirely unjustified though, as many of the improvements in algorithm and model design have carried over to multi-label classification \parencite{resnet, chen2019gnn_multi_label, benbaruch2020asymmetric_multilabel_loss}.  While improvements have been made there still remains a considerable gap between multi-class performance and multi-label performance. 

This gap is even more pronounced when considering the difference in robustness difference from multi-class to multi-label.  Previous work in robustness focused on small whole-image anomaly detection with surprisingly no research studying the multi-label setting.  In the previous two chapters, we have focused on scaling up to large-scale datasets which presents unique challenges such as a plethora of fine-grained object classes, see Chatper \ref{ch:anom_segmentation} and Chapter \ref{ch:neural_aug}.  In this chapter, we demonstrate that the maximum softmax probability (MSP) detector, a state-of-the-art method for small-scale problems, does not scale well to these challenging conditions. Moreover, in the multi-label setting the MSP detector cannot naturally be applied in the first place, as it requires softmax probabilities.    

Due to the limitations of the MSP, we modified it for use in the multi-label setting to the maximum logit which we covered in greater detail in Chapter \ref{ch:anom_segmentation}.  We also introduce a new technique that takes into account the correlations among the labels which is able to achieve comparable performance.

\section{Related Work}

Natural images often contain many objects of interest with complex relationships of co-occurrence. Multi-label image classification acknowledges this more realistic setting by allowing each image to have multiple overlapping labels. This problem has long been of interest \parencite{Everingham2009ThePV_pascal}, and recent web-scale multi-label datasets demonstrate its growing importance, including Tencent ML-Images \parencite{tencent-ml-images-2019} and Open Images \parencite{OpenImages}. 

There are a few distinct techniques that have made progress in the multi-label task beyond architectural improvements. One of the earlier techniques is combining recurrent neural networks with convolutional neural networks (CNNs) \parencite{jinseok2017multilabel_rnn}. The output of the CNN is fed into the RNN as a sequence to sequence task akin to language translation. 
Most recently graph neural networks (GNNs) have been used after the output of a CNN to learn the label dependencies \parencite{chen2019gnn_multi_label}. Others have expanded on utilizing GNNs by combining them with word embeddings \parencite{wang2019multilabel_word_embeddings}.  

Prior work addresses multi-label classification in various ways, such as by leveraging label dependencies \parencite{wang2016cnn}. While current work on out-of-distribution detection solely considers multi-class or unsupervised settings. Yet as classifiers learn to classify more objects and process larger images, the multi-label formulation becomes increasingly natural. To our knowledge, this problem setting has yet to be explored. We provide a baselines and evaluation setup.

\section{Methods}

\noindent\textbf{Datasets.}\quad
For multi-label classification we use the datasets PASCAL VOC \parencite{Everingham2009ThePV_pascal}, MS-COCO \parencite{coco}, and DeepFashion Remixed (DFR) \parencite{ge2019deepfashion2}.  Specifically for MS-COCO and PASCAL VOC, we evaluate the models trained on these datasets, by using 20 out-of-distribution classes from ImageNet-22K. These classes have no overlap with ImageNet-1K, PASCAL VOC, or MS-COCO. The 20 classes are chosen not to overlap with ImageNet-1K since the multi-label classifiers models are pre-trained on ImageNet-1K. We list the class WordNet IDs in the Supplementary Materials \ref{app:dfr_classes}.



In PASCAL VOC and MS-COCO both datasets are subject to changes in day-to-day camera operation which can cause shifts in attributes such as object size, object occlusion, camera viewpoint, and camera zoom. To measure this effect, we repurpose DeepFashion2 \parencite{ge2019deepfashion2} to create the DeepFashion Remixed (DFR) dataset.  
We designate a training set with 48K images and create eight out-of-distribution test sets to measure performance under shifts in object size, object occlusion, camera viewpoint, and camera zoom-in. DeepFashion Remixed is a multi-label classification task since images may contain more than one clothing item per image.  In this way we can control for changes in those attributes.

\noindent\textbf{Architecture.}\quad For our experiments we use a ResNet-101 backbone architecture pre-trained on ImageNet-1K. We replace the final layer with a fully connected layers and apply the logistic sigmoid function for multi-label prediction. 

$$\mathcal{L} = \sum\limits_i (y_i - ln(sigmoid( (logits)_i )) + (1 - y_i) \cdot ln(1 - sigmoid( (logits)_i ))$$

Where $y_i$ is a d-dimensional vector of {0,1} where 1 corresponds to the presence of that class. Note that it is not a one-hot vector so the entire vector can be all ones potentially.
We train each model for 50 epochs using the Adam optimizer \parencite{adam} with hyperparameter values $10^{-4}$ and $10^{-5}$ for $\beta_1$ and $\beta_2$\, respectively. For data augmentation we use standard resizing, random crops, and random flips to obtain images of size $256 \times 256 \times 3$. As a result of this training procedure, the mAP of the ResNet-101 on PASCAL VOC is 89.11\% and 72.0\% for MS-COCO.

For experiments on DFR data augmentation includes: a crop of random size in the (0.5 to 2.0) of the original size and a random aspect ratio of $3/4$ to $4/3$ of
the original aspect ratio, which is finally resized to create a $256 \times 256$ image. For data augmentation we randomly horizontally flip the image with probability 0.5.

\noindent\textbf{Detection Methods.}\quad We evaluate the trained MS-COCO and PASCAL VOC models using four different detectors described below.  Even though the models are multi-label detectors because we are feeding in single class images from Imagenet-22K we should expect all of the logits from the network to be low or zero.  For the descriptions of the dectectors below the ``logits'' refers to the aggregate vector composed of the prediction for each class.  Results are in \ref{tab:oodmultilabel}.

\begin{itemize}
  \item MaxLogit \label{MaxLogit_def} denotes taking the negative of the maximum value of a logits vector as the anomaly score.  The logits are formed by combining all the scores from each class taken from the last layer of a neural network.

  \item LogitAvg is the negative of the average of the logits values taken from the last layer of the neural network.

  \item Isolation Forest \parencite{Liu2008IsolationF}, denoted by IForest, works by randomly partitioning the input space into half spaces to form a decision tree.  IForest needs a ``training step'' or setup phase before it can be used.  More specifically the algorithm is as follows: 
  
  Step 1) select a feature to split on.  \\
  Step 2) choose a random split between min and max range for feature. \\
  Step 3) repeat steps 1 and 2 until all elements are singletons.  \\
  Step 4) Repeat steps 1-3 to construct a new tree. \\
  
  The isolation score is evaluated based on the average distance to reach a terminal leaf from the trees in the ensemble.  We train our isolation forest using in-distribution validation data.  Note that to train the isolation forest ground truth labels of the images are not required only the knowledge that they are in-distribution.  The Isolation forest can thus be considered an unsupervised learning algorithm as it does not use the image labels.  Finally, we tried two approaches to construct our space used for the isolation forest.  The first approach consisted of the aggregated logits vectors and the second approach consisted of using the maximum logit value.  So in the first approach the space is a d-dimensional space where d equals the number of classes.  The second approach consists of a 1-dimensional line based on the maximum logit possible from each image.  See the MaxLogit definition \ref{MaxLogit_def}.  We found that the second approach worked better and thus used that as our reported IForest values.  We use the default number of trees from \parencite{scikit-learn} which is that of 100 trees for the ensemble.
  
  Mathematically this works out to
  
  $$S(x_i) = -0.5 - \sum_j 2^{(
            -\text{depth}_j
            / (\#\text{trees} \enskip
               \cdot \enskip average\_path\_length_j))} $$
  \[
        \text{Anomaly Score}(x_i)= 
    \begin{cases}
        \text{True},  & \text{if } S(x_i)\geq 0\\
        \text{False}, & \text{otherwise}
    \end{cases}
  \]
  
  where $S(x_i)$ is the score of the i'th element.  The 0.5 is the default offset as presented by the authors who created Isolation Forest \parencite{Liu2008IsolationF}.  The variable j indexes the tree where $\text{depth}_{ij}$ is the number of ancestors of $x_i$ in tree $j$.  $\#trees$ is the total number of trees, and average\_path\_length is the avererge path length to get to the leaf for the j'th tree.  Finally the anomaly score of an element is determined by if the score is greater than or equal to 0.
  
  \item Local outlier factor (LOF) \parencite{lof}, computes a ratio of the local density between every element and the local density of its neighbors.  The algorithm works as follows.  We shall consider a point $A$ in the set and the points $B$ are elements of the k-Nearest Neighbors of A.  We first compute local reachability density of a point $A$ by taking the sum of the max of (distance of $A$ to $B$ and the distance of $B$ to its kth nearest neighbor) and finally dividing the resulting sum by k.  Given the the local reachability density (lrd) of $A$  we compute the lrd of $A's$ $k$ neighbors $B$ and take the ratio of lrd($B$)/(lrd($A$) * k) to give us the LOF.  $k$ is a hyper-parameter that needs to be set for up to which nearest neighbor to consider.  Here we set the number of neighbors considered to be $20$ as the default from \cite{scikit-learn}.  
  
  Similar to IForest we computed this method for both logits and maximum logit and reported the best result of the two, which turned out to be maximum logit.  Finally a value of $\leq$ 1 is considered and inlier while a value of $>$ 1 is considered an outlier.
  
  \item Typicality Score, computes how similar the set of output probabilities over all classes are to the average posterior distribution for a given set of classes.  To construct the typicality matrix we set a threshold $t$ and whenever a class probability exceeds $t$ we add the probability distribution to the typicality matrix corresponding to that class `c'.  In other words, if the posterior probability for label `c' is greater than 50\%, we add the entire probability distribution to entry `c'.  We repeat this process for every image in a validation set and finally normalize each row of the typicality matrix.  To test the typicality or get an anomaly score we apply a similar approach for each test image.  If class `c' of a test image exceeds threshold $t$ we compute the distance of the current output to the typical output of class `c'.  We repeat this process for each class in the output that exceeds the threshold and take the sum of the outputs to get our anomalous score.  We experimented with the thresholds as the $t$ used for construction can be different from $t$ used for evaluation but found it to only vary the results slightly giving extra added complexity but little benefit.  Note that this method does not require labels only the knowledge that a set of examples are in-distribution.
  
  We interpret the typicality matrix as a course measure of what is the probability to see other classes given the presence (or belief) of class `c'.  It is possible to construct the matrix from actual data labels as opposed to the model's output, however we found that produce inferior results compared to using the output class probabilities.  The final resulting matrix is of dimensions c by c. A row corresponds to the normalized sum of the output probabilities of all images of class `c' that the model outputted a belief that class is present.  
  
  \[
        \text{row}_i= 
    \begin{cases}
        \sum p / \sum\limits_j p_j,  & \forall p_i\geq 0.5\\
        \frac{1}{n}, & \text{if} \enskip \forall p_i < 0.5
     \end{cases}
  \]
  
  Mathematically the matrix is constructed as follows. $p$ corresponds to a concentation of all of the probabilities per class.  The i'th row is the sum of all concatenated probabilities where the i'th class has greater than 50\% probability.  After the summation the resulting row is normalized. If there are no such instances where that is true then the row defaults to a uniform probability distribution over all classes. 

  \item Non-typicality Score computes an addionational matrix for dissimilarity.  This score builds of off the previous Typicality Score, with an added non-typicality matrix that defines what the distribution of what objects looks like given the absence of label `c'.  This can be used in conjunction with the typicality matrix to add or subtract to the previous values. However, the addition of this matrix yielded slightly worse results, so we removed it for the final version.  
\end{itemize}



\paragraph{Data Collection.}

Similar to SVSF in section \ref{ch:neuraug_sec:svsf}, we fix one value for each of the four metadata attributes in the training distribution.
Specifically, the DFR training set contains images with medium scale, medium occlusion, side/back viewpoint, and no zoom-in.
After sampling an IID test set, we construct eight OOD test distributions by altering one attribute at a time, obtaining test sets with minimal and heavy occlusion; small and large scale; frontal and not-worn viewpoints; and medium and large zoom-in.
Including the in-distribution test set, this gives us a total of nine test sets.
See Appendix \ref{app:classes} for details on test set sizes.
Since DeepFashion Remixed is a multi-label classification task, we use sigmoid outputs.
To measure performance, we calculate mAP (mean Average Precision) as a frequency-weighted average over all 13 class AP scores.

\section{Results}

\begin{table*}
	\centering
	\begin{tabularx}{1\textwidth}{ *{1}{>{\hsize=.5\hsize}X} | *{5}{>{\hsize=0.5\hsize}Y} | *{5}{>{\hsize=0.5\hsize}Y}  }
		\multicolumn{1}{c}{} & \multicolumn{5}{c}{FPR95 $\downarrow$} \\ \cline{2-6}
		\multicolumn{1}{l|}{$\mathcal{D}_\text{in}$} & IForest & LogitAvg & LOF & MaxLogit & Typical \\ \hline
		VOC           & 98.6 & 98.2 & 84.0 & 35.6 & \textbf{28.1} \\
		COCO          & 95.6 & 94.5 & 78.4 & 40.4 & \textbf{39.7} \\
		\hline
		\multicolumn{1}{c}{} & \multicolumn{5}{c}{AUROC $\uparrow$} \\ \cline{2-6}
		\multicolumn{1}{l|}{$\mathcal{D}_\text{in}$} & IForest & LogitAvg & LOF & MaxLogit & Typical \\ \hline
		VOC           & 46.3 & 47.9 & 68.4 & \textbf{90.9} & 88.1 \\
		COCO          & 41.4 & 55.5 & 70.2 & \textbf{90.3} & 88.7 \\
		\Xhline{3\arrayrulewidth}
	\end{tabularx}
	\caption{Multi-label out-of-distribution detection comparison of the maximum logit, typicality matrix, logit average, Local Outlier Factor, and Isolation Forest anomaly detectors on PASCAL VOC and MS-COCO. The same network architecture is used for all three detectors. All results shown are percentages.}
	\label{tab:oodmultilabel}
	\vspace{5mm}
\end{table*}

Results are shown in table \ref{tab:oodmultilabel}. We observe that the MaxLogit method outperforms the average logit and LOF by a significant margin. The MaxLogit method bears similarity to the MSP baseline \parencite{hendrycks2017baseline}, but is naturally applicable to multi-label problems. Indeed, forcing a softmax output on the multi-label logits in order to use MSP detector results in a 19.6\% drop in AUROC on MS-COCO. These results establish the MaxLogit as an effective and natural baseline for large-scale multi-label problems. Further, the evaluation setup enables future work in out-of-distribution detection with multi-label datasets.


\begin{table}[h]
\footnotesize
\begin{center}
{\setlength\tabcolsep{2pt}%
\begin{tabular}{@{}l | c | c | c c | c c | c c | c c}
\multicolumn{3}{c}{} & \multicolumn{2}{c}{Size} & \multicolumn{2}{c}{Occlusion} & \multicolumn{2}{c}{Viewpoint} & \multicolumn{2}{c}{Zoom} \\
\hline
Network & \multicolumn{1}{c|}{\,IID\,} & \multicolumn{1}{c|}{\,OOD\,}
& Small & Large 
& Slight/None & Heavy
& No Wear & Side/Back
& Medium & Large
    \\ \hline 
ResNet-50	&	77.6    & 55.1	&	39.4	&	73.0	&	51.5	&	41.2	&	50.5	&	63.2	&	48.7	&	73.3	\\
+ ImageNet-21K \emph{Pretraining} &  80.8   & 58.3  & 40.0  & 73.6  &  55.2 & 43.0  &  63.0 & 67.3  &  50.5 &  73.9 \\
+ SE (\emph{Self-Attention})  & 77.4  & 55.3 &  38.9 & 72.7 & 52.1  & 40.9  & 52.9  & 64.2  & 47.8  & 72.8 \\
+ Random Erasure    & 78.9  &  56.4 &  39.9  &  75.0 &  52.5 &  42.6 &  53.4 &  66.0 & 48.8  &  73.4 \\
+ Speckle Noise	&	78.9	& 55.8  &	38.4	&  74.0 	&	52.6	&	40.8	&	55.7	&	63.8	&	47.8	&	73.6	\\
+ Style Transfer	&	80.2	&   57.1   &	37.6	&	76.5	&	54.6	&	43.2	&	58.4	&	65.1	&	49.2	&	72.5	\\
+ DeepAugment	&	79.7	&   56.3   &	38.3	&	74.5	&	52.6	&	42.8	&	54.6	&	65.5	&	49.5	&	72.7	\\
+ AugMix	&	80.4	&   57.3   &	39.4	&	74.8	&	55.3	&	42.8	&	57.3	&	66.6	&	49.0	&	73.1	\\ \midrule
ResNet-152 (\emph{Larger Models})  & 80.0  &    57.1   & 40.0  & 75.6 & 52.3  & 42.0 &  57.7 &  65.6 &  48.9 & 74.4  \\
\Xhline{2\arrayrulewidth}
\end{tabular}}
\end{center}
\caption{DeepFashion Remixed results. Unlike the previous tables, higher is better since all values are mAP scores for this multi-label classification benchmark. The ``OOD'' column is the average of the row's rightmost eight OOD values. All techniques do little to close the IID/OOD generalization gap. 
}
\vspace{5pt}
\label{tab:deepfashion}
\end{table}

\begin{table}
\small
\begin{center}
{\setlength\tabcolsep{3pt}%
\begin{tabular}{@{}l | c | c c | c c | c c | c c}
\multicolumn{2}{c}{} & \multicolumn{2}{c}{Scale} & \multicolumn{2}{c}{Occlusion} & \multicolumn{2}{c}{Viewpoint} & \multicolumn{2}{c}{Zoom} \\
\hline
Network & \multicolumn{1}{c|}{\,IID\,}
& Small & Large 
& Slight/None & Heavy
& No Wear & Side/Back
& Medium & Large
    \\ \hline 
ResNet-101	&	78.7	&	38.0	&	75.2	&	53.7	&	42.9	&	60.7	&	65.4	&	49.8	&	74.0	\\
+ ImageNet-21K \emph{Pretraining} & 80.7   & 38.3  & 74.2  & 51.5  & 43.3  & 59.2  & 68.5  & 50.6  & 73.0  \\
+ CBAM (\emph{Self-Attention})              & 80.9  & 41.0  & 75.8  & 53.0  & 43.9  & 67.3  & 66.3  & 50.7  & 74.9  \\
+ Random Erasure    & 80.1  & 37.4  & 77.6  & 54.8  & 43.7  & 64.9  & 67.5  & 50.4  & 75.4  \\
+ Speckle Noise	&	79.8	&	38.0	&	73.5	&	51.1	&	43.0	&	63.2	&	65.0	&	49.9	&	73.9	\\
+ Style Transfer	&	81.7	&	39.2	&	75.4	&	54.9	&	43.5	&	64.5	&	66.4	&	51.5	&	73.8	\\
+ DeepAugment	&	81.3	&	38.4	&	74.7	&	53.3	&	43.3	&	63.2	&	65.9	&	51.0	&	75.3	\\
+ AugMix	&	81.8	&	40.2	&	74.5	&	52.5	&	42.4	&	65.0	&	65.9	&	51.1	&	74.3	\\ \midrule
ResNet-152 (\emph{Larger Models})  & 81.0  & 39.7  & 73.5  & 51.2  & 44.2  & 65.1  & 66.1  & 50.3  & 74.1  \\
    
\Xhline{2\arrayrulewidth}
\end{tabular}}
\end{center}
\caption{For these results we took a different partitioning of the DeepFashion Remixed dataset namely using all of the training data (excluding the same combinations we test against), and then only splitting up the validation data.  All values are mAP scores. ResNet-152 tests the \emph{Larger Models} hypothesis, ImageNet-21K Pretraining tests \emph{Pretraining}, CBAM tests \emph{Self-Attention}, and the other techniques test \emph{Diverse Data Augmentation}. All techniques have limited effects.
}
\label{tab:deepfashionResNet101}
\end{table}

\noindent \textbf{DeepFashion Remixed.}\quad \label{ch:multilabel_sec:DFR}
Table \ref{tab:deepfashion} shows our experimental findings on DFR, in which all evaluated methods have an average OOD mAP that is close to the baseline. In fact, most OOD mAP increases track IID mAP increases. In general, DFR's size and occlusion shifts hurt performance the most. We also evaluate with Random Erasure augmentation, which deletes rectangles within the image, to simulate occlusion \parencite{Zhong2017RandomED}.
Random Erasure improved occlusion performance, but Style Transfer helped even more.
Nothing substantially improved OOD performance beyond what is explained by IID performance, so here it would appear that \emph{Only IID Accuracy Matters}.
Our results do not provide clear evidence for the \emph{Larger Models}, \emph{Self-Attention}, \emph{Diverse Data Augmentation}, and \emph{Pretraining} hypotheses as discussed in chpater \ref{ch:neural_aug}.




\section{Conclusion}

We have evaluated several classic techniques and introduced two new techniques for measuring the out-of-distribution robustness in the multi-label setting.  We show that both are able achieve comparable results and extend the previous baseline of maximum softmax probability to better handle diverse and complex images.  


\chapter{Robustness in Few-Shot Learning}\label{ch:fs_robustness}

\section{Overview}

Few-shot learning has gained recent popularity with the advent of novel meta-learning techniques \parencite{finn2017MAML}.  Few-shot learning is the problem of training a classification model on a task given a small number of training examples (the so-called ``shots'').  Due to the difficulty in generalizing from only a few instances, and to be robust to overfitting, a successful few-shot learning model must efficiently re-use what it has learned.  

Given the goal of efficient reusability of learned material, we can see a clear connection between few-shot learning, and the aims of robustness.  One of the aims in robustness research is to detect out-of-distribution (OOD) examples which in many cases the number of OOD examples greatly outnumbers that of in-distribution examples.  In this way, the problem of few-shot learning is simply a scaling down of the original problem of OOD detection.  

Previous formulations of robustness focused exclusively within the data-rich setting \parencite{kreo2018robust, tsipras2018robustness, Orhan2019RobustnessPO}.  However, in practice there are many naturally occurring scenarios which only have a few instances.  There many only exist a few instances present in either the training set, the anomalies, or even both.  An example of such a phenomenon is for detecting rare species of animals \parencite{weinstein2018visionspecies} or rare cosmic events \parencite{ackermann2018cosmicphysics}, whereby both events are rare and collecting more data may be cost prohibitive or physically impossible.  

We attempt to overcome these issues by reformulating the problem with meta-learning.  Specifically we look at previous meta-learning approaches such as Simple-Shot and ProtoNet, to compare to our own approach of set-membership. We hypothesize that reformulating the problem as a set-membership task would perform better than enforcing arbitrary classification during updates to the base learner.

Meta-learning still presents challenges as the interactions between the meta-algorithm and base-learners are opaque.  The extra hyperparameters and in some cases two optimizations create an extra hurdle to determine the optimal set of parameters.  

Through our experimentation, we fail to show how the reformulation of the task to set-membership benefits the problem.  We hypothesize that the task may be ill-suited to learning from few examples.  We experiment with different architectures, and optimization methods to arrive at our conclusion.    

\section{Set-Membership}

Set-membership has a rich history \parencite{Kosut1992SetmembershipIO, Gollamudi1998SetmembershipFA,Werner2001SetmembershipAP} and has broad applications in network protocol analysis, routing-table lookup, online traffic measurement, peer-to-peer systems, cooperative caching, firewall design, intrusion detection, bioinformatics, database QUERY processing, stream computing, and distributed storage systems \parencite{Broder2003BloomFiltersSurvey}.  

Set-membership is the task of determining if a query element is a member of some set.  Given the simple definition there are surprisingly few methods for the task.  Given sufficient memory (and time) then the problem is solvable with hash-table complete dictionaries.  Once memory is a constraint then approximations to the full solution need to be considered. The methods for approximation are hash compacted hash-table dictionaries, Bloom filters, and derivatives of Bloom filters.

\section{Methods}

\begin{figure}
	\centering
      \includegraphics[width=\linewidth]{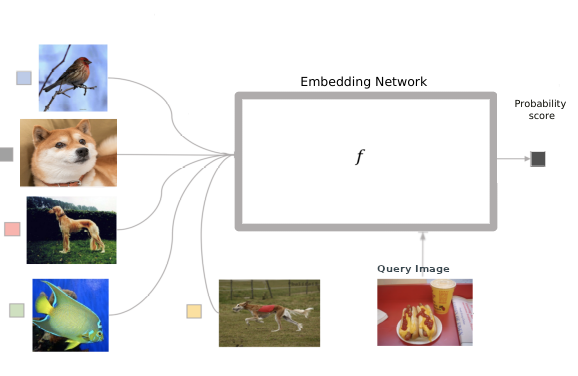}
  \caption{Set membership model overview. We concatenate the shot images with the single query images to feed into the network.  We use different architectures as our embedding network to learn whether the query image belongs in the set of images.  This is an example of a 5 shot (k=5) set membership task in which we ask if the hot dog belongs to the set of images of animals.
  }
  \label{fig:set_membership_model}
\end{figure}

\textbf{Datasets.}
The miniImageNet dataset \parencite{Vinyals2016miniimagenet} is a subset of the popularly used ImageNet \parencite{imagenet} dataset. The dataset contains 100 classes and has a total of 600
examples per class.  We follow \parencite{Ravi2017OptimizationAA} and the subsequent work to split the dataset into 64 base classes, 16 validation
classes, and 20 novel classes.  We pre-process the dataset as the original authors  \parencite{Vinyals2016miniimagenet} and subsequent studies do, by resizing the images to 84 × 84 pixels via rescaling and center cropping.

\textbf{Evaluation protocol.}
To evaluate our model and the others we compared to, we conduct 10,000 classification runs of K-shot C-way tasks from the novel classes.  Each task consists of selecting C of the novel classes, uses K labeled images, and 15 test images per class.  For our experiments we set $K = \{1,5\}$ for one-shot and five-shot experiments as per \cite{Vinyals2016miniimagenet,wang2019simpleshot}.  For the final accuracies we average over all of the tasks and test images to report the resulting average accuracy and 95\% confidence interval.  

\textbf{Model and Implementation details.}
We evaluate the following models for the set-membership problem. After we describe the neural network architectures we will describe the modifications we made to each to accomodate the new task.
We study the following five network architectures following \parencite{wang2019simpleshot}:

\textit{Conv-4}:  A four-layer convolutional neural network. We follow \cite{Vinyals2016miniimagenet} in their implementation. The implementation consists of 4 convolution blocks of 3 x 3 filter, followed by batch normalization, rectified linear unit, and max pooling.  The final layer is a linear projection to our prediction used for set-membership.  

\textit{MobileNet} \parencite{howard2017mobilenets}: We use the same architecture as published, but we remove the first two down-sampling layers from the network.

\textit{ResNet-10/18} \parencite{resnet}: We use the
standard 18-layer architecture but we remove the first
two down-sampling layers and we change the first convolutional layer to use a kernel of size 3 × 3 (rather
than 7 × 7) pixels. Our ResNet-10 contains 4 residual
blocks; the ResNet-18 contains 8 blocks.

\textit{WRN-28-10} Wide residual networks \parencite{wideresnet}: We use the architecture with 28 convolutional layers and a widening factor of 10.

\textit{DenseNet-121} \parencite{densenet}: We use the standard 121-layer architecture but similarly to MobileNet we remove
the first two down-sampling layers.

\begin{figure}
	\centering
      \includegraphics[width=\linewidth]{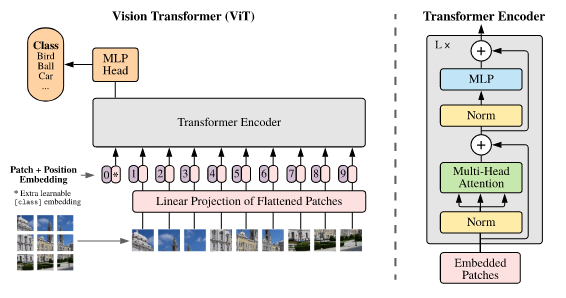}
  \caption{Vision Transformer model. The model takes as input non-overlapping patches of the original image whereby each are linearly embedded and position embeddings are added onto each. Finally the embeddings are fed into the Transformer along with a ``classification token'' to use for classification. The above reflects the original ViT.  We modified the architecture from instead of accepting only 1 image it accepts a batch of images as a single query.  The change corresponds to roughly increasing the model size by 6 to account for the extra 4 input images and the 1 query image.  The MLP head changes to output a single number instead of n class probabilities.  Image source: \cite{2021visiontransformer}.
  }
  \label{fig:vit_model}
\end{figure}

\textit{Vision Transformer} \parencite{2021visiontransformer}: We use the recent Vision Transformer with the following parameters: embedding dimensions of size 1024, 6 layers deep, 8 attention heads, and a logit layer dimensions of 2048 before the final linear layer to predict the binary class. The model is depicted in figure \ref{fig:vit_model}.

\textbf{Model training details.} We trained all of the networks for 90 epochs from scratch using stochastic gradient descent (SGD).  For all of the above networks we modified their final output layer to be a single logit which we treated as the set-membership score.  We use SGD to minimize binary cross entropy of the set-membership score with the labels $\{0, 1\}$ representing negative and positive membership respectively.  For SGD, we set the initial learning rate to 0.1 and decrease the learning rate by 10 at epochs 45 and 66 respectively.  We use a batch size of 200 images for all of our experiments.   We follow the data augmentation from \cite{resnet}, which is resize, scale, shift, and horizontal flip. 
Another modification we made to all of the above networks is modifying the input size by concatenating the shot and query images.  This would create a channel dimension of 6 and 18 for 1-shot and 5-shot respectively.  Finally we employed early stopping to select the best model.

\textbf{Experimental Setup.} For these experiments we utilize miniImageNet.  The training procedure is as follows: we sample a set of $k$ training examples these will serve as our key set (we use $k={1,5}$ for our experiments).  Given this set of $k$ examples from one class, sample a random image from the training set, this will serve as our query example.  The label is determined by whether the example belongs to the same class or different class.  During training any potential set of k examples from a given class can serve as key set.   

Once the key and query sets are selected from the training data, the resulting images are concatenated together along the channel dimension.  We modify the first layer of each network to accept a $(k+1)\cdot3$ channel image while maintaining the rest of the architecture constant.  The last layer of the networks are also modified to output a single probability of inclusion or exclusion.  We train for 600 iterations before retrieving a new set of keys and repeat the training procedures.  

At test time we are only allowed $k$ labeled examples per class as is commonly known as $k$-shot classification.  There are no common classes between the test time classes and those from training.  The $k$ samples are randomly chosen per class and all of the remaining examples serve as our query examples.  Finally we test all of the query examples to determine if they belong to key set.  We repeat the above procedure 10 times of choosing $k$ labeled examples and running every other query image against our network and use the average of the results as our reported result. We set $k=5$ for 5-shot and $k=1$ for 1-shot classification. 

We studied what effects does pretraining a model have in this setting.  Specifically we used MobileNet and ResNet-18 pretrained on Places365.  The reasoning for choosing this dataset to pretrain on is that it does not have any of the same classes as ImageNet so there are no conflicts with respect to the classes.

\section{Results}

\begin{table*}[ht]
\begin{center}
\begin{tabular}{lc}
\toprule
                Accuracy    & Mini ImageNet (\%) (both k=1 \& k=5)      \\ \midrule
Conv-4              & 50 \\
MobileNet           & 50 \\
ResNet-10           & 50 \\
ResNet-18           & 50 \\
WRN-28-10           & 50 \\
DenseNet-121        & 50 \\
Vision Transformer  & 50 \\ \midrule
Random Baseline     & 50 \\
\bottomrule
\end{tabular}
\end{center}
\caption{Accuracy of different architectures on miniImageNet \parencite{Vinyals2016miniimagenet}.  These results highlight that no architecture can achieve results than a random baseline.}
\label{tab:settraining_results}
\end{table*}

The results suggest that the approach implemented here utilizing neural networks for the task of set-membership needs improvement as it fails to generalize to new tasks.  Even with varying network architecture and optimization hyper-parameters such as SGD, Adam, weight decay and going from 1 to 5 shot seems to make no difference in generalizing from the training examples to test examples.  Furthermore even utilizing a different architecture, the transformer, results in a similar poor performance which leads to conclusion that the current approach is insensitive to architecture.
 
 \begin{table*}[ht]
\begin{center}
\begin{tabular}{lcc}
\toprule
                Accuracy    & Mini ImageNet (\%) k=1 & k=5      \\ \midrule
MobileNet           & 51 & 53 \\
ResNet-18           & 53 & 57 \\ \midrule
Random Baseline     & 50 & 50 \\
\bottomrule
\end{tabular}
\end{center}
\caption{Accuracy of different architectures on miniImageNet \parencite{Vinyals2016miniimagenet}.  These results that pretraining has some small effects on performance.}
\label{tab:settraining_results_pretrained}
\end{table*}

The results in table \ref{tab:settraining_results_pretrained} highlight how pretraining the models can improve the performance in this regime.  However the results are far below that of what the previous benchmarks and performance can achieve with prototypical networks \parencite{snell2017prototypical} for example.  This highlights how pretraining helps but this has also been demonstrated by prior work. 

\section{Conclusion}

We introduced a reformulation of a recent task of meta-learning into an older problem of set-membership.  However, it seems that this naive reinterpretation is unsuccessful in improving in the task.  We highlight some approaches and leave it as an open problem or as a warning to avoid the same pitfalls.  Our analysis shows that the reinterpretation of the task as a set-membership task is insensitive to model size, and optimization scheme.

\chapter{Conclusions}\label{ch:conclusion}

We have tested and presented several novel robustness techniques throughout the paper.  We began an exploration of few-shot robustness and demonstrated a how the task of set-membership as posed is ill-suited for robustness. We also have demonstrated a fragility in the loose definition of what robustness is. How in natural settings robustness is a multivariate concept that is not captured by a single metric even though the different concepts can be captured by a single word.

With the introduction of several techniques that scale better to larger images such as Maximum Logit, and Typicality Scores, we have helped the research field.  Deep Augment also presents a novel augmentation technique that creates an entirely new under explored area of neural based augmentations.  It remains difficult to still classify the types of augmentations learned and applied as they are data dependent augmentations.  

Finally we presented several new datasets to test out our methods and techniques.  Naturally Filtered Examples highlight the fragility of current models including that of transformers which is a completely different architecture type.  CAOS provides a synthetic test bed for reproducible anomaly detection.  Last but not least is Imagenet-R which is a representation version of a subset of ImageNet classes.  Together these contributions of datasets, and techniques will hopefully advance the area of robustness research in machine learning.  

\renewcommand\thechapter{Appendix A}
\makeatletter
\renewcommand{\@chapapp}{}
\makeatother
\chapter{}

\renewcommand\thechapter{A}

\section{\textsc{ImageNet-A} Classes}\label{app:imageneta-classes}
The 200 ImageNet classes that we selected for \textsc{ImageNet-A} are as follows.
goldfish, \quad 
great white shark, \quad 
hammerhead, \quad 
stingray, \quad 
hen, \quad 
ostrich, \quad 
goldfinch, \quad 
junco, \quad 
bald eagle, \quad 
vulture, \quad 
newt, \quad 
axolotl, \quad 
tree frog, \quad 
iguana, \quad 
African chameleon, \quad 
cobra, \quad 
scorpion, \quad 
tarantula, \quad 
centipede, \quad 
peacock, \quad 
lorikeet, \quad 
hummingbird, \quad 
toucan, \quad 
duck, \quad 
goose, \quad 
black swan, \quad 
koala, \quad 
jellyfish, \quad 
snail, \quad 
lobster, \quad 
hermit crab, \quad 
flamingo, \quad 
american egret, \quad 
pelican, \quad 
king penguin, \quad 
grey whale, \quad 
killer whale, \quad 
sea lion, \quad 
chihuahua, \quad 
shih tzu, \quad 
afghan hound, \quad 
basset hound, \quad 
beagle, \quad 
bloodhound, \quad 
italian greyhound, \quad 
whippet, \quad 
weimaraner, \quad 
yorkshire terrier, \quad 
boston terrier, \quad 
scottish terrier, \quad 
west highland white terrier, \quad 
golden retriever, \quad 
labrador retriever, \quad 
cocker spaniels, \quad 
collie, \quad 
border collie, \quad 
rottweiler, \quad 
german shepherd dog, \quad 
boxer, \quad 
french bulldog, \quad 
saint bernard, \quad 
husky, \quad 
dalmatian, \quad 
pug, \quad 
pomeranian, \quad 
chow chow, \quad 
pembroke welsh corgi, \quad 
toy poodle, \quad 
standard poodle, \quad 
timber wolf, \quad 
hyena, \quad 
red fox, \quad 
tabby cat, \quad 
leopard, \quad 
snow leopard, \quad 
lion, \quad 
tiger, \quad 
cheetah, \quad 
polar bear, \quad 
meerkat, \quad 
ladybug, \quad 
fly, \quad 
bee, \quad 
ant, \quad 
grasshopper, \quad 
cockroach, \quad 
mantis, \quad 
dragonfly, \quad 
monarch butterfly, \quad 
starfish, \quad 
wood rabbit, \quad 
porcupine, \quad 
fox squirrel, \quad 
beaver, \quad 
guinea pig, \quad 
zebra, \quad 
pig, \quad 
hippopotamus, \quad 
bison, \quad 
gazelle, \quad 
llama, \quad 
skunk, \quad 
badger, \quad 
orangutan, \quad 
gorilla, \quad 
chimpanzee, \quad 
gibbon, \quad 
baboon, \quad 
panda, \quad 
eel, \quad 
clown fish, \quad 
puffer fish, \quad 
accordion, \quad 
ambulance, \quad 
assault rifle, \quad 
backpack, \quad 
barn, \quad 
wheelbarrow, \quad 
basketball, \quad 
bathtub, \quad 
lighthouse, \quad 
beer glass, \quad 
binoculars, \quad 
birdhouse, \quad 
bow tie, \quad 
broom, \quad 
bucket, \quad 
cauldron, \quad 
candle, \quad 
cannon, \quad 
canoe, \quad 
carousel, \quad 
castle, \quad 
mobile phone, \quad 
cowboy hat, \quad 
electric guitar, \quad 
fire engine, \quad 
flute, \quad 
gasmask, \quad 
grand piano, \quad 
guillotine, \quad 
hammer, \quad 
harmonica, \quad 
harp, \quad 
hatchet, \quad 
jeep, \quad 
joystick, \quad 
lab coat, \quad 
lawn mower, \quad 
lipstick, \quad 
mailbox, \quad 
missile, \quad 
mitten, \quad 
parachute, \quad 
pickup truck, \quad 
pirate ship, \quad 
revolver, \quad 
rugby ball, \quad 
sandal, \quad 
saxophone, \quad 
school bus, \quad 
schooner, \quad 
shield, \quad 
soccer ball, \quad 
space shuttle, \quad 
spider web, \quad 
steam locomotive, \quad 
scarf, \quad 
submarine, \quad 
tank, \quad 
tennis ball, \quad 
tractor, \quad 
trombone, \quad 
vase, \quad 
violin, \quad 
military aircraft, \quad 
wine bottle, \quad 
ice cream, \quad 
bagel, \quad 
pretzel, \quad 
cheeseburger, \quad 
hotdog, \quad 
cabbage, \quad 
broccoli, \quad 
cucumber, \quad 
bell pepper, \quad 
mushroom, \quad 
Granny Smith, \quad 
strawberry, \quad 
lemon, \quad 
pineapple, \quad 
banana, \quad 
pomegranate, \quad 
pizza, \quad 
burrito, \quad 
espresso, \quad 
volcano, \quad 
baseball player, \quad 
scuba diver, \quad 
acorn, \quad 

n01443537, \quad 
n01484850, \quad 
n01494475, \quad 
n01498041, \quad 
n01514859, \quad 
n01518878, \quad 
n01531178, \quad 
n01534433, \quad 
n01614925, \quad 
n01616318, \quad 
n01630670, \quad 
n01632777, \quad 
n01644373, \quad 
n01677366, \quad 
n01694178, \quad 
n01748264, \quad 
n01770393, \quad 
n01774750, \quad 
n01784675, \quad 
n01806143, \quad 
n01820546, \quad 
n01833805, \quad 
n01843383, \quad 
n01847000, \quad 
n01855672, \quad 
n01860187, \quad 
n01882714, \quad 
n01910747, \quad 
n01944390, \quad 
n01983481, \quad 
n01986214, \quad 
n02007558, \quad 
n02009912, \quad 
n02051845, \quad 
n02056570, \quad 
n02066245, \quad 
n02071294, \quad 
n02077923, \quad 
n02085620, \quad 
n02086240, \quad 
n02088094, \quad 
n02088238, \quad 
n02088364, \quad 
n02088466, \quad 
n02091032, \quad 
n02091134, \quad 
n02092339, \quad 
n02094433, \quad 
n02096585, \quad 
n02097298, \quad 
n02098286, \quad 
n02099601, \quad 
n02099712, \quad 
n02102318, \quad 
n02106030, \quad 
n02106166, \quad 
n02106550, \quad 
n02106662, \quad 
n02108089, \quad 
n02108915, \quad 
n02109525, \quad 
n02110185, \quad 
n02110341, \quad 
n02110958, \quad 
n02112018, \quad 
n02112137, \quad 
n02113023, \quad 
n02113624, \quad 
n02113799, \quad 
n02114367, \quad 
n02117135, \quad 
n02119022, \quad 
n02123045, \quad 
n02128385, \quad 
n02128757, \quad 
n02129165, \quad 
n02129604, \quad 
n02130308, \quad 
n02134084, \quad 
n02138441, \quad 
n02165456, \quad 
n02190166, \quad 
n02206856, \quad 
n02219486, \quad 
n02226429, \quad 
n02233338, \quad 
n02236044, \quad 
n02268443, \quad 
n02279972, \quad 
n02317335, \quad 
n02325366, \quad 
n02346627, \quad 
n02356798, \quad 
n02363005, \quad 
n02364673, \quad 
n02391049, \quad 
n02395406, \quad 
n02398521, \quad 
n02410509, \quad 
n02423022, \quad 
n02437616, \quad 
n02445715, \quad 
n02447366, \quad 
n02480495, \quad 
n02480855, \quad 
n02481823, \quad 
n02483362, \quad 
n02486410, \quad 
n02510455, \quad 
n02526121, \quad 
n02607072, \quad 
n02655020, \quad 
n02672831, \quad 
n02701002, \quad 
n02749479, \quad 
n02769748, \quad 
n02793495, \quad 
n02797295, \quad 
n02802426, \quad 
n02808440, \quad 
n02814860, \quad 
n02823750, \quad 
n02841315, \quad 
n02843684, \quad 
n02883205, \quad 
n02906734, \quad 
n02909870, \quad 
n02939185, \quad 
n02948072, \quad 
n02950826, \quad 
n02951358, \quad 
n02966193, \quad 
n02980441, \quad 
n02992529, \quad 
n03124170, \quad 
n03272010, \quad 
n03345487, \quad 
n03372029, \quad 
n03424325, \quad 
n03452741, \quad 
n03467068, \quad 
n03481172, \quad 
n03494278, \quad 
n03495258, \quad 
n03498962, \quad 
n03594945, \quad 
n03602883, \quad 
n03630383, \quad 
n03649909, \quad 
n03676483, \quad 
n03710193, \quad 
n03773504, \quad 
n03775071, \quad 
n03888257, \quad 
n03930630, \quad 
n03947888, \quad 
n04086273, \quad 
n04118538, \quad 
n04133789, \quad 
n04141076, \quad 
n04146614, \quad 
n04147183, \quad 
n04192698, \quad 
n04254680, \quad 
n04266014, \quad 
n04275548, \quad 
n04310018, \quad 
n04325704, \quad 
n04347754, \quad 
n04389033, \quad 
n04409515, \quad 
n04465501, \quad 
n04487394, \quad 
n04522168, \quad 
n04536866, \quad 
n04552348, \quad 
n04591713, \quad 
n07614500, \quad 
n07693725, \quad 
n07695742, \quad 
n07697313, \quad 
n07697537, \quad 
n07714571, \quad 
n07714990, \quad 
n07718472, \quad 
n07720875, \quad 
n07734744, \quad 
n07742313, \quad 
n07745940, \quad 
n07749582, \quad 
n07753275, \quad 
n07753592, \quad 
n07768694, \quad 
n07873807, \quad 
n07880968, \quad 
n07920052, \quad 
n09472597, \quad 
n09835506, \quad 
n10565667, \quad 
n12267677, \quad 

\noindent`Stingray;' `goldfinch, Carduelis carduelis;' `junco, snowbird;' `robin, American robin, Turdus migratorius;' `jay;' `bald eagle, American eagle, Haliaeetus leucocephalus;' `vulture;' `eft;' `bullfrog, Rana catesbeiana;' `box turtle, box tortoise;' `common iguana, iguana, Iguana iguana;' `agama;' `African chameleon, Chamaeleo chamaeleon;' `American alligator, Alligator mississipiensis;' `garter snake, grass snake;' `harvestman, daddy longlegs, Phalangium opilio;' `scorpion;' `tarantula;' `centipede;' `sulphur-crested cockatoo, Kakatoe galerita, Cacatua galerita;' `lorikeet;' `hummingbird;' `toucan;' `drake;' `goose;' `koala, koala bear, kangaroo bear, native bear, Phascolarctos cinereus;' `jellyfish;' `sea anemone, anemone;' `flatworm, platyhelminth;' `snail;' `crayfish, crawfish, crawdad, crawdaddy;' `hermit crab;' `flamingo;' `American egret, great white heron, Egretta albus;' `oystercatcher, oyster catcher;' `pelican;' `sea lion;' `Chihuahua;' `golden retriever;' `Rottweiler;' `German shepherd, German shepherd dog, German police dog, alsatian;' `pug, pug-dog;' `red fox, Vulpes vulpes;' `Persian cat;' `lynx, catamount;' `lion, king of beasts, Panthera leo;' `American black bear, black bear, Ursus americanus, Euarctos americanus;' `mongoose;' `ladybug, ladybeetle, lady beetle, ladybird, ladybird beetle;' `rhinoceros beetle;' `weevil;' `fly;' `bee;' `ant, emmet, pismire;' `grasshopper, hopper;' `walking stick, walkingstick, stick insect;' `cockroach, roach;' `mantis, mantid;' `leafhopper;' `dragonfly, darning needle, devil's darning needle, sewing needle, snake feeder, snake doctor, mosquito hawk, skeeter hawk;' `monarch, monarch butterfly, milkweed butterfly, Danaus plexippus;' `cabbage butterfly;' `lycaenid, lycaenid butterfly;' `starfish, sea star;' `wood rabbit, cottontail, cottontail rabbit;' `porcupine, hedgehog;' `fox squirrel, eastern fox squirrel, Sciurus niger;' `marmot;' `bison;' `skunk, polecat, wood pussy;' `armadillo;' `baboon;' `capuchin, ringtail, Cebus capucinus;' `African elephant, Loxodonta africana;' `puffer, pufferfish, blowfish, globefish;' `academic gown, academic robe, judge's robe;' `accordion, piano accordion, squeeze box;' `acoustic guitar;' `airliner;' `ambulance;' `apron;' `balance beam, beam;' `balloon;' `banjo;' `barn;' `barrow, garden cart, lawn cart, wheelbarrow;' `basketball;' `beacon, lighthouse, beacon light, pharos;' `beaker;' `bikini, two-piece;' `bow;' `bow tie, bow-tie, bowtie;' `breastplate, aegis, egis;' `broom;' `candle, taper, wax light;' `canoe;' `castle;' `cello, violoncello;' `chain;' `chest;' `Christmas stocking;' `cowboy boot;' `cradle;' `dial telephone, dial phone;' `digital clock;' `doormat, welcome mat;' `drumstick;' `dumbbell;' `envelope;' `feather boa, boa;' `flagpole, flagstaff;' `forklift;' `fountain;' `garbage truck, dustcart;' `goblet;' `go-kart;' `golfcart, golf cart;' `grand piano, grand;' `hand blower, blow dryer, blow drier, hair dryer, hair drier;' `iron, smoothing iron;' `jack-o'-lantern;' `jeep, landrover;' `kimono;' `lighter, light, igniter, ignitor;' `limousine, limo;' `manhole cover;' `maraca;' `marimba, xylophone;' `mask;' `mitten;' `mosque;' `nail;' `obelisk;' `ocarina, sweet potato;' `organ, pipe organ;' `parachute, chute;' `parking meter;' `piggy bank, penny bank;' `pool table, billiard table, snooker table;' `puck, hockey puck;' `quill, quill pen;' `racket, racquet;' `reel;' `revolver, six-gun, six-shooter;' `rocking chair, rocker;' `rugby ball;' `saltshaker, salt shaker;' `sandal;' `sax, saxophone;' `school bus;' `schooner;' `sewing machine;' `shovel;' `sleeping bag;' `snowmobile;' `snowplow, snowplough;' `soap dispenser;' `spatula;' `spider web, spider's web;' `steam locomotive;' `stethoscope;' `studio couch, day bed;' `submarine, pigboat, sub, U-boat;' `sundial;' `suspension bridge;' `syringe;' `tank, army tank, armored combat vehicle, armoured combat vehicle;' `teddy, teddy bear;' `toaster;' `torch;' `tricycle, trike, velocipede;' `umbrella;' `unicycle, monocycle;' `viaduct;' `volleyball;' `washer, automatic washer, washing machine;' `water tower;' `wine bottle;' `wreck;' `guacamole;' `pretzel;' `cheeseburger;' `hotdog, hot dog, red hot;' `broccoli;' `cucumber, cuke;' `bell pepper;' `mushroom;' `lemon;' `banana;' `custard apple;' `pomegranate;' `carbonara;' `bubble;' `cliff, drop, drop-off;' `volcano;' `ballplayer, baseball player;' `rapeseed;' `yellow lady's slipper, yellow lady-slipper, Cypripedium calceolus, Cypripedium parviflorum;' `corn;' `acorn.'

Their WordNet IDs are as follows.

\noindent n01498041, \quad n01531178, \quad n01534433, \quad n01558993, \quad n01580077, \quad n01614925, \quad n01616318, \quad n01631663, \quad n01641577, \quad n01669191, \quad n01677366, \quad n01687978, \quad n01694178, \quad n01698640, \quad n01735189, \quad n01770081, \quad n01770393, \quad n01774750, \quad n01784675, \quad n01819313, \quad n01820546, \quad n01833805, \quad n01843383, \quad n01847000, \quad n01855672, \quad n01882714, \quad n01910747, \quad n01914609, \quad n01924916, \quad n01944390, \quad n01985128, \quad n01986214, \quad n02007558, \quad n02009912, \quad n02037110, \quad n02051845, \quad n02077923, \quad n02085620, \quad n02099601, \quad n02106550, \quad n02106662, \quad n02110958, \quad n02119022, \quad n02123394, \quad n02127052, \quad n02129165, \quad n02133161, \quad n02137549, \quad n02165456, \quad n02174001, \quad n02177972, \quad n02190166, \quad n02206856, \quad n02219486, \quad n02226429, \quad n02231487, \quad n02233338, \quad n02236044, \quad n02259212, \quad n02268443, \quad n02279972, \quad n02280649, \quad n02281787, \quad n02317335, \quad n02325366, \quad n02346627, \quad n02356798, \quad n02361337, \quad n02410509, \quad n02445715, \quad n02454379, \quad n02486410, \quad n02492035, \quad n02504458, \quad n02655020, \quad n02669723, \quad n02672831, \quad n02676566, \quad n02690373, \quad n02701002, \quad n02730930, \quad n02777292, \quad n02782093, \quad n02787622, \quad n02793495, \quad n02797295, \quad n02802426, \quad n02814860, \quad n02815834, \quad n02837789, \quad n02879718, \quad n02883205, \quad n02895154, \quad n02906734, \quad n02948072, \quad n02951358, \quad n02980441, \quad n02992211, \quad n02999410, \quad n03014705, \quad n03026506, \quad n03124043, \quad n03125729, \quad n03187595, \quad n03196217, \quad n03223299, \quad n03250847, \quad n03255030, \quad n03291819, \quad n03325584, \quad n03355925, \quad n03384352, \quad n03388043, \quad n03417042, \quad n03443371, \quad n03444034, \quad n03445924, \quad n03452741, \quad n03483316, \quad n03584829, \quad n03590841, \quad n03594945, \quad n03617480, \quad n03666591, \quad n03670208, \quad n03717622, \quad n03720891, \quad n03721384, \quad n03724870, \quad n03775071, \quad n03788195, \quad n03804744, \quad n03837869, \quad n03840681, \quad n03854065, \quad n03888257, \quad n03891332, \quad n03935335, \quad n03982430, \quad n04019541, \quad n04033901, \quad n04039381, \quad n04067472, \quad n04086273, \quad n04099969, \quad n04118538, \quad n04131690, \quad n04133789, \quad n04141076, \quad n04146614, \quad n04147183, \quad n04179913, \quad n04208210, \quad n04235860, \quad n04252077, \quad n04252225, \quad n04254120, \quad n04270147, \quad n04275548, \quad n04310018, \quad n04317175, \quad n04344873, \quad n04347754, \quad n04355338, \quad n04366367, \quad n04376876, \quad n04389033, \quad n04399382, \quad n04442312, \quad n04456115, \quad n04482393, \quad n04507155, \quad n04509417, \quad n04532670, \quad n04540053, \quad n04554684, \quad n04562935, \quad n04591713, \quad n04606251, \quad n07583066, \quad n07695742, \quad n07697313, \quad n07697537, \quad n07714990, \quad n07718472, \quad n07720875, \quad n07734744, \quad n07749582, \quad n07753592, \quad n07760859, \quad n07768694, \quad n07831146, \quad n09229709, \quad n09246464, \quad n09472597, \quad n09835506, \quad n11879895, \quad n12057211, \quad n12144580, \quad n12267677.

\section{\textsc{ImageNet-O} Classes}\label{app:imageneto-classes}

The 200 ImageNet classes that we selected for \textsc{ImageNet-O} are as follows.

\noindent`goldfish, Carassius auratus;' `triceratops;' `harvestman, daddy longlegs, Phalangium opilio;' `centipede;' `sulphur-crested cockatoo, Kakatoe galerita, Cacatua galerita;' `lorikeet;' `jellyfish;' `brain coral;' `chambered nautilus, pearly nautilus, nautilus;' `dugong, Dugong dugon;' `starfish, sea star;' `sea urchin;' `hog, pig, grunter, squealer, Sus scrofa;' `armadillo;' `rock beauty, Holocanthus tricolor;' `puffer, pufferfish, blowfish, globefish;' `abacus;' `accordion, piano accordion, squeeze box;' `apron;' `balance beam, beam;' `ballpoint, ballpoint pen, ballpen, Biro;' `Band Aid;' `banjo;' `barbershop;' `bath towel;' `bearskin, busby, shako;' `binoculars, field glasses, opera glasses;' `bolo tie, bolo, bola tie, bola;' `bottlecap;' `brassiere, bra, bandeau;' `broom;' `buckle;' `bulletproof vest;' `candle, taper, wax light;' `car mirror;' `chainlink fence;' `chain saw, chainsaw;' `chime, bell, gong;' `Christmas stocking;' `cinema, movie theater, movie theatre, movie house, picture palace;' `combination lock;' `corkscrew, bottle screw;' `crane;' `croquet ball;' `dam, dike, dyke;' `digital clock;' `dishrag, dishcloth;' `dogsled, dog sled, dog sleigh;' `doormat, welcome mat;' `drilling platform, offshore rig;' `electric fan, blower;' `envelope;' `espresso maker;' `face powder;' `feather boa, boa;' `fireboat;' `fire screen, fireguard;' `flute, transverse flute;' `folding chair;' `fountain;' `fountain pen;' `frying pan, frypan, skillet;' `golf ball;' `greenhouse, nursery, glasshouse;' `guillotine;' `hamper;' `hand blower, blow dryer, blow drier, hair dryer, hair drier;' `harmonica, mouth organ, harp, mouth harp;' `honeycomb;' `hourglass;' `iron, smoothing iron;' `jack-o'-lantern;' `jigsaw puzzle;' `joystick;' `lawn mower, mower;' `library;' `lighter, light, igniter, ignitor;' `lipstick, lip rouge;' `loupe, jeweler's loupe;' `magnetic compass;' `manhole cover;' `maraca;' `marimba, xylophone;' `mask;' `matchstick;' `maypole;' `maze, labyrinth;' `medicine chest, medicine cabinet;' `mortar;' `mosquito net;' `mousetrap;' `nail;' `neck brace;' `necklace;' `nipple;' `ocarina, sweet potato;' `oil filter;' `organ, pipe organ;' `oscilloscope, scope, cathode-ray oscilloscope, CRO;' `oxygen mask;' `paddlewheel, paddle wheel;' `panpipe, pandean pipe, syrinx;' `park bench;' `pencil sharpener;' `Petri dish;' `pick, plectrum, plectron;' `picket fence, paling;' `pill bottle;' `ping-pong ball;' `pinwheel;' `plate rack;' `plunger, plumber's helper;' `pool table, billiard table, snooker table;' `pot, flowerpot;' `power drill;' `prayer rug, prayer mat;' `prison, prison house;' `punching bag, punch bag, punching ball, punchball;' `quill, quill pen;' `radiator;' `reel;' `remote control, remote;' `rubber eraser, rubber, pencil eraser;' `rule, ruler;' `safe;' `safety pin;' `saltshaker, salt shaker;' `scale, weighing machine;' `screw;' `screwdriver;' `shoji;' `shopping cart;' `shower cap;' `shower curtain;' `ski;' `sleeping bag;' `slot, one-armed bandit;' `snowmobile;' `soap dispenser;' `solar dish, solar collector, solar furnace;' `space heater;' `spatula;' `spider web, spider's web;' `stove;' `strainer;' `stretcher;' `submarine, pigboat, sub, U-boat;' `swimming trunks, bathing trunks;' `swing;' `switch, electric switch, electrical switch;' `syringe;' `tennis ball;' `thatch, thatched roof;' `theater curtain, theatre curtain;' `thimble;' `throne;' `tile roof;' `toaster;' `tricycle, trike, velocipede;' `turnstile;' `umbrella;' `vending machine;' `waffle iron;' `washer, automatic washer, washing machine;' `water bottle;' `water tower;' `whistle;' `Windsor tie;' `wooden spoon;' `wool, woolen, woollen;' `crossword puzzle, crossword;' `traffic light, traffic signal, stoplight;' `ice lolly, lolly, lollipop, popsicle;' `bagel, beigel;' `pretzel;' `hotdog, hot dog, red hot;' `mashed potato;' `broccoli;' `cauliflower;' `zucchini, courgette;' `acorn squash;' `cucumber, cuke;' `bell pepper;' `Granny Smith;' `strawberry;' `orange;' `lemon;' `pineapple, ananas;' `banana;' `jackfruit, jak, jack;' `pomegranate;' `chocolate sauce, chocolate syrup;' `meat loaf, meatloaf;' `pizza, pizza pie;' `burrito;' `bubble;' `volcano;' `corn;' `acorn;' `hen-of-the-woods, hen of the woods, Polyporus frondosus, Grifola frondosa.'

Their WordNet IDs are as follows.

n01443537, \quad n01704323, \quad n01770081, \quad n01784675, \quad n01819313, \quad n01820546, \quad n01910747, \quad n01917289, \quad n01968897, \quad n02074367, \quad n02317335, \quad n02319095, \quad n02395406, \quad n02454379, \quad n02606052, \quad n02655020, \quad n02666196, \quad n02672831, \quad n02730930, \quad n02777292, \quad n02783161, \quad n02786058, \quad n02787622, \quad n02791270, \quad n02808304, \quad n02817516, \quad n02841315, \quad n02865351, \quad n02877765, \quad n02892767, \quad n02906734, \quad n02910353, \quad n02916936, \quad n02948072, \quad n02965783, \quad n03000134, \quad n03000684, \quad n03017168, \quad n03026506, \quad n03032252, \quad n03075370, \quad n03109150, \quad n03126707, \quad n03134739, \quad n03160309, \quad n03196217, \quad n03207743, \quad n03218198, \quad n03223299, \quad n03240683, \quad n03271574, \quad n03291819, \quad n03297495, \quad n03314780, \quad n03325584, \quad n03344393, \quad n03347037, \quad n03372029, \quad n03376595, \quad n03388043, \quad n03388183, \quad n03400231, \quad n03445777, \quad n03457902, \quad n03467068, \quad n03482405, \quad n03483316, \quad n03494278, \quad n03530642, \quad n03544143, \quad n03584829, \quad n03590841, \quad n03598930, \quad n03602883, \quad n03649909, \quad n03661043, \quad n03666591, \quad n03676483, \quad n03692522, \quad n03706229, \quad n03717622, \quad n03720891, \quad n03721384, \quad n03724870, \quad n03729826, \quad n03733131, \quad n03733281, \quad n03742115, \quad n03786901, \quad n03788365, \quad n03794056, \quad n03804744, \quad n03814639, \quad n03814906, \quad n03825788, \quad n03840681, \quad n03843555, \quad n03854065, \quad n03857828, \quad n03868863, \quad n03874293, \quad n03884397, \quad n03891251, \quad n03908714, \quad n03920288, \quad n03929660, \quad n03930313, \quad n03937543, \quad n03942813, \quad n03944341, \quad n03961711, \quad n03970156, \quad n03982430, \quad n03991062, \quad n03995372, \quad n03998194, \quad n04005630, \quad n04023962, \quad n04033901, \quad n04040759, \quad n04067472, \quad n04074963, \quad n04116512, \quad n04118776, \quad n04125021, \quad n04127249, \quad n04131690, \quad n04141975, \quad n04153751, \quad n04154565, \quad n04201297, \quad n04204347, \quad n04209133, \quad n04209239, \quad n04228054, \quad n04235860, \quad n04243546, \quad n04252077, \quad n04254120, \quad n04258138, \quad n04265275, \quad n04270147, \quad n04275548, \quad n04330267, \quad n04332243, \quad n04336792, \quad n04347754, \quad n04371430, \quad n04371774, \quad n04372370, \quad n04376876, \quad n04409515, \quad n04417672, \quad n04418357, \quad n04423845, \quad n04429376, \quad n04435653, \quad n04442312, \quad n04482393, \quad n04501370, \quad n04507155, \quad n04525305, \quad n04542943, \quad n04554684, \quad n04557648, \quad n04562935, \quad n04579432, \quad n04591157, \quad n04597913, \quad n04599235, \quad n06785654, \quad n06874185, \quad n07615774, \quad n07693725, \quad n07695742, \quad n07697537, \quad n07711569, \quad n07714990, \quad n07715103, \quad n07716358, \quad n07717410, \quad n07718472, \quad n07720875, \quad n07742313, \quad n07745940, \quad n07747607, \quad n07749582, \quad n07753275, \quad n07753592, \quad n07754684, \quad n07768694, \quad n07836838, \quad n07871810, \quad n07873807, \quad n07880968, \quad n09229709, \quad n09472597, \quad n12144580, \quad n12267677, \quad n13052670.

\newpage

\section{Expanded Results}

\subsection{Full Architecture Results}
Full results with various architectures are in \ref{tab:nae-fullarch}.
\begin{table*}
\begin{center}
\begin{tabular}{lcc}
\toprule
                        & ImageNet-A (Acc \%) & ImageNet-O (AUPR \%) \\\midrule
AlexNet                 & 1.77  & 15.44     \\
SqueezeNet1.1           & 1.12  & 15.31     \\
VGG16                   & 2.63  & 16.58     \\
VGG19                   & 2.11  & 16.80     \\
VGG19+BN                & 2.95  & 16.57     \\
DenseNet121             & 2.16  & 16.11     \\
\hline
ResNet-18               & 1.15  & 15.23     \\
ResNet-34               & 1.87  & 16.00     \\
ResNet-50               & 2.17  & 16.20     \\
ResNet-101              & 4.72  & 17.20     \\
ResNet-152              & 6.05  & 18.00     \\
\hline
ResNet-50+Squeeze-and-Excite            & 6.17  & 17.52     \\
ResNet-101+Squeeze-and-Excite           & 8.55  & 17.91     \\
ResNet-152+Squeeze-and-Excite           & 9.35  & 18.65     \\
\hline
Res2Net-50 (v1b)        & 14.59 & 19.50     \\
Res2Net-101 (v1b)       & 21.84 & 22.69     \\
Res2Net-152 (v1b)       & 22.4  & 23.90     \\
\hline
ResNeXt-50 ($32\times4$d)   & 4.81  & 17.60 \\
ResNeXt-101 ($32\times4$d)  & 5.85  & 19.60 \\
ResNeXt-101 ($32\times8$d)   & 10.2  & 20.51 \\
\hline
DPN 68                  & 3.53  & 17.78     \\
DPN 98                  & 9.15  & 21.10     \\
\bottomrule
\end{tabular}
\end{center}
\caption{Expanded \textsc{ImageNet-A} and \textsc{ImageNet-O} architecture results.}\label{tab:nae-fullarch}
\end{table*}

\subsection{Calibration}\label{app:calibration}
In this section we show \textsc{ImageNet-A} calibration results.

\begin{figure}
\vspace{-10pt}
\centering
	\includegraphics[width=0.16\textwidth]{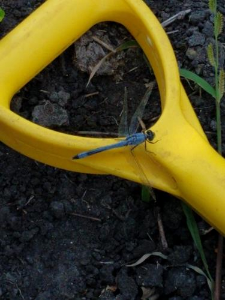}%
	\includegraphics[width=0.16\textwidth]{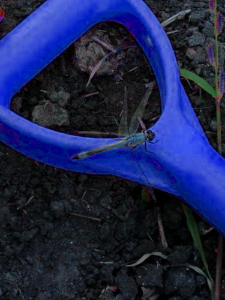}%
	\includegraphics[width=0.16\textwidth]{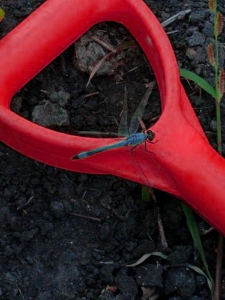}
	\caption{A demonstration of color sensitivity. While the leftmost image is classified as ``banana'' with high confidence, the images with modified color are correctly classified. Not only would we like models to be more accurate, we would like them to be calibrated if they wrong.}\label{fig:dragonfly}
	\vspace{-10pt}
\end{figure}

\noindent\textbf{Uncertainty Metrics.} \quad The \textit{$\ell_2$ Calibration Error} is how we measure miscalibration. We would like classifiers that can reliably forecast their accuracy. Concretely, we want classifiers which give examples 60\% confidence to be correct 60\% of the time. We judge a classifier's miscalibration with the $\ell_2$ Calibration Error \cite{kumar2019calibration}.


\begin{figure}[t]
	\centering
	\includegraphics[width=0.48\textwidth]{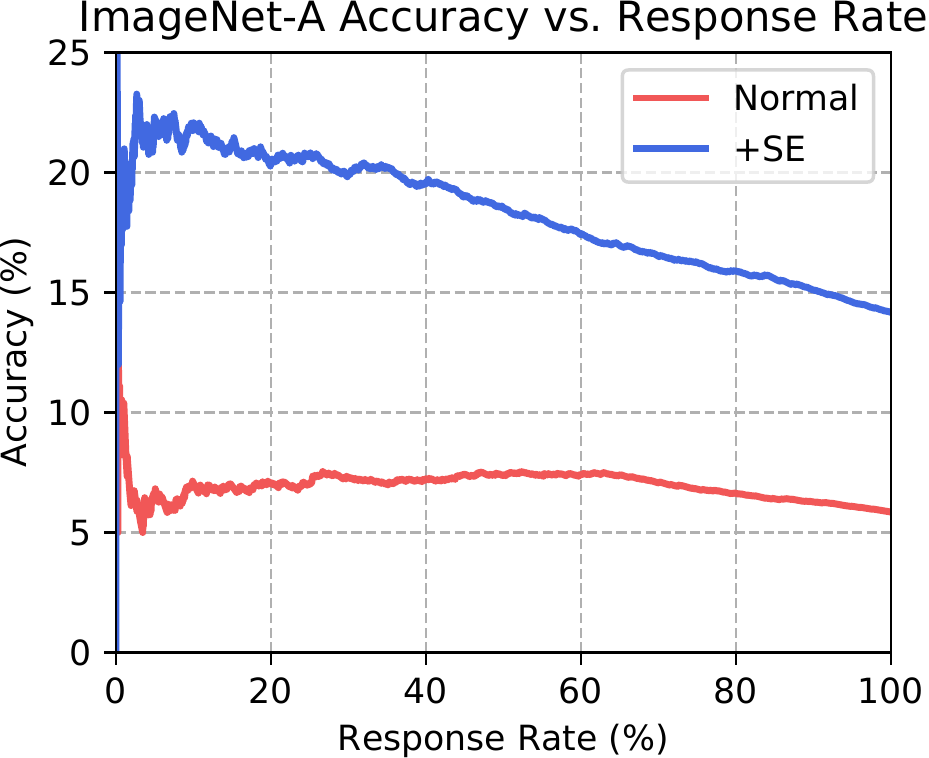}
	\caption{The Response Rate Accuracy curve for a ResNeXt-101 (32$\times$4d) with and without Squeeze-and-Excitation (SE). The Response Rate is the percent classified. The accuracy at a $n$\% response rate is the accuracy on the $n$\% of examples where the classifier is most confident.
	}
	\label{fig:rra}
\end{figure}

Our second uncertainty estimation metric is the \textit{Area Under the Response Rate Accuracy Curve (AURRA).} Responding only when confident is often preferable to predicting falsely.
In these experiments, we allow classifiers to respond to a subset of the test set and abstain from predicting the rest. Classifiers with quality uncertainty estimates should be capable identifying examples it is likely to predict falsely and abstain. If a classifier is required to abstain from predicting on 90\% of the test set, or equivalently respond to the remaining 10\% of the test set, then we should like the classifier's uncertainty estimates to separate correctly and falsely classified examples and have high accuracy on the selected 10\%. At a fixed response rate, we should like the accuracy to be as high as possible. At a 100\% response rate, the classifier accuracy is the usual test set accuracy. We vary the response rates and compute the corresponding accuracies to obtain the Response Rate Accuracy (RRA) curve. The area under the Response Rate Accuracy curve is the AURRA. To compute the AURRA in this paper, we use the maximum softmax probability. For response rate $p$, we take the $p$ fraction of examples with highest maximum softmax probability. If the response rate is 10\%, we select the top 10\% of examples with the highest confidence and compute the accuracy on these examples. An example RRA curve is in \ref{fig:rra} .\looseness=-1


\begin{figure}
\begin{subfigure}{.48\textwidth}
    \centering
    \includegraphics[width=\textwidth]{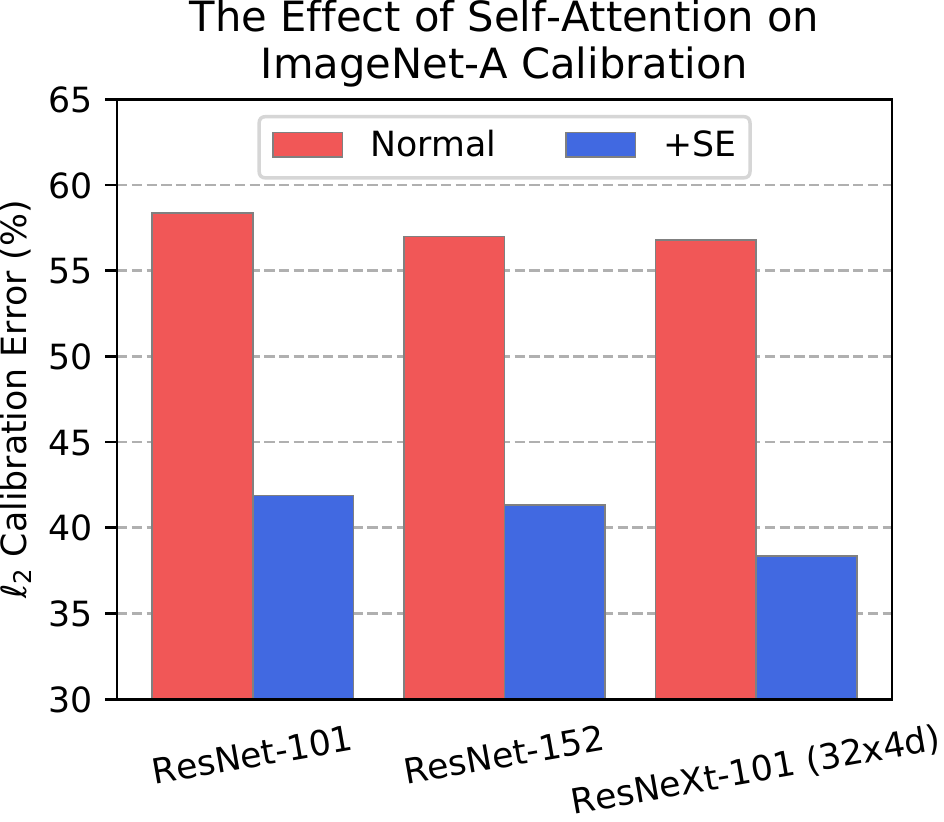}
\end{subfigure}
\begin{subfigure}{.48\textwidth}
    \centering
    \includegraphics[width=\textwidth]{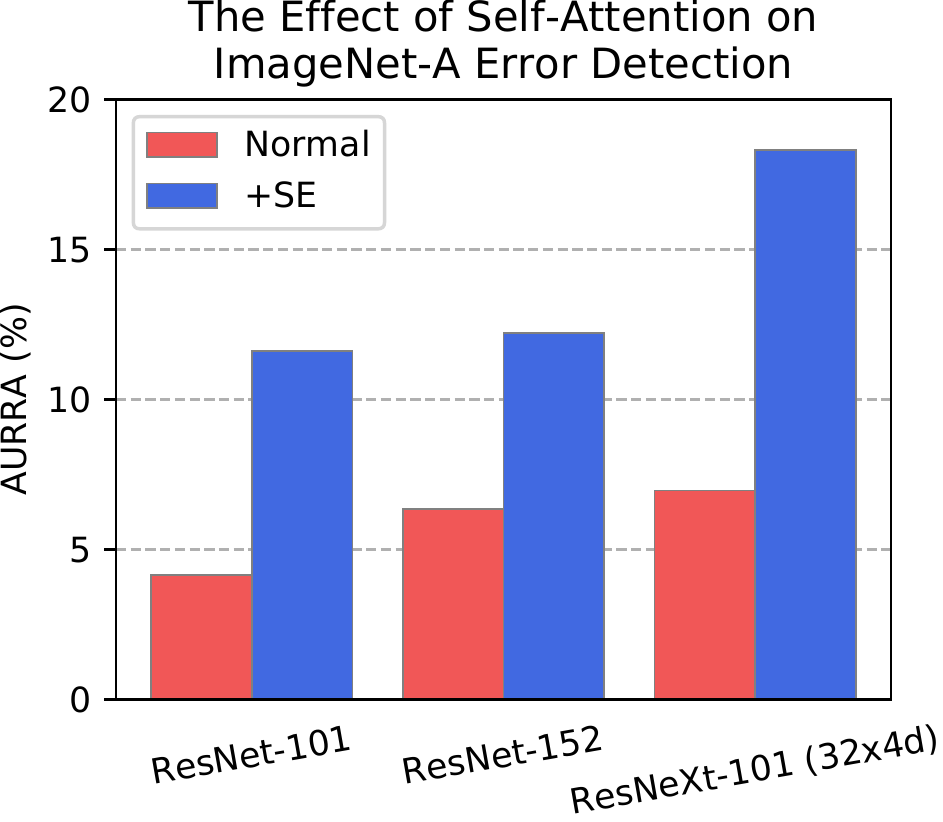}
\end{subfigure}%
\caption{Self-attention's influence on \textsc{ImageNet-A} $\ell_2$ calibration and error detection.}
\end{figure}


\begin{figure}
\begin{subfigure}{.48\textwidth}
    \centering
    \includegraphics[width=\textwidth]{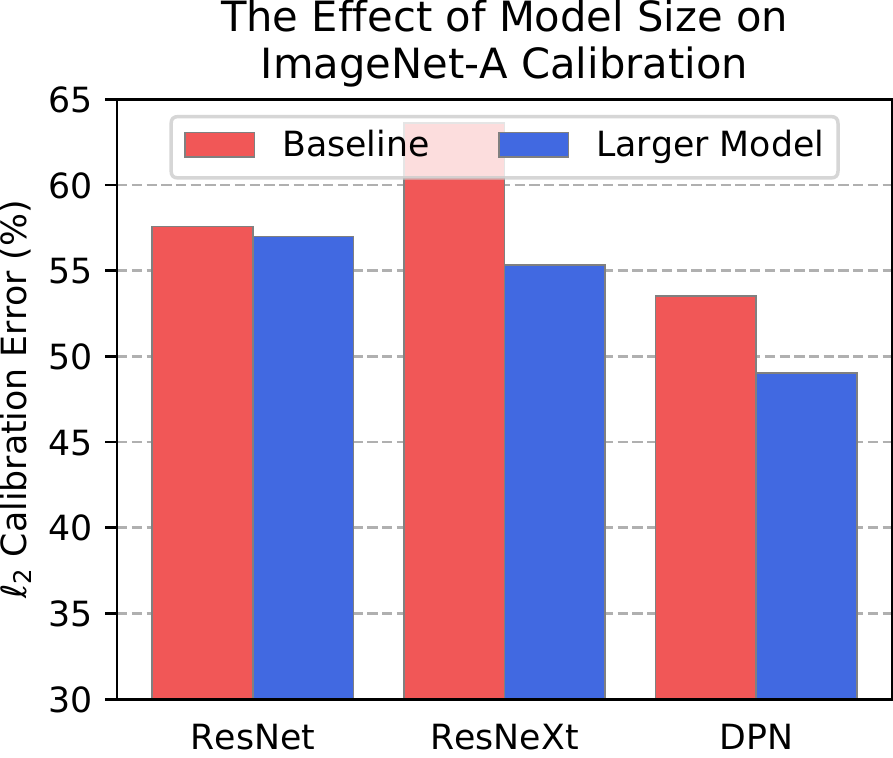}
\end{subfigure}
\begin{subfigure}{.48\textwidth}
    \centering
    \includegraphics[width=\textwidth]{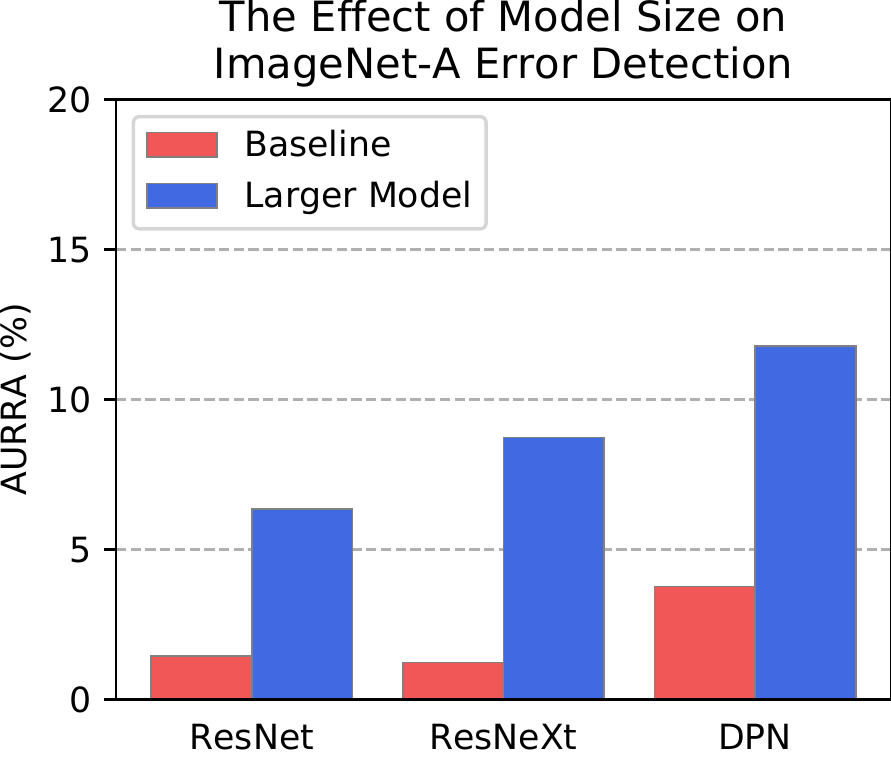}
\end{subfigure}%
\caption{Model size's influence on \textsc{ImageNet-A} $\ell_2$ calibration and error detection.}
\end{figure}

\section{Real Blurry Images and ImageNet-C}\label{app:blur}

\begin{figure}
\begin{center}
\includegraphics[width=\textwidth]{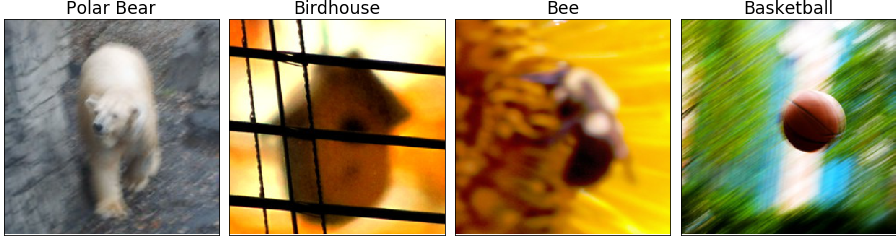}
\end{center}
\caption{
Examples of real-world blurry images from our collected dataset.
}\label{fig:blur}
\end{figure}

We collect 1,000 blurry images to see whether improvements on ImageNet-C's simulated blurs correspond to improvements on real-world blurry images. Each image belongs to an ImageNet class. Examples are in \ref{fig:blur}. Results from \ref{tab:imagenetblur} show that \emph{Larger Models}, \emph{Self-Attention}, \emph{Diverse Data Augmentation}, \emph{Pretraining} all help, just like ImageNet-C. Here DeepAugment+AugMix attains state-of-the-art. These results suggest ImageNet-C's simulated corruptions track real-world corruptions. In hindsight, this is expected since various computer vision problems have used synthetic corruptions as proxies for real-world corruptions, for decades. In short, ImageNet-C is a diverse and systematic benchmark that is correlated with improvements on real-world corruptions.

\begin{table}
\centering

\begin{tabularx}{\textwidth}{*{1}{>{\hsize=1.4\hsize}L} | *{4}{>{\hsize=0.25\hsize}Y} || *{1}{>{\hsize=0.6\hsize}Y} | *{1}{>{\hsize=0.6\hsize}Y}}
Network & Defocus Blur & Glass Blur & Motion Blur & Zoom Blur & {ImageNet-C Blur Mean} & Real Blurry Images
    \\ \hline 
ResNet-50	& 61 & 73 & 61 & 64 & 65 & 58.7 \\
+ ImageNet-21K \emph{Pretraining} & 56 & 69 & 53 & 59 & 59 & 54.8\\
+ CBAM (\emph{Self-Attention})  & 60 & 69 & 56 & 61 & 62 & 56.5 \\
+ $\ell_\infty$ Adversarial Training & 80 & 71 & 72 & 71 & 74 & 71.6 \\
+ Speckle Noise	& 57 & 68 & 60 & 64 & 62 & 56.9\\
+ Style Transfer	& 57 & 68 & 55 & 64 & 61 & 56.7\\
+ AugMix	& 52 & 65 & 46 & 51 & 54 & 54.4\\
+ DeepAugment	& 48 & 60 & 51 & 61 & 55 & 54.2\\
+ DeepAugment+AugMix	& 41 & 53 & 39 & 48 & 45 & 51.7 \\\midrule
ResNet-152 (\emph{Larger Models})  & 67 & 81 & 66 & 74 & 58 & 54.3 \\
\Xhline{2\arrayrulewidth}
\end{tabularx}
\caption{ImageNet-C vs Real Blurry Images. All values are error rates and percentages. The rank orderings of the models on Real Blurry Images are similar to the rank orderings for ``ImageNet-C Blur Mean,'' so ImageNet-C's simulated blurs track real-world blur performance. Hence synthetic image corruptions and real-world image corruptions are not loose and separate.}
\label{tab:imagenetblur}
\end{table}

\section{Additional Results}\label{app:additional}

\noindent \textbf{ImageNet-R.}\quad
Expanded ImageNet-R results are in \ref{tab:imagenetr_full}.


WSL pretraining on Instagram images appears to yield dramatic improvements on ImageNet-R, but the authors note the prevalence of artistic renditions of object classes on the Instagram platform. While ImageNet's data collection process actively excluded renditions, we do not have reason to believe the Instagram dataset excluded renditions. On a ResNeXt-101 32$\times$8d model, WSL pretraining improves ImageNet-R performance by a massive 37.5\% from 57.5\% top-1 error to 24.2\%. Ultimately, without examining the training images we are unable to determine whether ImageNet-R represents an actual distribution shift to the Instagram WSL models. However, we also observe that with greater controls, that is with ImageNet-21K pre-training, pretraining hardly helped ImageNet-R performance, so it is not clear that more pretraining data improves ImageNet-R performance.

\begin{figure}[t]
\begin{center}
\begin{subfigure}{.48\textwidth}
\includegraphics[width=0.96\textwidth]{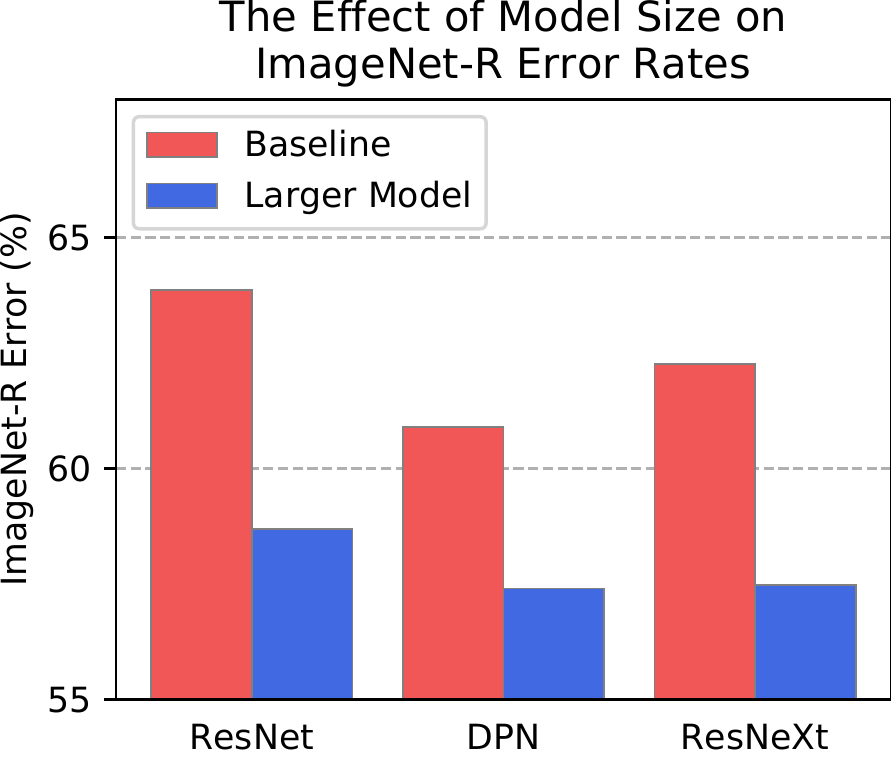}
\caption{Larger models improve robustness on ImageNet-R. The baseline models are ResNet-50, DPN-68, and ResNeXt-50 ($32\times4$d). The larger models are ResNet-152, DPN-98, and ResNeXt-101 ($32\times8$d). The baseline ResNeXt has a 7.1\% ImageNet error rate, while the large has a 6.2\% error rate.}\label{fig:imagenetr}
\end{subfigure}\quad%
\begin{subfigure}{.48\textwidth}
\includegraphics[width=0.96\textwidth]{figs/chap5_figs/imagenet_c.pdf}
\caption{
Accuracy as a function of corruption severity. Severity ``0'' denotes clean data. DeepAugment with AugMix shifts the entire Pareto frontier outward.\phantom{abcddas dfsdafsdfasads fsdfasdfadfsafd adsfdfssdfdfs}}
\end{subfigure}
\end{center}
\end{figure}

Increasing model size appears to automatically improve ImageNet-R performance, as shown in \ref{fig:imagenetr}. A ResNet-50 (25.5M parameters) has 63.9\% error, while a ResNet-152 (60M) has 58.7\% error. ResNeXt-50 32$\times$4d (25.0M) attains 62.3\% error and ResNeXt-101 32$\times$8d (88M) attains 57.5\% error.

\noindent \textbf{ImageNet-C.}\quad
Expanded ImageNet-C results are \ref{tab:imagenetc_table}. We also tested whether model size improves performance on ImageNet-C for even larger models. With a different codebase, we trained ResNet-50, ResNet-152, and ResNet-500 models which achieved 80.6, 74.0, and 68.5 mCE respectively.

\noindent \textbf{ImageNet-A.}\quad
ImageNet-A \cite{Hendrycks2019NaturalAE} is an adversarially filtered test set. This dataset contains examples that are difficult for a ResNet-50 to classify, so examples solvable by simple spurious cues are are especially infrequent in this dataset.
Results are in \ref{tab:imageneta}. Notice Res2Net architectures \cite{Gao2019Res2NetAN} can greatly improve accuracy. Results also show that \emph{Larger Models}, \emph{Self-Attention}, and \emph{Pretraining} help, while \emph{Diverse Data Augmentation} usually does not help substantially.

\noindent \textbf{Implications for the Four Method Hypotheses.}\quad \\
The \emph{Larger Models} hypothesis has support with ImageNet-C ($+$), ImageNet-A ($+$), ImageNet-R ($+$), yet does not markedly improve DFR ($-$) performance.\\
The \emph{Self-Attention} hypothesis has support with ImageNet-C ($+$), ImageNet-A ($+$), yet does not help ImageNet-R ($-$) and DFR ($-$) performance.\\
The \emph{Diverse Data Augmentation} hypothesis has support with ImageNet-C ($+$), ImageNet-R ($+$), yet does not markedly improve ImageNet-A ($-$), DFR($-$), nor SVSF ($-$) performance.\\
The \emph{Pretraining} hypothesis has support with ImageNet-C ($+$), ImageNet-A ($+$), yet does not markedly improve DFR ($-$) nor ImageNet-R ($-$) performance.

\begin{table}
\centering
\begin{tabular}{lcccccc}
  Hypothesis       & ImageNet-C & ImageNet-A & ImageNet-R  & DFR & SVSF \\ 
\hline
\emph{Larger Models}    &  $+$ & $+$ & $+$ & $-$ \\
\emph{Self-Attention}   & $+$ & $+$ & $-$ & $-$  \\
\emph{Diverse Data Augmentation}       & $+$ & $-$ & $+$ & $-$ & $-$  \\
\emph{Pretraining}      & $+$ & $+$ & $-$ & $-$  \\
\bottomrule
\end{tabular}
\caption{A highly simplified account of each hypothesis when tested against different datasets. This table includes ImageNet-A results.}
\label{tab:hypothesissummaryimageneta}
\end{table}

\newpage

\begin{table*}[t]
\begin{center}
\begin{tabular}{lcc>{\color{gray}}c}
\toprule
                                        &  ImageNet-200 (\%)    & ImageNet-R (\%)   &  Gap      \\ \midrule
ResNet-50 \cite{resnet}                              &  7.9                  &  63.9             &  56.0     \\
+ ImageNet-21K \emph{Pretraining} ($10\times$ data)         &  7.0                  &  62.8             &  55.8     \\
+ CBAM (\emph{Self-Attention})                      &  7.0                  &  63.2             &  56.2     \\ 
+ $\ell_\infty$ Adversarial Training       &  25.1                 &  68.6             &  43.5     \\
+ Speckle Noise              &  8.1                  &  62.1             &  54.0     \\
+ Style Transfer             &  8.9                  &  58.5             &  49.6     \\ 
+ AugMix                    &  7.1                  &  58.9             &  51.8     \\
+ DeepAugment                 &  7.5                  &  57.8             &  50.3     \\
+ DeepAugment + AugMix        &  8.0                  &  53.2    &  45.2     \\ \midrule
ResNet-101 (\emph{Larger Models})                             &  7.1                  &  60.7             &  53.6     \\
+ SE (\emph{Self-Attention})                        &  6.7                  &  61.0             &  54.3     \\ \midrule
ResNet-152 (\emph{Larger Models})                             &  6.8                  &  58.7             &  51.9     \\
+ SE (\emph{Self-Attention})                         &  6.6                  &  60.0             &  53.4     \\ \midrule
ResNeXt-101 32$\times$\text{4d} (\emph{Larger Models})        &  6.8                  &  58.0             &  51.2     \\
+ SE (\emph{Self-Attention})    &  5.9                  &  59.6             &  53.7     \\ \midrule
ResNeXt-101 32$\times$\text{8d} (\emph{Larger Models})         &  6.2                  &  57.5             &  51.3     \\
+ WSL \emph{Pretraining} ($1000\times$ data)  &  4.1   &  24.2             &  20.1     \\
+ DeepAugment + AugMix        &  6.1                  &  47.9             &  41.8     \\

\bottomrule
\end{tabular}
\end{center}
\caption{ImageNet-200 and ImageNet-Renditions error rates. ImageNet-21K and WSL Pretraining test the \emph{Pretraining} hypothesis, and here pretraining gives mixed benefits. CBAM and SE test the \emph{Self-Attention} hypothesis, and these \emph{hurt} robustness. ResNet-152 and ResNeXt-101 32$\times$\text{8d} test the \emph{Larger Models} hypothesis, and these help. Other methods augment data, and Style Transfer, AugMix, and DeepAugment provide support for the \emph{Diverse Data Augmentation} hypothesis.
}
\label{tab:imagenetr_full}
\end{table*}

\begin{table}
\scriptsize
\begin{center}
{\setlength\tabcolsep{0.9pt}%
\begin{tabular}{@{}l | c | c | c c c | c c c c | c c c  c | c c c c@{}}
 & \multicolumn{2}{c}{} & \multicolumn{3}{c}{Noise} & \multicolumn{4}{c}{Blur} & \multicolumn{4}{c}{Weather} & \multicolumn{4}{c}{Digital}\\
\cline{1-18}
 & \multicolumn{1}{c|}{\,\emph{Clean}\,} & {\,\emph{mCE}\,} & \scriptsize{Gauss.}
    & \scriptsize{Shot} & \scriptsize{Impulse} & \scriptsize{Defocus} & \scriptsize{Glass} & \scriptsize{Motion} & \scriptsize{Zoom} & \scriptsize{Snow} & \scriptsize{Frost} & \scriptsize{Fog} & \scriptsize{Bright} & \scriptsize{Contrast} & \scriptsize{Elastic} & \scriptsize{Pixel} & \scriptsize{JPEG} \\ \hline 
ResNet-50                   & 23.9 & 76.7 & 80 & 82 & 83 & 75 & 89 & 78 & 80 & 78 & 75 & 66 & 57 & 71 & 85 & 77 & 77\\
+ ImageNet-21K \emph{Pretraining}     & 22.4 & 65.8 & 61 & 64 & 63 & 69 & 84 & 68 & 74 & 69 & 71 & 61 & 53 & 53 & 81 & 54 & 63 \\
+ SE (\emph{Self-Attention})  & 22.4 & 68.2 & 63 & 66 & 66 & 71 & 82 & 67 & 74 & 74 & 72 & 64 & 55 & 71 & 73 & 60 & 67 \\
+ CBAM (\emph{Self-Attention})  & 22.4 & 70.0 & 67 & 68 & 68 & 74 & 83 & 71 & 76 & 73 & 72 & 65 & 54 & 70 & 79 & 62 & 67 \\
+ $\ell_\infty$ Adversarial Training      & 46.2 & 94.0 & 91 & 92 & 95 & 97 & 86 & 92 & 88 & 93 & 99 & 118 & 104 & 111 & 90 & 72 & 81  \\
+ Speckle Noise             & 24.2 & 68.3 & 51 & 47 & 55 & 70 & 83 & 77 & 80 & 76 & 71 & 66 & 57 & 70 & 82 & 72 & 69 \\
+ Style Transfer            & 25.4 & 69.3 & 66 & 67 & 68 & 70 & 82 & 69 & 80 & 68 & 71 & 65 & 58 & 66 & 78 & 62 & 70 \\
+ AugMix                    & 22.5 & 65.3 & 67 & 66 & 68 & 64 & 79 & 59 & 64 & 69 & 68 & 65 & 54 & 57 & 74 & 60 & 66  \\
+ DeepAugment               & 23.3 & 60.4 & 49 & 50 & 47 & 59 & 73 & 65 & 76 & 64 & 60 & 58 & 51 & 61 & 76 & 48 & 67  \\
+ DeepAugment + AugMix      & 24.2 & 53.6 & 46 & 45 & 44 & 50 & 64 & 50 & 61& 58 & 57 & 54 & 52 & 48 & 71 & 43 & 61  \\ \midrule
ResNet-152 (\emph{Larger Models}) & 21.7 & 69.3 & 73 & 73 & 76 & 67 & 81 & 66 & 74 & 71 & 68 & 62 & 51 & 67 & 76 & 69 & 65 \\ \midrule
ResNeXt-101 32$\times$\text{8d} (\emph{Larger Models}) & 20.7 & 66.7 & 68 & 69 & 71 & 65 & 79 & 66 & 71 & 69 & 66 & 60 & 50 & 66 & 74 & 61 & 64  \\ 
+ WSL \emph{Pretraining} ($1000\times$ data)           & 17.8 & 51.7 & 49 & 50 & 51 & 53 & 72 & 55 & 63 & 53 & 51 & 42 & 37 & 41 & 67 & 40 & 51 \\
+ DeepAugment + AugMix          & 20.1 & 44.5 & 36 & 35 & 34 & 43 & 55 & 42 & 55 & 48 & 48 & 47 & 43 & 39 & 59 & 34 & 50 \\

\end{tabular}}
\caption{Clean Error, Corruption Error (CE), and mean CE (mCE) values for various models and training methods on ImageNet-C. The mCE value is computed by averaging across all 15 CE values. A CE value greater than 100 (e.g. adversarial training on contrast) denotes worse performance than AlexNet. DeepAugment+AugMix improves robustness by over 23 mCE.}
\label{tab:imagenetc_table}
\end{center}
\end{table}

\begin{table*}
\begin{center}
\begin{tabular}{lc}
\toprule
                        & ImageNet-A (\%)  \\\midrule
ResNet-50   & 2.2 \\
+ ImageNet-21K \emph{Pretraining} ($10\times$ data)    & 11.4 \\
+ Squeeze-and-Excitation (\emph{Self-Attention}) & 6.2 \\
+ CBAM (\emph{Self-Attention}) & 6.9 \\
+ $\ell_\infty$ Adversarial Training    & 1.7 \\
+ Style Transfer  & 2.0 \\
+ AugMix  & 3.8\\
+ DeepAugment & 3.5 \\
+ DeepAugment + AugMix  & 3.9\\
\hline
ResNet-152 (\emph{Larger Models})             & 6.1        \\
ResNet-152+Squeeze-and-Excitation (\emph{Self-Attention})           & 9.4        \\
\hline
Res2Net-50 v1b        & 14.6       \\
Res2Net-152 v1b (\emph{Larger Models})       & 22.4        \\
\hline
ResNeXt-101 ($32\times8$d) (\emph{Larger Models})   & 10.2  \\
+ WSL \emph{Pretraining} ($1000\times$ data) & 45.4 \\
+ DeepAugment + AugMix & 11.5 \\
\bottomrule
\end{tabular}
\end{center}
\caption{ImageNet-A top-1 accuracy.}\label{tab:imageneta}
\end{table*}

\newpage
\section{Further Dataset Descriptions}
\paragraph{ImageNet-R Classes.}\label{app:classes}
The 200 ImageNet classes and their WordNet IDs in ImageNet-R are as follows.

Goldfish, \quad great white shark, \quad hammerhead, \quad stingray, \quad hen, \quad ostrich, \quad goldfinch, \quad junco, \quad bald eagle, \quad vulture, \quad newt, \quad axolotl, \quad tree frog, \quad iguana, \quad African chameleon, \quad cobra, \quad scorpion, \quad tarantula, \quad centipede, \quad peacock, \quad lorikeet, \quad hummingbird, \quad toucan, \quad duck, \quad goose, \quad black swan, \quad koala, \quad jellyfish, \quad snail, \quad lobster, \quad hermit crab, \quad flamingo, \quad american egret, \quad pelican, \quad king penguin, \quad grey whale, \quad killer whale, \quad sea lion, \quad chihuahua, \quad shih tzu, \quad afghan hound, \quad basset hound, \quad beagle, \quad bloodhound, \quad italian greyhound, \quad whippet, \quad weimaraner, \quad yorkshire terrier, \quad boston terrier, \quad scottish terrier, \quad west highland white terrier, \quad golden retriever, \quad labrador retriever, \quad cocker spaniels, \quad collie, \quad border collie, \quad rottweiler, \quad german shepherd dog, \quad boxer, \quad french bulldog, \quad saint bernard, \quad husky, \quad dalmatian, \quad pug, \quad pomeranian, \quad chow chow, \quad pembroke welsh corgi, \quad toy poodle, \quad standard poodle, \quad timber wolf, \quad hyena, \quad red fox, \quad tabby cat, \quad leopard, \quad snow leopard, \quad lion, \quad tiger, \quad cheetah, \quad polar bear, \quad meerkat, \quad ladybug, \quad fly, \quad bee, \quad ant, \quad grasshopper, \quad cockroach, \quad mantis, \quad dragonfly, \quad monarch butterfly, \quad starfish, \quad wood rabbit, \quad porcupine, \quad fox squirrel, \quad beaver, \quad guinea pig, \quad zebra, \quad pig, \quad hippopotamus, \quad bison, \quad gazelle, \quad llama, \quad skunk, \quad badger, \quad orangutan, \quad gorilla, \quad chimpanzee, \quad gibbon, \quad baboon, \quad panda, \quad eel, \quad clown fish, \quad puffer fish, \quad accordion, \quad ambulance, \quad assault rifle, \quad backpack, \quad barn, \quad wheelbarrow, \quad basketball, \quad bathtub, \quad lighthouse, \quad beer glass, \quad binoculars, \quad birdhouse, \quad bow tie, \quad broom, \quad bucket, \quad cauldron, \quad candle, \quad cannon, \quad canoe, \quad carousel, \quad castle, \quad mobile phone, \quad cowboy hat, \quad electric guitar, \quad fire engine, \quad flute, \quad gasmask, \quad grand piano, \quad guillotine, \quad hammer, \quad harmonica, \quad harp, \quad hatchet, \quad jeep, \quad joystick, \quad lab coat, \quad lawn mower, \quad lipstick, \quad mailbox, \quad missile, \quad mitten, \quad parachute, \quad pickup truck, \quad pirate ship, \quad revolver, \quad rugby ball, \quad sandal, \quad saxophone, \quad school bus, \quad schooner, \quad shield, \quad soccer ball, \quad space shuttle, \quad spider web, \quad steam locomotive, \quad scarf, \quad submarine, \quad tank, \quad tennis ball, \quad tractor, \quad trombone, \quad vase, \quad violin, \quad military aircraft, \quad wine bottle, \quad ice cream, \quad bagel, \quad pretzel, \quad cheeseburger, \quad hotdog, \quad cabbage, \quad broccoli, \quad cucumber, \quad bell pepper, \quad mushroom, \quad Granny Smith, \quad strawberry, \quad lemon, \quad pineapple, \quad banana, \quad pomegranate, \quad pizza, \quad burrito, \quad espresso, \quad volcano, \quad baseball player, \quad scuba diver, \quad acorn.

n01443537, \quad n01484850, \quad n01494475, \quad n01498041, \quad n01514859, \quad n01518878, \quad n01531178, \quad n01534433, \quad n01614925, \quad n01616318, \quad n01630670, \quad n01632777, \quad n01644373, \quad n01677366, \quad n01694178, \quad n01748264, \quad n01770393, \quad n01774750, \quad n01784675, \quad n01806143, \quad n01820546, \quad n01833805, \quad n01843383, \quad n01847000, \quad n01855672, \quad n01860187, \quad n01882714, \quad n01910747, \quad n01944390, \quad n01983481, \quad n01986214, \quad n02007558, \quad n02009912, \quad n02051845, \quad n02056570, \quad n02066245, \quad n02071294, \quad n02077923, \quad n02085620, \quad n02086240, \quad n02088094, \quad n02088238, \quad n02088364, \quad n02088466, \quad n02091032, \quad n02091134, \quad n02092339, \quad n02094433, \quad n02096585, \quad n02097298, \quad n02098286, \quad n02099601, \quad n02099712, \quad n02102318, \quad n02106030, \quad n02106166, \quad n02106550, \quad n02106662, \quad n02108089, \quad n02108915, \quad n02109525, \quad n02110185, \quad n02110341, \quad n02110958, \quad n02112018, \quad n02112137, \quad n02113023, \quad n02113624, \quad n02113799, \quad n02114367, \quad n02117135, \quad n02119022, \quad n02123045, \quad n02128385, \quad n02128757, \quad n02129165, \quad n02129604, \quad n02130308, \quad n02134084, \quad n02138441, \quad n02165456, \quad n02190166, \quad n02206856, \quad n02219486, \quad n02226429, \quad n02233338, \quad n02236044, \quad n02268443, \quad n02279972, \quad n02317335, \quad n02325366, \quad n02346627, \quad n02356798, \quad n02363005, \quad n02364673, \quad n02391049, \quad n02395406, \quad n02398521, \quad n02410509, \quad n02423022, \quad n02437616, \quad n02445715, \quad n02447366, \quad n02480495, \quad n02480855, \quad n02481823, \quad n02483362, \quad n02486410, \quad n02510455, \quad n02526121, \quad n02607072, \quad n02655020, \quad n02672831, \quad n02701002, \quad n02749479, \quad n02769748, \quad n02793495, \quad n02797295, \quad n02802426, \quad n02808440, \quad n02814860, \quad n02823750, \quad n02841315, \quad n02843684, \quad n02883205, \quad n02906734, \quad n02909870, \quad n02939185, \quad n02948072, \quad n02950826, \quad n02951358, \quad n02966193, \quad n02980441, \quad n02992529, \quad n03124170, \quad n03272010, \quad n03345487, \quad n03372029, \quad n03424325, \quad n03452741, \quad n03467068, \quad n03481172, \quad n03494278, \quad n03495258, \quad n03498962, \quad n03594945, \quad n03602883, \quad n03630383, \quad n03649909, \quad n03676483, \quad n03710193, \quad n03773504, \quad n03775071, \quad n03888257, \quad n03930630, \quad n03947888, \quad n04086273, \quad n04118538, \quad n04133789, \quad n04141076, \quad n04146614, \quad n04147183, \quad n04192698, \quad n04254680, \quad n04266014, \quad n04275548, \quad n04310018, \quad n04325704, \quad n04347754, \quad n04389033, \quad n04409515, \quad n04465501, \quad n04487394, \quad n04522168, \quad n04536866, \quad n04552348, \quad n04591713, \quad n07614500, \quad n07693725, \quad n07695742, \quad n07697313, \quad n07697537, \quad n07714571, \quad n07714990, \quad n07718472, \quad n07720875, \quad n07734744, \quad n07742313, \quad n07745940, \quad n07749582, \quad n07753275, \quad n07753592, \quad n07768694, \quad n07873807, \quad n07880968, \quad n07920052, \quad n09472597, \quad n09835506, \quad n10565667, \quad n12267677.

\paragraph{SVSF.} The classes are
\begin{multicols}{3}
    \begin{itemize}
        \item auto shop
        \item bakery
        \item bank
        \item beauty salon
        \item car dealer
        \item car wash
        \item cell phone store
        \item dentist
        \item discount store
        \item dry cleaner
        \item furniture store
        \item gas station
        \item gym
        \item hardware store
        \item hotel
        \item liquor store
        \item pharmacy
        \item religious institution
        \item storage facility
        \item veterinary care.
    \end{itemize}
\end{multicols}

\paragraph{DeepFashion Remixed.} The classes are
\begin{multicols}{3}
    \begin{itemize}
        \item short sleeve top
        \item long sleeve top
        \item short sleeve outerwear
        \item long sleeve outerwear
        \item vest
        \item sling
        \item shorts
        \item trousers
        \item skirt
        \item short sleeve dress
        \item long sleep dress
        \item vest dress
        \item sling dress.
    \end{itemize}
\end{multicols} \label{app:dfr_classes}

Size (small, moderate, or large) defines how much of the image the article of clothing takes up. Occlusion (slight, medium, or heavy) defines the degree to which the object is occluded from the camera. Viewpoint (front, side/back, or not worn) defines the camera position relative to the article of clothing. Zoom (no zoom, medium, or large) defines how much camera zoom was used to take the picture.  

\begin{table*}
\begin{center}
\begin{tabular}{lc}
\toprule
                        & Represented Distribution Shifts           \\\midrule
ImageNet-Renditions     & artistic renditions (cartoons, graffiti, embroidery, graphics, origami, \\
                        & paintings, sculptures, sketches, tattoos, toys, ...)       \\
DeepFashion Remixed     & occlusion, size, viewpoint, zoom         \\
StreetView StoreFronts  & camera, capture year, country             \\
\bottomrule
\end{tabular}
\caption{Various distribution shifts represented in our three new benchmarks. ImageNet-Renditions is a new test set for ImageNet trained models measuring robustness to various object renditions. DeepFashion Remixed and StreetView StoreFronts each contain a training set and multiple test sets capturing a variety of distribution shifts.}
\label{tab:datasetshifts}
\end{center}
\end{table*}

\begin{table*}
\begin{center}
\begin{tabular}{lcc}
\toprule
                        & Training set              &  Testing images \\\midrule
ImageNet-R     & 1281167 &  30000          \\
DFR     & 48000                    &  42640, 7440, 28160, 10360, 480, 11040, 10520, 10640    \\
SVSF  & 200000                   &  10000, 10000, 10000, 8195, 9788  \\
\bottomrule
\end{tabular}
\caption{Number of images in each training and test set. ImageNet-R training set refers to the ILSVRC 2012 training set \cite{imagenet}. DeepFashion Remixed test sets are: in-distribution, occlusion - none/slight, occlusion - heavy, size - small, size - large, viewpoint - frontal, viewpoint - not-worn, zoom-in - medium, zoom-in - large. StreetView StoreFronts test sets are: in-distribution, capture year - 2018, capture year - 2017, camera system - new, country - France.}
\label{tab:datasetsizes}
\end{center}
\end{table*}

\newpage
\section{DeepAugment Details}\label{app:deepaugment}
\begin{lstlisting}[language=Python]
def main():
    net.apply_weights(deepAugment_getNetwork())  # EDSR, CAE, ...
    for image in dataset:  # May be the ImageNet training set
        if np.random.uniform() < 0.05:  # Arbitrary refresh prob
            net.apply_weights(deepAugment_getNetwork())
        new_image = net.deepAugment_forwardPass(image)

def deepAugment_getNetwork():
    weights = load_clean_weights()
    weight_distortions = sample_weight_distortions()
    for d in weight_distortions:
        weights = apply_distortion(d, weights)
    return weights

def sample_weight_distortions():
    distortions = [
        negate_weights,
        zero_weights,
        flip_transpose_weights,
        ...
    ]
    
    return random_subset(distortions)
    
def sample_signal_distortions():
    distortions = [
        dropout,
        gelu,
        negate_signal_random_mask,
        flip_signal,        
        ...
    ]
    
    return random_subset(distortions)


class Network():
    def apply_weights(weights):
        # Apply given weight tensors to network
        ...
    
    # Clean forward pass. Compare to deepAugment_forwardPass()
    def clean_forwardPass(X):
        X = network.block1(X)
        X = network.block2(X)
        ...
        X = network.blockN(X)
        return X

    # Our forward pass. Compare to clean_forwardPass()
    def deepAugment_forwardPass(X):
        # Returns a list of distortions, each of which 
        # will be applied at a different layer.
        signal_distortions = sample_signal_distortions()
	       
        X = network.block1(X)
        apply_layer_1_distortions(X, signal_distortions)
        X = network.block2(X)
        apply_layer_2_distortions(X, signal_distortions)
        ...
        apply_layer_N-1_distortions(X, signal_distortions)
        X = network.blockN(X)
        apply_layer_N_distortions(X, signal_distortions)
	
    return X
\end{lstlisting}
\paragraph{Pseudocode.} Above is Pythonic pseudocode for DeepAugment. The basic structure of DeepAugment is agnostic to the backbone network used, but specifics such as which layers are chosen for various transforms may vary as the backbone architecture varies. We do not need to train many different image-to-image models to get diverse distortions \cite{zhang2018perceptual,Lee2020NetworkRA}. We only use two existing models, the EDSR super-resolution model \cite{lim2017enhanced} and the CAE image compression model \cite{theis2017lossy}. See full code for such details.

At a high level, we process each image with an image-to-image network. The image-to-image's weights and feedforward signal pass are distorted with each pass. The distortion is made possible by, for example, negating the network's weights and applying dropout to the feedforward signal. The resulting image is distorted and saved. This process generates an augmented dataset.


\makebibliography



@techreport{achanta2010slic,
    title       = {Slic superpixels},
    author      = {Achanta, Radhakrishna and Shaji, Appu and Smith, Kevin and Lucchi, Aurelien and Fua, Pascal and S{\"u}sstrunk, Sabine},
    year        = 2010
}

@article{adam,
    title       = {Adam: {A} Method for Stochastic Optimization},
    author      = {Diederik P. Kingma and Jimmy Ba},
    year        = 2014,
    journal     = {ICLR}
}

@article{Ahmed2019DetectingSA,
    title       = {Detecting semantic anomalies},
    author      = {Faruk Ahmed and Aaron C. Courville},
    year        = 2019,
    journal     = {ArXiv},
    volume      = {abs/1908.04388}
}

@article{AlexNet,
    title       = {{I}mage{N}et Classification with Deep Convolutional Neural Networks},
    author      = {Alex Krizhevsky and Sutskever, Ilya and Hinton, Geoffrey E},
    year        = 2012,
    journal     = {NIPS}
}

@inproceedings{Altman1992KNN,
    title       = {An Introduction to Kernel and Nearest Neighbor Nonparametric Regression},
    author      = {Naomi S. Altman},
    year        = 1992
}

@article{anguelov2010google,
    title       = {Google street view: Capturing the world at street level},
    author      = {Anguelov, Dragomir and Dulong, Carole and Filip, Daniel and Frueh, Christian and Lafon, St{\'e}phane and Lyon, Richard and Ogale, Abhijit and Vincent, Luc and Weaver, Josh},
    year        = 2010,
    journal     = {Computer},
    publisher   = {IEEE},
    volume      = 43,
    number      = 6,
    pages       = {32--38}
}

@misc{arazo2019unsupervised,
    title       = {Unsupervised Label Noise Modeling and Loss Correction},
    author      = {Eric Arazo and Diego Ortego and Paul Albert and Noel E. O'Connor and Kevin McGuinness},
    year        = 2019,
    eprint      = {1904.11238},
    archiveprefix = {arXiv},
    primaryclass = {cs.CV}
}

@article{arbelaez2010contour,
    title       = {Contour detection and hierarchical image segmentation},
    author      = {Arbelaez, Pablo and Maire, Michael and Fowlkes, Charless and Malik, Jitendra},
    year        = 2010,
    journal     = {IEEE transactions on pattern analysis and machine intelligence},
    publisher   = {IEEE},
    volume      = 33,
    number      = 5,
    pages       = {898--916}
}

@article{Arjovsky2019InvariantRM,
    title       = {Invariant Risk Minimization},
    author      = {Mart{\'i}n Arjovsky and L{\'e}on Bottou and Ishaan Gulrajani and David Lopez-Paz},
    year        = 2019,
    journal     = {ArXiv},
    volume      = {abs/1907.02893}
}

@article{athalye2017synthesizing,
    title       = {Synthesizing robust adversarial examples},
    author      = {Athalye, Anish and Engstrom, Logan and Ilyas, Andrew and Kwok, Kevin},
    year        = 2017,
    journal     = {arXiv preprint arXiv:1707.07397}
}

@incollection{auprbaseline,
    title       = {The precision-recall plot is more informative than the {ROC} plot when evaluating binary classifiers on imbalanced datasets},
    author      = {Takaya Saito and Marc Rehmsmeier},
    year        = 2015,
    booktitle   = {PLoS ONE}
}

@article{Baluja2017AdversarialTN,
    title       = {Adversarial Transformation Networks: Learning to Generate Adversarial Examples},
    author      = {Shumeet Baluja and Ian Fischer},
    year        = 2017,
    journal     = {CoRR},
    volume      = {abs/1703.09387}
}

@article{Baur_2019,
    title       = {Deep Autoencoding Models for Unsupervised Anomaly Segmentation in Brain MR Images},
    author      = {Baur, Christoph and Wiestler, Benedikt and Albarqouni, Shadi and Navab, Nassir},
    year        = 2019,
    journal     = {Lecture Notes in Computer Science},
    publisher   = {Springer International Publishing},
    pages       = {161–169},
    doi         = {10.1007/978-3-030-11723-8_16},
    isbn        = 9783030117238,
    issn        = {1611-3349}
}

@article{bdd100k,
    title       = {{BDD100K:} {A} Diverse Driving Video Database with Scalable Annotation Tooling},
    author      = {Fisher Yu and Wenqi Xian and Yingying Chen and Fangchen Liu and Mike Liao and Vashisht Madhavan and Trevor Darrell},
    year        = 2018,
    journal     = {CoRR},
    volume      = {abs/1805.04687},
    archiveprefix = {arXiv},
    eprint      = {1805.04687},
    bibsource   = {dblp computer science bibliography, https://dblp.org}
}

@inproceedings{beede2020human,
    title       = {A Human-Centered Evaluation of a Deep Learning System Deployed in Clinics for the Detection of Diabetic Retinopathy},
    author      = {Beede, Emma and Baylor, Elizabeth and Hersch, Fred and Iurchenko, Anna and Wilcox, Lauren and Ruamviboonsuk, Paisan and Vardoulakis, Laura M},
    year        = 2020,
    booktitle   = {Proceedings of the 2020 CHI Conference on Human Factors in Computing Systems},
    pages       = {1--12}
}

@misc{bev2018discriminative,
    title       = {Discriminative out-of-distribution detection for semantic segmentation},
    author      = {Petra Bevandić and Ivan Krešo and Marin Oršić and Siniša Šegvić},
    year        = 2018,
    eprint      = {1808.07703},
    archiveprefix = {arXiv},
    primaryclass = {cs.CV}
}

@article{Bhagavatula2019AbductiveCR,
    title       = {Abductive Commonsense Reasoning},
    author      = {Chandra Bhagavatula and Ronan Le Bras and Chaitanya Malaviya and Keisuke Sakaguchi and Ari Holtzman and Hannah Rashkin and Doug Downey and Scott Yih and Yejin Choi},
    year        = 2019,
    journal     = {ArXiv},
    volume      = {abs/1908.05739}
}

@article{Bhattad2019BigBI,
author = {Bhattad, Anand and Chong, Min and Liang, Kaizhao and Li, Bo and Forsyth, David},
year = {2019},
title = {Big but Imperceptible Adversarial Perturbations via Semantic Manipulation},
journal     = {abs/1904.06347}
}

@inproceedings{Bhattad2020UnrestrictedAEcolorizationAttack,
    title       = {Unrestricted Adversarial Examples via Semantic Manipulation},
    author      = {Anand Bhattad and Min Jin Chong and Kaizhao Liang and Bo Li and David A Forsyth},
    year        = 2020,
    booktitle   = {ICLR}
}

@article{biederman1988surface,
    title       = {Surface versus edge-based determinants of visual recognition},
    author      = {Biederman, Irving and Ju, Ginny},
    year        = 1988,
    journal     = {Cognitive psychology},
    publisher   = {Elsevier},
    volume      = 20,
    number      = 1,
    pages       = {38--64}
}

@article{Bisk2019PIQARA,
    title       = {PIQA: Reasoning about Physical Commonsense in Natural Language},
    author      = {Yonatan Bisk and Rowan Zellers and Ronan Le Bras and Jianfeng Gao and Yejin Choi},
    year        = 2019,
    journal     = {ArXiv},
    volume      = {abs/1911.11641}
}

@misc{blum2019fishyscapes,
    title       = {The Fishyscapes Benchmark: Measuring Blind Spots in Semantic Segmentation},
    author      = {Hermann Blum and Paul-Edouard Sarlin and Juan Nieto and Roland Siegwart and Cesar Cadena},
    year        = 2019,
    eprint      = {1904.03215},
    archiveprefix = {arXiv},
    primaryclass = {cs.CV}
}

@article{Brendel2018ApproximatingCW,
    title       = {Approximating CNNs with Bag-of-local-Features models works surprisingly well on ImageNet},
    author      = {Wieland Brendel and Matthias Bethge},
    year        = 2018,
    journal     = {CoRR},
    volume      = {abs/1904.00760}
}

@inproceedings{Breunig2000LOFID,
    title       = {LOF: identifying density-based local outliers},
    author      = {M. Breunig and H. Kriegel and R. Ng and J. Sander},
    year        = 2000,
    booktitle   = {SIGMOD '00}
}

@article{brown2017adversarial,
    title       = {Adversarial patch},
    author      = {Brown, Tom B and Man{\'e}, Dandelion and Roy, Aurko and Abadi, Mart{\'i}n and Gilmer, Justin},
    year        = 2017,
    journal     = {arXiv preprint arXiv:1712.09665}
}

@article{Brown2018UnrestrictedAE,
    title       = {Unrestricted Adversarial Examples},
    author      = {Tom B. Brown and Nicholas Carlini and Chiyuan Zhang and Catherine Olsson and Paul Francis Christiano and Ian J. Goodfellow},
    year        = 2018,
    journal     = {CoRR},
    volume      = {abs/1809.08352}
}

@inproceedings{BSDSmartin2001,
    title       = {A database of human segmented natural images and its application to evaluating segmentation algorithms and measuring ecological statistics},
    author      = {Martin, David and Fowlkes, Charless and Tal, Doron and Malik, Jitendra},
    year        = 2001,
    booktitle   = {Proceedings Eighth IEEE International Conference on Computer Vision. ICCV 2001},
    volume      = 2,
    pages       = {416--423},
    organization = {IEEE}
}

@inproceedings{Cai2017PayAT,
    title       = {Pay Attention to the Ending: Strong Neural Baselines for the ROC Story Cloze Task},
    author      = {Zheng Cai and Lifu Tu and Kevin Gimpel},
    year        = 2017,
    booktitle   = {ACL}
}

@inproceedings{carla,
    title       = {{CARLA}: {An} Open Urban Driving Simulator},
    author      = {Alexey Dosovitskiy and German Ros and Felipe Codevilla and Antonio Lopez and Vladlen Koltun},
    year        = 2017,
    booktitle   = {Proceedings of the 1st Annual Conference on Robot Learning},
    pages       = {1--16}
}

@article{Carlini2017Wagner,
    title       = {Towards Evaluating the Robustness of Neural Networks},
    author      = {Nicholas Carlini and David A. Wagner},
    year        = 2017,
    journal     = {2017 IEEE Symposium on Security and Privacy (SP)},
    pages       = {39--57}
}

@article{carlini2019evaluating,
    title       = {On evaluating adversarial robustness},
    author      = {Carlini, Nicholas and Athalye, Anish and Papernot, Nicolas and Brendel, Wieland and Rauber, Jonas and Tsipras, Dimitris and Goodfellow, Ian and Madry, Aleksander and Kurakin, Alexey},
    year        = 2019,
    journal     = {arXiv preprint arXiv:1902.06705}
}

@inproceedings{carreira2010constrained,
    title       = {Constrained parametric min-cuts for automatic object segmentation},
    author      = {Carreira, Joao and Sminchisescu, Cristian},
    year        = 2010,
    booktitle   = {2010 IEEE Computer Society Conference on Computer Vision and Pattern Recognition},
    pages       = {3241--3248},
    organization = {IEEE}
}

@misc{chao2020revisiting,
    title       = {Revisiting Meta-Learning as Supervised Learning},
    author      = {Wei-Lun Chao and Han-Jia Ye and De-Chuan Zhan and Mark Campbell and Kilian Q. Weinberger},
    year        = 2020,
    eprint      = {2002.00573},
    archiveprefix = {arXiv},
    primaryclass = {cs.LG}
}

@article{cifar,
    title       = {{Learning multiple layers of features from tiny images.}},
    author      = {A. Krizhevsky and G. Hinton},
    year        = 2009,
    journal     = {Tech Report},
    primaryclass = {cs.CV}
}

@article{cimpoi14describing,
    title       = {Describing Textures in the Wild},
    author      = {Mircea Cimpoi and Subhransu Maji and Iasonas Kokkinos and Sammy Mohamed and Andrea Vedaldi},
    year        = 2014,
    journal     = {Computer Vision and Pattern Recognition}
}

@article{coco,
    title       = {Microsoft {COCO}: Common Objects in Context},
    author      = {Tsung-Yi Lin and Michael Maire and Serge Belongie and Lubomir Bourdev and Ross Girshick and James Hays and Pietro Perona and Deva Ramanan and C. Lawrence Zitnick and Piotr Dollar},
    year        = 2014,
    journal     = {ECCV}
}

@inproceedings{Cohen2019SmoothingAdversarialCertification,
    title       = {Certified Adversarial Robustness via Randomized Smoothing},
    author      = {Jeremy M. Cohen and Elan Rosenfeld and J. Zico Kolter},
    year        = 2019,
    booktitle   = {ICML}
}

@inproceedings{Cordts2016Cityscapes,
    title       = {The Cityscapes Dataset for Semantic Urban Scene Understanding},
    author      = {Cordts, Marius and Omran, Mohamed and Ramos, Sebastian and Rehfeld, Timo and Enzweiler, Markus and Benenson, Rodrigo and Franke, Uwe and Roth, Stefan and Schiele, Bernt},
    year        = 2016,
    booktitle   = {Proc. of the IEEE Conference on Computer Vision and Pattern Recognition (CVPR)}
}

@article{Cubuk2018AutoAugmentLA,
    title       = {{AutoAugment}: Learning Augmentation Policies from Data},
    author      = {Ekin Dogus Cubuk and Barret Zoph and Dandelion Man{\'e} and Vijay Vasudevan and Quoc V. Le},
    year        = 2018,
    journal     = {CVPR}
}

@techreport{deng2012large,
    title       = {Large scale visual recognition},
    author      = {Deng, Jia},
    year        = 2012,
    institution = {PRINCETON UNIV NJ DEPT OF COMPUTER SCIENCE}
}

@inproceedings{densenet,
    title       = {Densely connected convolutional networks},
    author      = {Huang, Gao and Liu, Zhuang and van der Maaten, Laurens and Weinberger, Kilian Q},
    year        = 2017,
    booktitle   = {Proceedings of the IEEE Conference on Computer Vision and Pattern Recognition}
}

@article{devries,
    title       = {Learning Confidence for Out-of-Distribution Detection in Neural Networks},
    author      = {DeVries, Terrance and Taylor, Graham W},
    year        = 2018,
    journal     = {arXiv preprint arXiv:1802.04865}
}

@article{Devries2017ImprovedRO,
    title       = {Improved Regularization of Convolutional Neural Networks with Cutout},
    author      = {Terrance Devries and Graham W. Taylor},
    year        = 2017,
    journal     = {arXiv preprint arXiv:1708.04552}
}

@inproceedings{dodge2017study,
    title       = {A study and comparison of human and deep learning recognition performance under visual distortions},
    author      = {Dodge, Samuel and Karam, Lina},
    year        = 2017,
    booktitle   = {2017 26th international conference on computer communication and networks (ICCCN)},
    pages       = {1--7},
    organization = {IEEE}
}

@inproceedings{Dua2019DROPAR,
    title       = {DROP: A Reading Comprehension Benchmark Requiring Discrete Reasoning Over Paragraphs},
    author      = {Dheeru Dua and Yizhong Wang and Pradeep Dasigi and Gabriel Stanovsky and Sameer Singh and Matt Gardner},
    year        = 2019,
    booktitle   = {NAACL-HLT}
}

@inproceedings{Dziedzic2019BandlimitedTA,
    title       = {Band-limited Training and Inference for Convolutional Neural Networks},
    author      = {Adam Dziedzic and John Paparrizos and Sanjay Krishnan and Aaron J. Elmore and Michael J. Franklin},
    year        = 2019,
    booktitle   = {ICML}
}

@misc{emmott2015metaanalysis,
    title       = {A Meta-Analysis of the Anomaly Detection Problem},
    author      = {Andrew Emmott and Shubhomoy Das and Thomas Dietterich and Alan Fern and Weng-Keen Wong},
    year        = 2015,
    eprint      = {1503.01158},
    archiveprefix = {arXiv},
    primaryclass = {cs.AI}
}

@article{engstrom2020identifying,
    title       = {Identifying Statistical Bias in Dataset Replication},
    author      = {Logan Engstrom and Andrew Ilyas and Shibani Santurkar and Dimitris Tsipras and Jacob Steinhardt and Aleksander Madry},
    year        = 2020,
    journal     = {ICML}
}

@article{Everingham2009ThePV_pascal,
    title       = {The Pascal Visual Object Classes (VOC) Challenge},
    author      = {Mark Everingham and Luc Van Gool and Christopher K. I. Williams and John M. Winn and Andrew Zisserman},
    year        = 2009,
    journal     = {International Journal of Computer Vision},
    volume      = 88,
    pages       = {303--338}
}

@article{Eykholt2018PhysicalAE,
    title       = {Physical Adversarial Examples for Object Detectors},
    author      = {Kevin Eykholt and Ivan Evtimov and Earlence Fernandes and Bo Li and Amir Rahmati and Florian Tram{\`e}r and Atul Prakash and Tadayoshi Kohno and Dawn Xiaodong Song},
    year        = 2018,
    journal     = {ArXiv},
    volume      = {abs/1807.07769}
}

@article{finn2017MAML,
    title       = {Model-agnostic meta-learning for fast adaptation of deep networks},
    author      = {Finn, Chelsea and Abbeel, Pieter and Levine, Sergey},
    year        = 2017,
    journal     = {arXiv preprint arXiv:1703.03400}
}

@article{Fischler1981Ransac,
    title       = {Random sample consensus: a paradigm for model fitting with applications to image analysis and automated cartography},
    author      = {M. Fischler and R. Bolles},
    year        = 1981,
    journal     = {Commun. ACM},
    volume      = 24,
    pages       = {381--395}
}

@misc{franceschi2018bilevel,
    title       = {Bilevel Programming for Hyperparameter Optimization and Meta-Learning},
    author      = {Luca Franceschi and Paolo Frasconi and Saverio Salzo and Riccardo Grazzi and Massimilano Pontil},
    year        = 2018,
    eprint      = {1806.04910},
    archiveprefix = {arXiv},
    primaryclass = {stat.ML}
}

@article{gal,
    title       = {Dropout as a Bayesian Approximation: Representing Model Uncertainty in Deep Learning},
    author      = {Yarin Gal and Zoubin Ghahramani},
    year        = 2016,
    journal     = {International Conference on Machine Learning}
}

@article{Gao2019Res2NetAN,
    title       = {Res2Net: A New Multi-scale Backbone Architecture},
    author      = {Shanghua Gao and Ming-Ming Cheng and Kai Zhao and Xinyu Zhang and Ming-Hsuan Yang and Philip H. S. Torr},
    year        = 2019,
    journal     = {IEEE transactions on pattern analysis and machine intelligence}
}

@inproceedings{ge2019deepfashion2,
    title       = {Deepfashion2: A versatile benchmark for detection, pose estimation, segmentation and re-identification of clothing images},
    author      = {Ge, Yuying and Zhang, Ruimao and Wang, Xiaogang and Tang, Xiaoou and Luo, Ping},
    year        = 2019,
    booktitle   = {Proceedings of the IEEE Conference on Computer Vision and Pattern Recognition},
    pages       = {5337--5345}
}

@article{geirhos,
    title       = {Generalisation in humans and deep neural networks},
    author      = {Geirhos, Robert and Temme, Carlos R. M. and Rauber, Jonas and Sch\"{u}tt, Heiko H. and Bethge, Matthias and Wichmann, Felix A.},
    year        = 2018,
    journal     = {NeurIPS}
}

@article{geirhos2019,
    title       = {ImageNet-trained CNNs are biased towards texture; increasing shape bias improves accuracy and robustness},
    author      = {Geirhos, Robert and Rubisch, Patricia and Michaelis, Claudio and Bethge, Matthias and Wichmann, Felix A and Brendel, Wieland},
    year        = 2019,
    journal     = {ICLR}
}

@article{geirhos2020shortcut,
    title       = {Shortcut Learning in Deep Neural Networks},
    author      = {Geirhos, Robert and Jacobsen, J{\"o}rn-Henrik and Michaelis, Claudio and Zemel, Richard and Brendel, Wieland and Bethge, Matthias and Wichmann, Felix A},
    year        = 2020,
    journal     = {arXiv preprint arXiv:2004.07780}
}

@article{Geirhos2020ShortcutLI,
    title       = {Shortcut Learning in Deep Neural Networks},
    author      = {Robert Geirhos and Jorn-Henrik Jacobsen and Claudio Michaelis and Richard S. Zemel and Wieland Brendel and Matthias Bethge and Felix A. Wichmann},
    year        = 2020,
    journal     = {ArXiv},
    volume      = {abs/2004.07780}
}

@article{gelu,
    title       = {Gaussian Error Linear Units (GELUs)},
    author      = {Hendrycks, Dan and Gimpel, Kevin},
    year        = 2016,
    journal     = {arXiv preprint 1606.08415}
}

@article{Gilmer2018MotivatingTR,
    title       = {Motivating the Rules of the Game for Adversarial Example Research},
    author      = {Justin Gilmer and Ryan P. Adams and Ian J. Goodfellow and David Andersen and George E. Dahl},
    year        = 2018,
    journal     = {CoRR},
    volume      = {abs/1807.06732}
}

@article{goodfellow2014explaining,
    title       = {Explaining and harnessing adversarial examples},
    author      = {Goodfellow, Ian J and Shlens, Jonathon and Szegedy, Christian},
    year        = 2014,
    journal     = {arXiv preprint arXiv:1412.6572}
}

@article{Goodfellow2015ExplainingAH,
    title       = {Explaining and Harnessing Adversarial Examples},
    author      = {Ian J. Goodfellow and Jonathon Shlens and Christian Szegedy},
    year        = 2015,
    journal     = {CoRR},
    volume      = {abs/1412.6572}
}

@misc{Goodfellow2016comparemodels,
    title       = {Adversarial Examples and Adversarial Training},
    author      = {Ian J. Goodfellow},
    year        = 2016,
    url         = {http://engineering.purdue.edu/~mark/puthesis}
}

@article{goodfellowblog,
    title       = {Attacking Machine Learning with Adversarial Examples},
    author      = {Ian Goodfellow and Nicolas Papernot and Sandy Huang and Yan Duan and Peter Abbeel},
    year        = 2017,
    journal     = {OpenAI Blog}
}

@article{Gururangan2018AnnotationAI,
    title       = {Annotation Artifacts in Natural Language Inference Data},
    author      = {Suchin Gururangan and Swabha Swayamdipta and Omer Levy and Roy Schwartz and Samuel R. Bowman and Noah A. Smith},
    year        = 2018,
    journal     = {ArXiv},
    volume      = {abs/1803.02324}
}

@inproceedings{haselmann2018anomaly,
    title       = {Anomaly Detection Using Deep Learning Based Image Completion},
    author      = {Haselmann, Matthias and Gruber, Dieter P and Tabatabai, Paul},
    year        = 2018,
    booktitle   = {2018 17th IEEE International Conference on Machine Learning and Applications (ICMLA)},
    pages       = {1237--1242},
    organization = {IEEE}
}

@article{He2015DelvingDI,
  author    = {He, Kaiming and Zhang, Xiangyu and Ren, Shaoqing},
  title     = {Delving Deep into Rectifiers: Surpassing Human-Level Performance on ImageNet Classification},
  journal   = {IEEE International Conference on Computer Vision (ICCV)},
  year      = {2015}
}

@inproceedings{he2017mask,
    title       = {Mask r-cnn},
    author      = {He, Kaiming and Gkioxari, Georgia and Doll{\'a}r, Piotr and Girshick, Ross},
    year        = 2017,
    booktitle   = {Proceedings of the IEEE international conference on computer vision},
    pages       = {2961--2969}
}

@article{hendrycks2017baseline,
    title       = {A Baseline for Detecting Misclassified and Out-of-Distribution Examples in Neural Networks},
    author      = {Dan Hendrycks and Kevin Gimpel},
    year        = 2017,
    journal     = {ICLR},
    volume      = {abs/1610.02136}
}

@article{hendrycks2019augmix,
    title       = {AugMix: A Simple Data Processing Method to Improve Robustness and Uncertainty},
    author      = {Hendrycks, Dan and Mu, Norman and Cubuk, Ekin D and Zoph, Barret and Gilmer, Justin and Lakshminarayanan, Balaji},
    year        = 2020,
    journal     = {ICLR}
}

@article{Hendrycks2019NaturalAE,
    title       = {Natural Adversarial Examples},
    author      = {Dan Hendrycks and Kevin Zhao and Steven Basart and Jacob Steinhardt and Dawn Song},
    year        = 2019,
    journal     = {ArXiv},
    volume      = {abs/1907.07174}
}

@article{hendrycks2019oe,
    title       = {Deep Anomaly Detection with Outlier Exposure},
    author      = {Hendrycks, Dan and Mazeika, Mantas and Dietterich, Thomas},
    year        = 2019,
    journal     = {ICLR}
}

@inproceedings{Hendrycks2019Pretraining,
    title       = {Using Pre-Training Can Improve Model Robustness and Uncertainty},
    author      = {Dan Hendrycks and Kimin Lee and Mantas Mazeika},
    year        = 2019,
    booktitle   = {ICML}
}

@article{hendrycks2019robustness,
    title       = {Benchmarking Neural Network Robustness to Common Corruptions and Perturbations},
    author      = {Dan Hendrycks and Thomas Dietterich},
    year        = 2019,
    journal     = {ICLR}
}

@article{hendrycks2020pretrained,
    title       = {Pretrained Transformers Improve Out-of-Distribution Robustness},
    author      = {Hendrycks, Dan and Liu, Xiaoyuan and Wallace, Eric and Dziedzic, Adam and Krishnan, Rishabh and Song, Dawn},
    year        = 2020,
    journal     = {ACL}
}

@inproceedings{Hopfield1988NeuralNA,
    title       = {Neural networks and physical systems with emergent collective computational abilities},
    author      = {John J. Hopfield},
    year        = 1988
}

@misc{hospedales2020metalearning,
    title       = {Meta-Learning in Neural Networks: A Survey},
    author      = {Timothy Hospedales and Antreas Antoniou and Paul Micaelli and Amos Storkey},
    year        = 2020,
    eprint      = {2004.05439},
    archiveprefix = {arXiv},
    primaryclass = {cs.LG}
}

@article{Hosseini2018SemanticAE,
    title       = {Semantic Adversarial Examples},
    author      = {Hossein Hosseini and Radha Poovendran},
    year        = 2018,
    journal     = {2018 IEEE/CVF Conference on Computer Vision and Pattern Recognition Workshops (CVPRW)},
    pages       = {1695--16955}
}

@inproceedings{Hu2018GatherExciteE,
    title       = {Gather-Excite : Exploiting Feature Context in Convolutional Neural Networks},
    author      = {Jie Hu and Li Shen and Samuel Albanie and Gang Sun and Andrea Vedaldi},
    year        = 2018,
    booktitle   = {NeurIPS}
}

@article{Hu2018SqueezeandExcitationN,
    title       = {Squeeze-and-Excitation Networks},
    author      = {Jie Hu and Li Shen and Gang Sun},
    year        = 2018,
    journal     = {2018 IEEE/CVF Conference on Computer Vision and Pattern Recognition}
}

@inproceedings{huang2017speed,
    title       = {Speed/accuracy trade-offs for modern convolutional object detectors},
    author      = {Huang, Jonathan and Rathod, Vivek and Sun, Chen and Zhu, Menglong and Korattikara, Anoop and Fathi, Alireza and Fischer, Ian and Wojna, Zbigniew and Song, Yang and Guadarrama, Sergio and others},
    year        = 2017,
    booktitle   = {Proceedings of the IEEE conference on computer vision and pattern recognition},
    pages       = {7310--7311}
}

@article{hypercol,
    title       = {{Hypercolumns for Object Segmentation and Fine-grained Localization}},
    author      = {{Hariharan}, B. and {Arbel{\'a}ez}, P. and {Girshick}, R. and {Malik}, J.},
    year        = 2014,
    month       = nov,
    journal     = {ArXiv e-prints},
    archiveprefix = {arXiv},
    eprint      = {1411.5752},
    primaryclass = {cs.CV},
    adsurl      = {http://adsabs.harvard.edu/abs/2014arXiv1411.5752H}
}

@inproceedings{Ilyas2019AdversarialAreBugs,
    title       = {Adversarial Examples Are Not Bugs, They Are Features},
    author      = {Andrew Ilyas and Shibani Santurkar and D. Tsipras and L. Engstrom and B. Tran and A. Madry},
    year        = 2019,
    booktitle   = {NeurIPS}
}

@article{imagenet,
    title       = {ImageNet Large Scale Visual Recognition Challenge},
    author      = {Olga Russakovsky and Jia Deng and Hao Su and Jonathan Krause and Sanjeev Satheesh and Sean Ma and Zhiheng Huang and Andrej Karpathy and Aditya Khosla and Michael S. Bernstein and Alexander C. Berg and Fei{-}Fei Li},
    year        = 2014,
    journal     = {CoRR},
    volume      = {abs/1409.0575},
    url         = {http://arxiv.org/abs/1409.0575},
    bibsource   = {dblp computer science bibliography, http://dblp.org}
}

@inproceedings{imagenet_cvpr09,
    title       = {{ImageNet: A Large-Scale Hierarchical Image Database}},
    author      = {Deng, J. and Dong, W. and Socher, R. and Li, L.-J. and Li, K. and Fei-Fei, L.},
    year        = 2009,
    booktitle   = {CVPR09},
    bibsource   = {http://www.image-net.org/papers/imagenet_cvpr09.bib}
}

@article{instagram2018,
    title       = {Exploring the Limits of Weakly Supervised Pretraining},
    author      = {Dhruv Mahajan and Ross Girshick and Vignesh Ramanathan and Kaiming He and Manohar Paluri abd Yixuan Li and Ashwin Bharambe and Laurens van der Maaten},
    year        = 2018,
    journal     = {ECCV}
}

@article{Itakura1994,
    title       = {Recognition of Line-Drawing Representations by a Chimpanzee},
    author      = {Shoji Itakura},
    year        = 1994,
    month       = jul,
    journal     = {The Journal of General Psychology},
    publisher   = {Informa {UK} Limited},
    volume      = 121,
    number      = 3,
    pages       = {189--197},
    doi         = {10.1080/00221309.1994.9921195}
}

@article{kang,
    title       = {Testing Robustness Against Unforeseen Adversaries},
    author      = {Daniel Kang and Yi Sun and Dan Hendrycks and Tom Brown and Jacob Steinhardt},
    year        = 2019,
    journal     = {CoRR},
    volume      = {abs/1908.08016}
}

@article{Kendall2015BayesianSM,
    title       = {Bayesian SegNet: Model Uncertainty in Deep Convolutional Encoder-Decoder Architectures for Scene Understanding},
    author      = {Alex Kendall and Vijay Badrinarayanan and Roberto Cipolla},
    year        = 2015,
    journal     = {ArXiv},
    volume      = {abs/1511.02680}
}

@article{kimin,
    title       = {Training Confidence-calibrated Classifiers for Detecting Out-of-Distribution Samples},
    author      = {Kimin Lee and Honglak Lee and Kibok Lee and Jinwoo Shin},
    year        = 2018,
    journal     = {ICLR}
}

@article{Kirillov2019PanopticS,
    title       = {Panoptic Segmentation},
    author      = {Alexander Kirillov and Kaiming He and Ross B. Girshick and Carsten Rother and Piotr Doll{\'a}r},
    year        = 2019,
    journal     = {2019 IEEE/CVF Conference on Computer Vision and Pattern Recognition (CVPR)},
    pages       = {9396--9405}
}

@article{kolesnikov2019large,
    title       = {Large Scale Learning of General Visual Representations for Transfer},
    author      = {Kolesnikov, Alexander and Beyer, Lucas and Zhai, Xiaohua and Puigcerver, Joan and Yung, Jessica and Gelly, Sylvain and Houlsby, Neil},
    year        = 2019,
    journal     = {arXiv preprint arXiv:1912.11370}
}

@article{Kornblith2018DoBI,
    title       = {Do Better ImageNet Models Transfer Better?},
    author      = {Simon Kornblith and Jonathon Shlens and Quoc V. Le},
    year        = 2018,
    journal     = {CoRR},
    volume      = {abs/1805.08974}
}

@misc{kreo2018robust,
    title       = {Robust Semantic Segmentation with Ladder-DenseNet Models},
    author      = {Ivan Krešo and Marin Oršić and Petra Bevandić and Siniša Šegvić},
    year        = 2018,
    eprint      = {1806.03465},
    archiveprefix = {arXiv},
    primaryclass = {cs.CV}
}

@inproceedings{Krizhevsky2017Alexnet,
    title       = {ImageNet classification with deep convolutional neural networks},
    author      = {A. Krizhevsky and Ilya Sutskever and Geoffrey E. Hinton},
    year        = 2017,
    booktitle   = {CACM}
}

@inproceedings{kumar2019calibration,
    title       = {Verified Uncertainty Calibration},
    author      = {A. Kumar and P. Liang and T. Ma},
    year        = 2019,
    booktitle   = {Advances in Neural Information Processing Systems (NeurIPS)}
}

@article{kurakin,
    title       = {Adversarial Machine Learning at Scale},
    author      = {Alexey Kurakin and Ian Goodfellow and Samy Bengio},
    year        = 2017,
    journal     = {ICLR}
}

@article{Kurakin2017AdversarialEI,
    title       = {Adversarial examples in the physical world},
    author      = {Alexey Kurakin and Ian J. Goodfellow and Samy Bengio},
    year        = 2017,
    journal     = {ArXiv},
    volume      = {abs/1607.02533}
}

@inproceedings{Lapuschkin2019UnmaskingCH,
    title       = {Unmasking Clever Hans predictors and assessing what machines really learn},
    author      = {Sebastian Lapuschkin and Stephan W{\"a}ldchen and Alexander Binder and Gr{\'e}goire Montavon and Wojciech Samek and Klaus-Robert M{\"u}ller},
    year        = 2019,
    booktitle   = {Nature Communications}
}

@inproceedings{Lee2018Mahalanobis,
    title       = {A Simple Unified Framework for Detecting Out-of-Distribution Samples and Adversarial Attacks},
    author      = {Lee, Kimin and Lee, Kibok and Lee, H and Shin, Jinwoo},
    year        = 2018,
    booktitle   = {NeurIPS}
}

@inproceedings{Lee2020NetworkRA,
    title       = {Network Randomization: A Simple Technique for Generalization in Deep Reinforcement Learning},
    author      = {Kimin Lee and Kibok Lee and Jinwoo Shin and Honglak Lee},
    year        = 2020,
    booktitle   = {ICLR}
}

@article{Li2020OnFN,
    title       = {On Feature Normalization and Data Augmentation},
    author      = {Bo-Yi Li and Felix Wu and Ser-Nam Lim and Serge J. Belongie and Kilian Q. Weinberger},
    year        = 2020,
    journal     = {ArXiv},
    volume      = {abs/2002.11102}
}

@article{Liang2018ODIN,
    title       = {Enhancing The Reliability of Out-of-distribution Image Detection in Neural Networks},
    author      = {Shiyu Liang and Yixuan Li and R. Srikant},
    year        = 2018,
    journal     = {arXiv: Learning}
}

@inproceedings{lim2017enhanced,
    title       = {Enhanced deep residual networks for single image super-resolution},
    author      = {Lim, Bee and Son, Sanghyun and Kim, Heewon and Nah, Seungjun and Mu Lee, Kyoung},
    year        = 2017,
    booktitle   = {Proceedings of the IEEE conference on computer vision and pattern recognition workshops},
    pages       = {136--144}
}

@article{Liu2008IsolationF,
    title       = {Isolation Forest},
    author      = {F. Liu and K. Ting and Z. Zhou},
    year        = 2008,
    journal     = {2008 Eighth IEEE International Conference on Data Mining},
    pages       = {413--422}
}

@article{Liu2019FeatureDD,
    title       = {Feature Distillation: DNN-Oriented JPEG Compression Against Adversarial Examples},
    author      = {Z. Liu and Qi Liu and T. Liu and Yanzhi Wang and W. Wen},
    year        = 2019,
    journal     = {2019 IEEE/CVF Conference on Computer Vision and Pattern Recognition (CVPR)},
    pages       = {860--868}
}

@inproceedings{lof,
    title       = {LOF: identifying density-based local outliers},
    author      = {Breunig, Markus M and Kriegel, Hans-Peter and Ng, Raymond T and Sander, J{\"o}rg},
    year        = 2000,
    booktitle   = {ACM sigmod record},
    volume      = 29,
    number      = 2,
    pages       = {93--104},
    organization = {ACM}
}

@article{Lopes2019ImprovingRW,
    title       = {Improving Robustness Without Sacrificing Accuracy with Patch {Gaussian} Augmentation},
    author      = {Raphael Gontijo Lopes and Dong Yin and Ben Poole and Justin Gilmer and Ekin Dogus Cubuk},
    year        = 2019,
    journal     = {arXiv preprint arXiv:1906.02611}
}

@article{lsun,
    title       = {{LSUN:} Construction of a Large-scale Image Dataset using Deep Learning with Humans in the Loop},
    author      = {Fisher Yu and Yinda Zhang and Shuran Song and Ari Seff and Jianxiong Xiao},
    year        = 2015,
    journal     = {CoRR}
}

@article{madry,
    title       = {Towards Deep Learning Models Resistant to Adversarial Attacks},
    author      = {Aleksander Madry and Aleksandar Makelov and Ludwig Schmidt and Dimitris Tsipras and Adrian Vladu},
    year        = 2018,
    journal     = {ICLR}
}

@article{Madry2018AdversarialTraining,
    title       = {Towards Deep Learning Models Resistant to Adversarial Attacks},
    author      = {A. Madry and Aleksandar Makelov and L. Schmidt and D. Tsipras and Adrian Vladu},
    year        = 2018,
    journal     = {ArXiv},
    volume      = {abs/1706.06083}
}

@article{martin2004learning,
    title       = {Learning to detect natural image boundaries using local brightness, color, and texture cues},
    author      = {Martin, David R and Fowlkes, Charless C and Malik, Jitendra},
    year        = 2004,
    journal     = {IEEE transactions on pattern analysis and machine intelligence},
    publisher   = {IEEE},
    volume      = 26,
    number      = 5,
    pages       = {530--549}
}

@article{maurer2005algorithmic,
    title       = {Algorithmic stability and meta-learning},
    author      = {Maurer, Andreas},
    year        = 2005,
    journal     = {Journal of Machine Learning Research},
    volume      = 6,
    number      = {Jun},
    pages       = {967--994}
}

@article{Meinke2019TowardsNN,
    title       = {Towards neural networks that provably know when they don't know},
    author      = {Alexander Meinke and Matthias Hein},
    year        = 2019,
    journal     = {ArXiv},
    volume      = {abs/1909.12180}
}

@book{menard2002appliedLR,
    title       = {Applied logistic regression analysis},
    author      = {Menard, Scott},
    year        = 2002,
    publisher   = {Sage},
    volume      = 106
}

@inproceedings{mnist,
    title       = {Gradient-based learning applied to document recognition},
    author      = {Yann Lecun and Léon Bottou and Yoshua Bengio and Patrick Haffner},
    year        = 1998,
    booktitle   = {Proceedings of the IEEE},
    pages       = {2278--2324}
}

@article{mordvintsev2015inceptionism,
    title       = {Inceptionism: Going deeper into neural networks},
    author      = {Mordvintsev, Alexander and Olah, Christopher and Tyka, Mike},
    year        = 2015,
    journal     = {arXiv}
}

@article{Nguyen2015DeepNN,
    title       = {Deep neural networks are easily fooled: High confidence predictions for unrecognizable images},
    author      = {Anh Mai Nguyen and Jason Yosinski and Jeff Clune},
    year        = 2015,
    journal     = {2015 IEEE Conference on Computer Vision and Pattern Recognition (CVPR)},
    pages       = {427--436}
}

@article{objectdetectionanalysis,
    title       = {Analyzing the Performance of Multilayer Neural Networks for Object Recognition},
    author      = {Pulkit Agrawal and Ross Girshick and Jitendra Malik},
    year        = 2014,
    journal     = {ECCV}
}

@inproceedings{OC-SVM,
    title       = {Support Vector Method for Novelty Detection},
    author      = {Sch\"{o}lkopf, Bernhard and Williamson, Robert and Smola, Alex and Shawe-Taylor, John and Platt, John},
    year        = 1999,
    booktitle   = {Proceedings of the 12th International Conference on Neural Information Processing Systems},
    location    = {Denver, CO},
    publisher   = {MIT Press},
    address     = {Cambridge, MA, USA},
    series      = {NIPS'99},
    pages       = {582--588},
    numpages    = 7,
    acmid       = 3009740
}

@article{olah2017DeepDream,
    title       = {Feature Visualization: How neural networks build up their understanding of images},
    author      = {Chris Olah and Alexander Mordvintsev and Ludwig Schubert },
    year        = 2017,
    journal     = {Distill},
    doi         = {10.23915/distill.00007},
    note        = {https://distill.pub/2017/feature-visualization}
}

@article{OpenImages,
    title       = {The Open Images Dataset V4: Unified image classification, object detection, and visual relationship detection at scale},
    author      = {Alina Kuznetsova and Hassan Rom and Neil Alldrin and Jasper Uijlings and Ivan Krasin and Jordi Pont-Tuset and Shahab Kamali and Stefan Popov and Matteo Malloci and Tom Duerig and Vittorio Ferrari},
    year        = 2018,
    journal     = {arXiv:1811.00982}
}

@article{Orhan2019RobustnessPO,
    title       = {Robustness properties of Facebook's {ResNeXt WSL} models},
    author      = {A. Emin Orhan},
    year        = 2019,
    journal     = {arxiv:1907.07640}
}

@article{Patrini2017LabelNoise,
    title       = {Making Deep Neural Networks Robust to Label Noise: a Loss Correction Approach},
    author      = {Patrini, Giorgio and Rozza, Alessandro and Menon, Aditya and Nock, Richard and Qu, Lizhen},
    year        = 2017,
    journal     = {CVPR},
    booktitle   = {Proceedings of the IEEE Conference on Computer Vision and Pattern Recognition},
    pages       = {1944--1952}
}

@article{Pinggera2016LostAF,
    title       = {Lost and Found: Detecting Small Road Hazards for Self-Driving Vehicles},
    author      = {Peter Pinggera and Sebastian Ramos and Stefan K. Gehrig and Uwe Franke and Carsten Rother and Rudolf Mester},
    year        = 2016,
    journal     = {2016 IEEE/RSJ International Conference on Intelligent Robots and Systems (IROS)},
    pages       = {1099--1106}
}

@article{Recht2019DoIC,
    title       = {Do ImageNet Classifiers Generalize to ImageNet?},
    author      = {Benjamin Recht and Rebecca Roelofs and Ludwig Schmidt and Vaishaal Shankar},
    year        = 2019,
    journal     = {ArXiv},
    volume      = {abs/1902.10811}
}

@article{resnet,
    title       = {Deep Residual Learning for Image Recognition},
    author      = {Kaiming He and Xiangyu Zhang and Shaoqing Ren and Jian Sun},
    year        = 2015,
    journal     = {CoRR},
    volume      = {abs/1512.03385},
    url         = {http://arxiv.org/abs/1512.03385},
    bibsource   = {dblp computer science bibliography, http://dblp.org}
}

@article{resnext,
    title       = {Aggregated Residual Transformations for Deep Neural Networks},
    author      = {Saining Xie and Ross Girshick and Piotr Dollár and Zhuowen Tu and Kaiming He},
    year        = 2016,
    journal     = {CVPR}
}

@article{Ronneberger_2015,
    title       = {U-Net: Convolutional Networks for Biomedical Image Segmentation},
    author      = {Ronneberger, Olaf and Fischer, Philipp and Brox, Thomas},
    year        = 2015,
    journal     = {Medical Image Computing and Computer-Assisted Intervention – MICCAI 2015},
    publisher   = {Springer International Publishing},
    pages       = {234–241},
    doi         = {10.1007/978-3-319-24574-4_28},
    isbn        = 9783319245744,
    issn        = {1611-3349}
}

@article{rusak2020increasing,
    title       = {Increasing the robustness of DNNs against image corruptions by playing the Game of Noise},
    author      = {Rusak, Evgenia and Schott, Lukas and Zimmermann, Roland and Bitterwolf, Julian and Bringmann, Oliver and Bethge, Matthias and Brendel, Wieland},
    year        = 2020,
    journal     = {arXiv preprint arXiv:2001.06057}
}

@article{Sakaguchi2019WINOGRANDEAA,
    title       = {WINOGRANDE: An Adversarial Winograd Schema Challenge at Scale},
    author      = {Keisuke Sakaguchi and Ronan Le Bras and Chandra Bhagavatula and Yejin Choi},
    year        = 2019,
    journal     = {ArXiv},
    volume      = {abs/1907.10641}
}

@inproceedings{Schlkopf1999OCSVM,
    title       = {Support Vector Method for Novelty Detection},
    author      = {B. Sch{\"o}lkopf and R. Williamson and A. Smola and John Shawe-Taylor and John C. Platt},
    year        = 1999,
    booktitle   = {NIPS}
}

@article{Selvaraju_2019GradCam,
    title       = {Grad-CAM: Visual Explanations from Deep Networks via Gradient-Based Localization},
    author      = {Selvaraju, Ramprasaath R. and Cogswell, Michael and Das, Abhishek and Vedantam, Ramakrishna and Parikh, Devi and Batra, Dhruv},
    year        = 2019,
    month       = 10,
    journal     = {International Journal of Computer Vision},
    publisher   = {Springer Science and Business Media LLC},
    volume      = 128,
    number      = 2,
    pages       = {336–359},
    doi         = {10.1007/s11263-019-01228-7},
    issn        = {1573-1405},
    url         = {http://dx.doi.org/10.1007/s11263-019-01228-7}
}

@article{Selvaraju2019GradCAMVE,
    title       = {Grad-CAM: Visual Explanations from Deep Networks via Gradient-Based Localization},
    author      = {Ramprasaath R. Selvaraju and Abhishek Das and Ramakrishna Vedantam and Michael Cogswell and Devi Parikh and Dhruv Batra},
    year        = 2019,
    journal     = {International Journal of Computer Vision},
    volume      = 128,
    pages       = {336--359}
}

@article{shapenet,
    title       = {{Semantically-Enriched 3D Models for Common-sense Knowledge}},
    author      = {Manolis Savva and Angel X. Chang and Pat Hanrahan},
    year        = 2015,
    journal     = {CVPR 2015 Workshop on Functionality, Physics, Intentionality and Causality}
}

@book{shapiro_stockman_2001,
    title       = {Computer vision},
    author      = {Shapiro, Linda G. and Stockman, George C.},
    year        = 2001,
    publisher   = {Prentice Hall},
    pages       = {279–325},
    place       = {Upper Saddle River, NJ}
}

@inproceedings{sharif2016accessorize,
    title       = {Accessorize to a crime: Real and stealthy attacks on state-of-the-art face recognition},
    author      = {Sharif, Mahmood and Bhagavatula, Sruti and Bauer, Lujo and Reiter, Michael K},
    year        = 2016,
    booktitle   = {Proceedings of the 2016 ACM SIGSAC Conference on Computer and Communications Security},
    pages       = {1528--1540},
    organization = {ACM}
}

@article{Silla2010ASO,
    title       = {A survey of hierarchical classification across different application domains},
    author      = {Carlos N. Silla and Alex Alves Freitas},
    year        = 2010,
    journal     = {Data Mining and Knowledge Discovery},
    volume      = 22,
    pages       = {31--72}
}

@inproceedings{Song2018ConstructingUA,
    title       = {Constructing Unrestricted Adversarial Examples with Generative Models},
    author      = {Yang Song and Rui Shu and Nate Kushman and Stefano Ermon},
    year        = 2018,
    booktitle   = {NeurIPS}
}

@inproceedings{Stock2018ConvNetsAI,
    title       = {ConvNets and ImageNet Beyond Accuracy: Understanding Mistakes and Uncovering Biases},
    author      = {Pierre Stock and Moustapha Ciss{\'e}},
    year        = 2018,
    booktitle   = {ECCV}
}

@article{Sun_2017_ICCV,
    title       = {Revisiting Unreasonable Effectiveness of Data in Deep Learning Era},
    author      = {Sun, Chen and Shrivastava, Abhinav and Singh, Saurabh and Gupta, Abhinav},
    year        = 2017,
    journal     = {ICCV}
}

@inproceedings{Sung1995LearningAE,
    title       = {Learning and example selection for object and pattern detection},
    author      = {Kah Kay Sung},
    year        = 1995
}

@article{svhn,
    title       = {{Multi-digit Number Recognition from Street View Imagery using Deep Convolutional Neural Networks}},
    author      = {{Goodfellow}, I.~J. and {Bulatov}, Y. and {Ibarz}, J. and {Arnoud}, S. and {Shet}, V.},
    year        = 2013,
    month       = dec,
    journal     = {ArXiv e-prints},
    archiveprefix = {arXiv},
    eprint      = {1312.6082},
    primaryclass = {cs.CV},
    adsurl      = {http://adsabs.harvard.edu/abs/2013arXiv1312.6082G}
}

@article{szegedy2013intriguing,
    title       = {Intriguing properties of neural networks},
    author      = {Szegedy, Christian and Zaremba, Wojciech and Sutskever, Ilya and Bruna, Joan and Erhan, Dumitru and Goodfellow, Ian and Fergus, Rob},
    year        = 2013,
    journal     = {arXiv preprint arXiv:1312.6199}
}

@article{Szegedy2014IntriguingPO,
    title       = {Intriguing properties of neural networks},
    author      = {Christian Szegedy and Wojciech Zaremba and Ilya Sutskever and Joan Bruna and Dumitru Erhan and Ian J. Goodfellow and Rob Fergus},
    year        = 2014,
    journal     = {CoRR},
    volume      = {abs/1312.6199}
}

@article{Tanaka2006,
    title       = {Recognition of pictorial representations by chimpanzees (Pan troglodytes)},
    author      = {Masayuki Tanaka},
    year        = 2006,
    month       = dec,
    journal     = {Animal Cognition},
    publisher   = {Springer Science and Business Media {LLC}},
    volume      = 10,
    number      = 2,
    pages       = {169--179},
    doi         = {10.1007/s10071-006-0056-1}
}

@misc{taori2020when,
    title       = {When Robustness Doesn{\textquoteright}t Promote Robustness: Synthetic vs. Natural Distribution Shifts on ImageNet},
    author      = {Rohan Taori and Achal Dave and Vaishaal Shankar and Nicholas Carlini and Benjamin Recht and Ludwig Schmidt},
    year        = 2020,
    url         = {https://openreview.net/forum?id=HyxPIyrFvH}
}

@article{tencent-ml-images-2019,
    title       = {Tencent ML-Images: A Large-Scale Multi-Label Image Database for Visual Representation Learning},
    author      = {Wu, Baoyuan and Chen, Weidong and Fan, Yanbo and Zhang, Yong and Hou, Jinlong and Huang, Junzhou and Liu, Wei and Zhang, Tong},
    year        = 2019,
    journal     = {arXiv preprint arXiv:1901.01703}
}

@inproceedings{textures,
    title       = {Describing Textures in the Wild},
    author      = {M. Cimpoi and S. Maji and I. Kokkinos and S. Mohamed and A. Vedaldi},
    year        = 2014,
    booktitle   = {Computer Vision and Pattern Recognition}
}

@article{theis2017lossy,
    title       = {Lossy image compression with compressive autoencoders},
    author      = {Theis, Lucas and Shi, Wenzhe and Cunningham, Andrew and Husz{\'a}r, Ferenc},
    year        = 2017,
    journal     = {arXiv preprint arXiv:1703.00395}
}

@misc{thesis_canny,
    title       = {Towards Pixel-Level OOD Detection for Semantic Segmentation},
    author      = {Matt Angus},
    year        = 2019,
    publisher   = {UWSpace}
}

@misc{snell2017prototypical,
      title={Prototypical Networks for Few-shot Learning}, 
      author={Jake Snell and Kevin Swersky and Richard S. Zemel},
      year={2017},
      eprint={1703.05175},
      archivePrefix={arXiv},
      primaryClass={cs.LG}
}

@article{tsipras2018robustness,
    title       = {Robustness may be at odds with accuracy},
    author      = {Tsipras, Dimitris and Santurkar, Shibani and Engstrom, Logan and Turner, Alexander and Madry, Aleksander},
    year        = 2018,
    journal     = {arXiv preprint arXiv:1805.12152}
}

@inproceedings{wang2016cnn,
    title       = {Cnn-rnn: A unified framework for multi-label image classification},
    author      = {Wang, Jiang and Yang, Yi and Mao, Junhua and Huang, Zhiheng and Huang, Chang and Xu, Wei},
    year        = 2016,
    booktitle   = {Proceedings of the IEEE conference on computer vision and pattern recognition},
    pages       = {2285--2294}
}

@misc{wang2019learning,
    title       = {Learning Robust Global Representations by Penalizing Local Predictive Power},
    author      = {Haohan Wang and Songwei Ge and Eric P. Xing and Zachary C. Lipton},
    year        = 2019,
    archiveprefix = {arXiv}
}

@inproceedings{wideresnet,
    title       = {Wide Residual Networks},
    author      = {Sergey Zagoruyko and Nikos Komodakis},
    year        = 2016,
    booktitle   = {BMVC}
}

@article{wong2020fast,
    title       = {Fast is better than free: Revisiting adversarial training},
    author      = {Wong, Eric and Rice, Leslie and Kolter, J Zico},
    year        = 2020,
    journal     = {arXiv preprint arXiv:2001.03994}
}

@inproceedings{woo2018cbam,
    title       = {Cbam: Convolutional block attention module},
    author      = {Woo, Sanghyun and Park, Jongchan and Lee, Joon-Young and So Kweon, In},
    year        = 2018,
    booktitle   = {Proceedings of the European Conference on Computer Vision (ECCV)},
    pages       = {3--19}
}

@inproceedings{wu2008interactive,
    title       = {Interactive foreground/background segmentation based on graph cut},
    author      = {Wu, Xiaoyu and Wang, Yangsheng},
    year        = 2008,
    booktitle   = {2008 Congress on Image and Signal Processing},
    volume      = 3,
    pages       = {692--696},
    organization = {IEEE}
}

@article{Xiao2010SUNDL,
    title       = {SUN database: Large-scale scene recognition from abbey to zoo},
    author      = {Jianxiong Xiao and James Hays and Krista A. Ehinger and Aude Oliva and Antonio Torralba},
    year        = 2010,
    journal     = {2010 IEEE Computer Society Conference on Computer Vision and Pattern Recognition},
    pages       = {3485--3492}
}

@article{Xiao2018SpatiallyTA,
    title       = {Spatially Transformed Adversarial Examples},
    author      = {Chaowei Xiao and Jun-Yan Zhu and Bo Li and Warren He and Mingyan Liu and Dawn Xiaodong Song},
    year        = 2018,
    journal     = {CoRR},
    volume      = {abs/1801.02612}
}

@inproceedings{Xie2020Intriguing,
    title       = {Intriguing Properties of Adversarial Training at Scale},
    author      = {Cihang Xie and Alan Yuille},
    year        = 2020,
    booktitle   = {International Conference on Learning Representations}
}

@article{Yakura_2019AudioAlexAttack,
    title       = {Robust Audio Adversarial Example for a Physical Attack},
    author      = {Yakura, Hiromu and Sakuma, Jun},
    year        = 2019,
    month       = 8,
    journal     = {Proceedings of the Twenty-Eighth International Joint Conference on Artificial Intelligence},
    publisher   = {International Joint Conferences on Artificial Intelligence Organization},
    doi         = {10.24963/ijcai.2019/741},
    isbn        = 9780999241141,
    url         = {http://dx.doi.org/10.24963/ijcai.2019/741}
}

@article{yin2019fourier,
    title       = {A {F}ourier perspective on model robustness in computer vision},
    author      = {Yin, Dong and Lopes, Raphael Gontijo and Shlens, Jonathon and Cubuk, Ekin D and Gilmer, Justin},
    year        = 2019,
    journal     = {arXiv preprint arXiv:1906.08988}
}

@article{You2019AdversarialNL,
    title       = {Adversarial Noise Layer: Regularize Neural Network by Adding Noise},
    author      = {Zhonghui You and Jinmian Ye and Kunming Li and Ping Wang},
    year        = 2019,
    journal     = {2019 IEEE International Conference on Image Processing (ICIP)},
    pages       = {909--913}
}

@article{Yun2019CutMixRS,
    title       = {CutMix: Regularization Strategy to Train Strong Classifiers With Localizable Features},
    author      = {Sangdoo Yun and Dongyoon Han and Seong Joon Oh and Sanghyuk Chun and Junsuk Choe and Youngjoon Yoo},
    year        = 2019,
    journal     = {2019 IEEE/CVF International Conference on Computer Vision (ICCV)},
    pages       = {6022--6031}
}

@inproceedings{Zellers2019HellaSwagCA,
    title       = {HellaSwag: Can a Machine Really Finish Your Sentence?},
    author      = {Rowan Zellers and Ari Holtzman and Yonatan Bisk and Ali Farhadi and Yejin Choi},
    year        = 2019,
    booktitle   = {ACL}
}

@inproceedings{zendel2018wilddash,
    title       = {Wilddash-creating hazard-aware benchmarks},
    author      = {Zendel, Oliver and Honauer, Katrin and Murschitz, Markus and Steininger, Daniel and Fernandez Dominguez, Gustavo},
    year        = 2018,
    booktitle   = {Proceedings of the European Conference on Computer Vision (ECCV)},
    pages       = {402--416}
}

@article{zhang2017mixup,
    title       = {mixup: Beyond Empirical Risk Minimization},
    author      = {Hongyi Zhang and Moustapha Ciss{\'e} and Yann Dauphin and David Lopez-Paz},
    year        = 2017,
    journal     = {arXiv preprint arXiv:1710.09412}
}

@inproceedings{zhang2018perceptual,
    title       = {The Unreasonable Effectiveness of Deep Features as a Perceptual Metric},
    author      = {Zhang, Richard and Isola, Phillip and Efros, Alexei A and Shechtman, Eli and Wang, Oliver},
    year        = 2018,
    booktitle   = {CVPR}
}

@inproceedings{Zhang2019TRADES,
    title       = {Theoretically Principled Trade-off between Robustness and Accuracy},
    author      = {Hongyang Zhang and Yaodong Yu and Jiantao Jiao and Eric P. Xing and Laurent El Ghaoui and Michael I. Jordan},
    year        = 2019,
    booktitle   = {ICML}
}

@article{Zhao2017PyramidSP,
    title       = {Pyramid Scene Parsing Network},
    author      = {Hengshuang Zhao and Jianping Shi and Xiaojuan Qi and Xiaogang Wang and Jiaya Jia},
    year        = 2017,
    journal     = {2017 IEEE Conference on Computer Vision and Pattern Recognition (CVPR)},
    pages       = {6230--6239}
}

@article{Zhong2017RandomED,
    title       = {Random Erasing Data Augmentation},
    author      = {Zhun Zhong and Liang Zheng and Guoliang Kang and Shaozi Li and Yi Yang},
    year        = 2017,
    journal     = {arXiv preprint arXiv:1708.04896}
}

@article{zhou2017places,
    title       = {Places: A 10 million Image Database for Scene Recognition},
    author      = {Zhou, Bolei and Lapedriza, Agata and Khosla, Aditya and Oliva, Aude and Torralba, Antonio},
    year        = 2017,
    journal     = {PAMI}
}

@inproceedings{zhou2018learning,
    title       = {Learning rich features for image manipulation detection},
    author      = {Zhou, Peng and Han, Xintong and Morariu, Vlad I and Davis, Larry S},
    year        = 2018,
    booktitle   = {Proceedings of the IEEE Conference on Computer Vision and Pattern Recognition},
    pages       = {1053--1061}
}

@article{Zhou2019InterpretingDV,
    title       = {Interpreting Deep Visual Representations via Network Dissection},
    author      = {Bolei Zhou and David Bau and Aude Oliva and Antonio Torralba},
    year        = 2019,
    journal     = {IEEE Transactions on Pattern Analysis and Machine Intelligence},
    volume      = 41,
    pages       = {2131--2145}
}

@article{Rolnick2017MassiveLabelNoise,
  title={Deep Learning is Robust to Massive Label Noise},
  author={D. Rolnick and Andreas Veit and Serge J. Belongie and N. Shavit},
  journal={ArXiv},
  year={2017},
  volume={abs/1705.10694}
}

@article{Watanabe2013AFR,
  title={A fuzzy RANSAC algorithm based on reinforcement learning concept},
  author={Toshihiko Watanabe},
  journal={2013 IEEE International Conference on Fuzzy Systems (FUZZ-IEEE)},
  year={2013},
  pages={1-6}
}

@article{Tramr2020FundamentalMnistSensitivityChangingLabel,
  title={Fundamental Tradeoffs between Invariance and Sensitivity to Adversarial Perturbations},
  author={Florian Tram{\`e}r and Jens Behrmann and N. Carlini and Nicolas Papernot and Jorn-Henrik Jacobsen},
  journal={ArXiv},
  year={2020},
  volume={abs/2002.04599}
}

@article{engstrom2019DiscussionofAdversarialBugs,
  author = {Engstrom, Logan and Gilmer, Justin and Goh, Gabriel and Hendrycks, Dan and Ilyas, Andrew and Madry, Aleksander and Nakano, Reiichiro and Nakkiran, Preetum and Santurkar, Shibani and Tran, Brandon and Tsipras, Dimitris and Wallace, Eric},
  title = {A Discussion of 'Adversarial Examples Are Not Bugs, They Are Features'},
  journal = {Distill},
  year = {2019},
  note = {https://distill.pub/2019/advex-bugs-discussion},
  doi = {10.23915/distill.00019}
}

@article{Tramr2018EnsembleAT,
  title={Ensemble Adversarial Training: Attacks and Defenses},
  author={Florian Tram{\`e}r and A. Kurakin and Nicolas Papernot and D. Boneh and P. McDaniel},
  journal={ArXiv},
  year={2018},
  volume={abs/1705.07204}
}

@article{Tramr2020OnAdaptiveAttacks,
  title={On Adaptive Attacks to Adversarial Example Defenses},
  author={Florian Tram{\`e}r and N. Carlini and W. Brendel and A. Madry},
  journal={ArXiv},
  year={2020},
  volume={abs/2002.08347}
}

@inproceedings{Cox1958logisticregression,
  title={The Regression Analysis of Binary Sequences},
  author={D. R. Cox},
  year={1958}
}

@inproceedings{Brhanie2016MultiLabelCM,
  title={Multi-Label Classification Methods for Image Annotation},
  author={Bekalu Mullu Brhanie},
  year={2016}
}

@article{Taori2020RechtDistributionShiftImagenet,
  title={Measuring Robustness to Natural Distribution Shifts in Image Classification},
  author={Rohan Taori and Achal Dave and V. Shankar and N. Carlini and B. Recht and L. Schmidt},
  journal={ArXiv},
  year={2020},
  volume={abs/2007.00644}
}

@inproceedings{emmott2013ocsvmrobustnessreview,
  title={Systematic construction of anomaly detection benchmarks from real data},
  author={Emmott, Andrew F and Das, Shubhomoy and Dietterich, Thomas and Fern, Alan and Wong, Weng-Keen},
  booktitle={Proceedings of the ACM SIGKDD workshop on outlier detection and description},
  pages={16--21},
  year={2013}
}

@article{Charikar2017SteinhardtsemiVerified,
  title={Learning from untrusted data},
  author={M. Charikar and J. Steinhardt and G. Valiant},
  journal={Proceedings of the 49th Annual ACM SIGACT Symposium on Theory of Computing},
  year={2017}
}

@article{Ren2018ReweightExamplesLabelCorruption,
  title={Learning to Reweight Examples for Robust Deep Learning},
  author={Mengye Ren and Wenyuan Zeng and B. Yang and R. Urtasun},
  journal={ArXiv},
  year={2018},
  volume={abs/1803.09050}
}

@inproceedings{wong2018adversarialpolytope,
  title={Provable defenses against adversarial examples via the convex outer adversarial polytope},
  author={Wong, Eric and Kolter, Zico},
  booktitle={International Conference on Machine Learning},
  pages={5286--5295},
  year={2018},
  organization={PMLR}
}

@article{Breiman2004RandomForests,
  title={Random Forests},
  author={L. Breiman},
  journal={Machine Learning},
  year={2004},
  volume={45},
  pages={5-32}
}

@article{Arnab2018ConditionalRFNN,
  title={Conditional Random Fields Meet Deep Neural Networks for Semantic Segmentation: Combining Probabilistic Graphical Models with Deep Learning for Structured Prediction},
  author={A. Arnab and Shuai Zheng and Sadeep Jayasumana and Bernardino Romera-Paredes and M. Larsson and A. Kirillov and Bogdan Savchynskyy and C. Rother and F. Kahl and P. Torr},
  journal={IEEE Signal Processing Magazine},
  year={2018},
  volume={35},
  pages={37-52}
}

@InProceedings{gilmer19advconsequnce_of_noise_iclr, 
    title = {Adversarial Examples Are a Natural Consequence of Test Error in Noise}, 
    author = {Gilmer, Justin and Ford, Nicolas and Carlini, Nicholas and Cubuk, Ekin}, 
    pages = {2280--2289}, 
    year = {2019}, 
    editor = {Kamalika Chaudhuri and Ruslan Salakhutdinov}, volume = {97},
    series = {Proceedings of Machine Learning Research}, address = {Long Beach, California, USA}, 
    month = {6}, 
    publisher = {PMLR}, 
    url = {http://proceedings.mlr.press/v97/gilmer19a.html},  
}

@misc{alex2019pointrend,
    title={PointRend: Image Segmentation as Rendering},
    author={Alexander Kirillov and Yuxin Wu and Kaiming He and Ross Girshick},
    year={2019},
    eprint={1912.08193},
    archivePrefix={arXiv},
    primaryClass={cs.CV}
}

@inproceedings{Ranzato2006LecunSparseEnergyModel,
  title={Efficient Learning of Sparse Representations with an Energy-Based Model},
  author={Marc'Aurelio Ranzato and Christopher S. Poultney and S. Chopra and Y. LeCun},
  booktitle={NIPS},
  year={2006}
}

@inproceedings{Beygelzimer2006CoverTreesNN,
  title={Cover trees for nearest neighbor},
  author={A. Beygelzimer and Sham M. Kakade and J. Langford},
  booktitle={ICML '06},
  year={2006}
}

@inproceedings{Cho2009KernelMethods,
  title={Kernel Methods for Deep Learning},
  author={Youngmin Cho and L. Saul},
  booktitle={NIPS},
  year={2009}
}

@InProceedings{corel5k,
author="Duygulu, P.
and Barnard, K.
and de Freitas, J. F. G.
and Forsyth, D. A.",
editor="Heyden, Anders
and Sparr, Gunnar
and Nielsen, Mads
and Johansen, Peter",
title="Object Recognition as Machine Translation: Learning a Lexicon for a Fixed Image Vocabulary",
booktitle="Computer Vision --- ECCV 2002",
year="2002",
publisher="Springer Berlin Heidelberg",
address="Berlin, Heidelberg",
pages="97--112",
isbn="978-3-540-47979-6"
}

@inproceedings{Kosut1992SetmembershipIO,
  title={Set-membership identification of systems with parametric and nonparametric uncertainty},
  author={R. Kosut and M. Lau and Stephen P. Boyd},
  year={1992}
}

@article{Gollamudi1998SetmembershipFA,
  title={Set-membership filtering and a set-membership normalized LMS algorithm with an adaptive step size},
  author={S. Gollamudi and S. Nagaraj and S. Kapoor and Yih-Fang Huang},
  journal={IEEE Signal Processing Letters},
  year={1998},
  volume={5},
  pages={111-114}
}

@article{Werner2001SetmembershipAP,
  title={Set-membership affine projection algorithm},
  author={S. Werner and P.S.R. Diniz},
  journal={IEEE Signal Processing Letters},
  year={2001},
  volume={8},
  pages={231-235}
}

@article{Broder2003BloomFiltersSurvey,
  title={Network Applications of Bloom Filters: A Survey},
  author={A. Broder and M. Mitzenmacher},
  journal={Internet Mathematics},
  year={2003},
  volume={1},
  pages={485 - 509}
}

@article{Vinyals2016miniimagenet,
  title={Matching Networks for One Shot Learning},
  author={Oriol Vinyals and Charles Blundell and T. Lillicrap and K. Kavukcuoglu and Daan Wierstra},
  journal={ArXiv},
  year={2016},
  volume={abs/1606.04080}
}

@inproceedings{Ravi2017OptimizationAA,
  title={Optimization as a Model for Few-Shot Learning},
  author={S. Ravi and H. Larochelle},
  booktitle={ICLR},
  year={2017}
}

@article{wang2019simpleshot,
  title={SimpleShot: Revisiting Nearest-Neighbor Classification for Few-Shot Learning},
  author={Wang, Yan and Chao, Wei-Lun and Weinberger, Kilian Q.  and van der Maaten, Laurens},
  journal={arXiv preprint arXiv:1911.04623},
  year={2019}
}

@misc{howard2017mobilenets,
    title={MobileNets: Efficient Convolutional Neural Networks for Mobile Vision Applications}, 
    author={Andrew G. Howard and Menglong Zhu and Bo Chen and Dmitry Kalenichenko and Weijun Wang and Tobias Weyand and Marco Andreetto and Hartwig Adam},
    year={2017},
    eprint={1704.04861},
    archivePrefix={arXiv},
    primaryClass={cs.CV}
}

@misc{2021visiontransformer,
      title={An Image is Worth 16x16 Words: Transformers for Image Recognition at Scale}, 
      author={Alexey Dosovitskiy and Lucas Beyer and Alexander Kolesnikov and Dirk Weissenborn and Xiaohua Zhai and Thomas Unterthiner and Mostafa Dehghani and Matthias Minderer and Georg Heigold and Sylvain Gelly and Jakob Uszkoreit and Neil Houlsby},
      year={2021},
      eprint={2010.11929},
      archivePrefix={arXiv},
      primaryClass={cs.CV}
}

@article{tidake2018multilabelsurvey,
  title={Multi-label Classification: a survey},
  author={Tidake, Vaishali S and Sane, Shirish S},
  journal={International Journal of Engineering and Technology},
  volume={7},
  number={1045},
  year={2018}
}

@misc{chen2019gnn_multi_label,
      title={Learning Semantic-Specific Graph Representation for Multi-Label Image Recognition}, 
      author={Tianshui Chen and Muxin Xu and Xiaolu Hui and Hefeng Wu and Liang Lin},
      year={2019},
      eprint={1908.07325},
      archivePrefix={arXiv},
      primaryClass={cs.CV}
}

@misc{benbaruch2020asymmetric_multilabel_loss,
      title={Asymmetric Loss For Multi-Label Classification}, 
      author={Emanuel Ben-Baruch and Tal Ridnik and Nadav Zamir and Asaf Noy and Itamar Friedman and Matan Protter and Lihi Zelnik-Manor},
      year={2020},
      eprint={2009.14119},
      archivePrefix={arXiv},
      primaryClass={cs.CV}
}

@misc{wang2019multilabel_word_embeddings,
      title={Multi-Label Classification with Label Graph Superimposing}, 
      author={Ya Wang and Dongliang He and Fu Li and Xiang Long and Zhichao Zhou and Jinwen Ma and Shilei Wen},
      year={2019},
      eprint={1911.09243},
      archivePrefix={arXiv},
      primaryClass={cs.CV}
}

@incollection{jinseok2017multilabel_rnn,
title = {Maximizing Subset Accuracy with Recurrent Neural Networks in Multi-label Classification},
author = {Nam, Jinseok and Loza Menc{\'i}a, Eneldo and Kim, Hyunwoo J and F{\"u}rnkranz, Johannes},
booktitle = {Advances in Neural Information Processing Systems 30},
editor = {I. Guyon and U. V. Luxburg and S. Bengio and H. Wallach and R. Fergus and S. Vishwanathan and R. Garnett},
pages = {5413--5423},
year = {2017},
publisher = {Curran Associates, Inc.},
url = {http://papers.nips.cc/paper/7125-maximizing-subset-accuracy-with-recurrent-neural-networks-in-multi-label-classification.pdf}
}

@inproceedings{mostajabi2015zoomout,
    title={Feedforward semantic segmentation with zoom-out features},
    author={Mohammadreza Mostajabi and Payman Yadollahpour and Gregory Shakhnarovich},
    booktitle={CVPR},
    year={2015}
}

@inproceedings{mildenhall2020nerf,
 title={NeRF: Representing Scenes as Neural Radiance Fields for View Synthesis},
 author={Ben Mildenhall and Pratul P. Srinivasan and Matthew Tancik and Jonathan T. Barron and Ravi Ramamoorthi and Ren Ng},
 year={2020},
 booktitle={ECCV},
}

@article{scikit-learn,
 title={Scikit-learn: Machine Learning in {P}ython},
 author={Pedregosa, F. and Varoquaux, G. and Gramfort, A. and Michel, V.
         and Thirion, B. and Grisel, O. and Blondel, M. and Prettenhofer, P.
         and Weiss, R. and Dubourg, V. and Vanderplas, J. and Passos, A. and
         Cournapeau, D. and Brucher, M. and Perrot, M. and Duchesnay, E.},
 journal={Journal of Machine Learning Research},
 volume={12},
 pages={2825--2830},
 year={2011}
}

@article{weinstein2018visionspecies,
  title={A computer vision for animal ecology},
  author={Weinstein, Ben G},
  journal={Journal of Animal Ecology},
  volume={87},
  number={3},
  pages={533--545},
  year={2018},
  publisher={Wiley Online Library}
}

@article{ackermann2018cosmicphysics,
  title={Using transfer learning to detect galaxy mergers},
  author={Ackermann, Sandro and Schawinski, Kevin and Zhang, Ce and Weigel, Anna K and Turp, M Dennis},
  journal={Monthly Notices of the Royal Astronomical Society},
  volume={479},
  number={1},
  pages={415--425},
  year={2018},
  publisher={Oxford University Press}
}
\end{document}